\documentclass{article} 
\usepackage{iclr2026_conference,times}

\usepackage{amsmath,amsfonts,bm}

\def\eqref#1{equation~\ref{#1}}

\def\1{\bm{1}}

\DeclareMathAlphabet{\mathsfit}{\encodingdefault}{\sfdefault}{m}{sl}
\SetMathAlphabet{\mathsfit}{bold}{\encodingdefault}{\sfdefault}{bx}{n}

\newcommand{\R}{\mathbb{R}}

\usepackage{hyperref}
\usepackage{url}
\usepackage{graphicx}
\usepackage{subcaption}
\usepackage{float}
\usepackage{bm}
\usepackage{float}

\usepackage{xcolor}
\usepackage{booktabs}

\definecolor{CausalRevIN}{HTML}{FF7F00}       
\definecolor{CausalRevINasinh}{HTML}{573030}  
\definecolor{RevIN}{HTML}{00CC66}             
\definecolor{RevINasinh}{HTML}{0055FF}        
\definecolor{WURevIN}{HTML}{FF3300}           
\definecolor{WURevINasinh}{HTML}{CC0099}      

\newcommand{\causalrevin}{\textcolor{CausalRevIN}{Causal}}
\newcommand{\causalrevinasinh}{\textcolor{CausalRevINasinh}{Causal+$\sinh^{-1}$}}
\newcommand{\revin}{\textcolor{RevIN}{RevIN}}
\newcommand{\revinasinh}{\textcolor{RevINasinh}{RevIN+$\sinh^{-1}$}}
\newcommand{\wurevin}{\textcolor{WURevIN}{Prefix@$k$}}
\newcommand{\wurevinasinh}{\textcolor{WURevINasinh}{Prefix@$k$+$\sinh^{-1}$}}

\newcommand{\wurevinabl}{\textcolor{WURevIN}{Prefix@$k$ V2}}
\newcommand{\wurevinasinhabl}{\textcolor{WURevINasinh}{Prefix@$k$ V2+$\sinh^{-1}$}}

\newcommand{\mytabsize}{\fontsize{6.9pt}{1pt}\selectfont}

\title{Does Normalization Choice Matter for Causal Large Time-Series Models?}

\author{Samy-Melwan Vilhes \& Gilles Gasso \\
INSA Rouen Normandie, Univ Rouen\\
Normandie Univ, LITIS UR4108\\
F-76000 Rouen, France \\
\texttt{\{samy-melwan.vilhes,gilles.gasso\}@insa-rouen.fr} \\
\And
Mokhtar Z. Alaya \\
Univ de Technologie de Compiègne \\
LMAC EA 222 \\
F-60203 Compiègne, France \\
\texttt{\{alayaelm\}@utc.fr} \\
}

\iclrfinalcopy

\track{Research}

\begin{document}

\maketitle

\begin{abstract}
Large models for time-series forecasting have been emerged as a promising paradigm for training models on heterogeneous collections of signals.
These models typically rely on causal autoregressive architectures, where each observation is sequentially predicted from past.
In practice, real-world time-series exhibit non-stationarities, which significantly influence predictive performance.  
To mitigate this, normalization is commonly employed. However, in efficient causal settings it might induce information leakage from future observations during training. 
Recent alternatives, including causal normalization and statistics computed from initial observations, have been proposed to address this issue, but their practical implications remain insufficiently understood.
In this work, we evaluate normalization strategies for transformer-based large time-series models trained with patching and efficient causal strategy. 
We showcase that normalization choice significantly influences both training convergence and forecasting performance.
\end{abstract}

\section{Introduction}

Motivated by the success of large language models (LLMs) \citep{gpt, llama}, large time-series models aim to learn transferable representations from large-scale and heterogeneous corpora of signals. 
Towards this end, modern architectures for forecasting \citep{toto, tirex, moirai2, timemoe, timesfm, stateflow,chronos2} typically adopt patch-based representations \citep{patchtst}.  
{\it Efficient causal training} in this context refers to schemes where each patch in a sequence is predicted based on past ones, either sequentially or in parallel, while computational efficiency is maintained through reuse of intermediate representations.

However, the statistical heterogeneity of real-world time-series, manifested in varying means, variances, and distributional shifts, poses substantial challenges for performance. 
While normalization techniques such as RevIN \citep{revin}, designed to remove and restore time-series statistics, are widely used in supervised forecasting, their integration into large models with efficient causal training is non-trivial. 
In particular, computing global statistics introduces look-ahead leakage that violates causal constraints during efficient training. 
Although recent approaches propose causal-based \citep{toto, stateflow} or prefix-based \citep{timesfm, moirai2} normalization schemes, a principled understanding of how normalization choices influence training dynamics and downstream performance remains limited.

In this work, we present a unified evaluation of normalization strategies for causal large time-series models. 
We isolate normalization as the varying factor within a fixed large-scale efficient causal training framework
While we primarily consider univariate settings, our analysis can be extended to multivariate cases, and we focus on Transformer architecture~\citep{transformer}.

The most closely related prior study, \citet{comparison}, compares normalization strategies in early large time-series models \citep{chronos, moirai, gtt}, concluding that mean--variance scaling leads to better performance. 
However, their analysis does not address the constraints imposed by efficient causal training.
Consequently, strategies such as causal-based and prefix-based are not examined.

Our contributions are twofold:
$(i)$ we formalize a categorization of normalization strategies for large time-series models, dividing them into {\it vanilla, prefix}, and {\it causal variants}; 
$(ii)$ we empirically demonstrate that normalization choice play a critical role in shaping both training convergence and forecasting accuracy.

\section{A Categorization of Normalizations in Large Time-Series Models}\label{normalization}

Patching for time-series was introduced in \citet{patchtst}, where a patch-based Transformer architecture was proposed for forecasting. 
This paradigm has since become a standard component of large time-series models. 
Patches correspond to contiguous segments of a time-series and play a role analogous to tokens in language models. 
Formally, an input signal \( \mathbf{x} \in \R^{\rm{w}}\) of length \(\rm{w}\) is divided into consecutive patches of length \({\rm L}\). 

Let \(\rm{N} = \lfloor \frac{\rm{w}}{ \rm{L} } \rfloor\) be the number of patches and 
\(
\mathbf{x}_{\rm P} =
\begin{pmatrix}
\mathbf{x}_{{\rm P}_1}, \cdots, \mathbf{x}_{{\rm P}_{ \rm{N} }}
\end{pmatrix}^{\top}
\in \mathbb{R}^{{\rm N} \times {\rm L}}
\)
where the \(i\)-th patch is denoted by \(\mathbf{x}_{{\rm P}_i}\).
Models are trained to predict the next patch \(\mathbf{x}_{{\rm P}_{i+1}}\) given the previous patches \(\mathbf{x}_{{\rm P}_1}, \ldots, \mathbf{x}_{{\rm P}_i}\) for \(i = 1, \ldots, \rm{N}-1\).
Thus, a single forecast estimates the \({\rm L}\) future observations.

However, unlike discrete tokens, time-series patches are continuous-valued, which necessitates normalization to ensure robustness across signals with varying scales.
Given a patch \(\mathbf{x}_{{\rm P}_i}\), normalization is defined as:
\begin{equation}
\tilde{\mathbf{x}}_{{\rm P}_i}
= \frac{\mathbf{x}_{{\rm P}_i} - \mu_i}{\sqrt{\sigma_i^2 + \epsilon}} ,
\end{equation}
where \(\mu_i\) and \(\sigma_i^2\) are the mean and variance used to normalize \(\mathbf{x}_{{\rm P}_i}\) and computed by a given {\it normalization strategy}. Here, the parameter \(\epsilon > 0\) ensures numerical stability.

It is standard to compute normalization statistics over the entire available window, yielding the most accurate estimate of the signal distribution. 
However, under an efficient causal setting, training a model using this strategy without relaxation is impossible.
Specifically, the normalization statistics differ when forecasting \(\mathbf{x}_{{\rm P}_i}\) and \(\mathbf{x}_{{\rm P}_{i+1}}\), that prevent the reuse of computations between consecutive patches (See Appendix~\ref{formalization}).
This motivates to modify normalization strategies.

\subsection{Vanilla Reversible Instance Normalization}

RevIN \citep{revin} is a normalization–denormalization framework that normalizes input sequences, feeds them into the model, and subsequently denormalizes the outputs to address distribution shifts between different signals. 
RevIN has been adopted in recent time-series models \citep{tirex, chronos2}.
In a vanilla RevIN strategy, normalization statistics are computed as
\begin{equation}
\mu_i = \texttt{Mean}\big(\text{concat}(\mathbf{x}_{{\rm P}_1},\ldots,\mathbf{x}_{{\rm P}_{ \rm{N} }})\big),\quad
\sigma_i^2 = \texttt{Var}\big(\text{concat}(\mathbf{x}_{{\rm P}_1},\ldots,\mathbf{x}_{{\rm P}_{ \rm{N} }})\big),
\end{equation}
where each patch is normalized using the same global statistics. This allows for efficient training, but it violates causal constraints.
However, during inference, this strategy provides the most accurate statistics for normalization.

\subsection{Prefix@\texorpdfstring{$k$}{k} Reversible Instance Normalization}

To avoid statistics leakage during training, a prefix-based strategy computes normalization statistics using only the first \(k\) patches (\(k=8\) in our experiments), as proposed in \citet{timesfm, moirai2}. 
While \citet{timesfm} use only the first patch, \citet{moirai2} generalize this approach to the first \(k\) patches, with \(1 \leq k < \rm{N}\). 
To maintain training consistency, we do not forecast the first \(k\) patches; instead, we predict \(\mathbf{x}_{{\rm P}_{k+1}}, \ldots, \mathbf{x}_{{\rm P}_{\rm{N}}}\), and the normalization statistics are computed as
\begin{equation}
\mu_i = \texttt{Mean}\big(\text{concat}(\mathbf{x}_{{\rm P}_1},\ldots,\mathbf{x}_{{\rm P}_k})\big),\quad
\sigma_i^2 = \texttt{Var}\big(\text{concat}(\mathbf{x}_{{\rm P}_1},\ldots,\mathbf{x}_{{\rm P}_k})\big).
\end{equation}
A principal limitation of this strategy is its sensitivity to non-stationarity that may lead to distribution shifts occurring after the prefix window, resulting in inaccurate normalization of subsequent patches.
During inference, statistics can be computed either over the first \(k\) patches for consistency with training or over the full context as in vanilla RevIN. In our main experiments, 
we use the \(k\)-patch statistics, additional ablation results (see Appendix~\ref{warmupmoreover}) indicate that full-context statistics do not improve performance.

\subsection{Causal Reversible Instance Normalization}
 
Causal normalization computes statistics for patch \(\mathbf{x}_{{\rm P}_i}\),  using only preceding patches \citep{stateflow, toto}, in the following way:
\begin{equation}
\mu_i = \texttt{Mean}\big(\text{concat}(\mathbf{x}_{{\rm P}_1},\ldots,\mathbf{x}_{{\rm P}_i})\big),\quad
\sigma_i^2 = \texttt{Var}\big(\text{concat}(\mathbf{x}_{{\rm P}_1},\ldots,\mathbf{x}_{{\rm P}_i})\big).
\end{equation}
This strategy preserves causal integrity and eliminates look-ahead bias. 
Moreover, it enables key-value caching~\citep{kvcache} during inference, which avoids recomputation of key and value projections for previously processed patches.
This mechanism significantly reduces inference cost with transformers (with an analogous reduction achieved through the reuse of hidden states in unidirectional models).
Unlike other normalization approaches, this strategy uses patch-specific means and variances, adding only negligible computations (see Appendix~\ref{computation}).
However, the normalized representation may exhibit non-smooth transitions across patches. 

\subsection{Sinh and denormalization}
The \(\sinh^{-1}\) transformation is a continuous and monotonic function that behaves similarly to the logarithm for large values. 
Recent approaches \citep{chronos2}, building on earlier work \citep{sinh1}, apply this transformation after normalization to mitigate the influence of outliers in time-series.
We evaluate this transformation under the three normalization strategies presented above. Thus, when incorporating the \(\sinh^{-1}\) transformation, the final input becomes
 \( \tilde{{{\mathbf x}}}_{{\rm P}_i} = \sinh^{-1}\big(\tilde{{\bf x}}_{{\rm P}_i}\big) \).
During denormalization, the inverse transformation is applied:
\begin{equation}
\hat{{\bf x}}_{{\rm P}_{i+1}} = \big(\sqrt{\sigma^2_i+\epsilon}\big) \cdot \mathcal{T}^{-1} \big(\widehat{\tilde{{{\bf x}}}}_{{\rm P}_{i+1}}\big) + \mu_i,
\end{equation}
where \(\widehat{\tilde{{\bf x}}}_{{\rm P}_{i+1}}\) denotes the model prediction in the normalized space before inverse transformation and \(\mathcal{T}^{-1}\) represents the inverse of the transformation used during normalization (identity or \(\sinh\)), enabling recovery of the original scale.

\section{Experiments}

We conduct experiments to assess how these normalization strategies affect Transformer-based time-series forecasting models under efficient causal training.
The models are trained with the pinball loss \citep{tirex, moirai2, chronos2} to estimate quantiles.
We restrict our study to univariate time-series; however, the presented strategies can extend to multivariate settings by computing statistics independently for each channel.
We use a 300M-parameter Transformer architecture with patch-based inputs, described in Appendix~\ref{architecture}. 
The models are trained on a large and diverse corpus of real-world time-series from the GIFT pretraining \citep{gift} and UTSD \citep{utsd} datasets, complemented with synthetic data, comprising approximately 715M patches, training and dataset details are provided in Appendix~\ref{training}.
Evaluation is performed on test sets drawn either from the same distribution as the training data (using evaluation segments from UTSD and newly generated synthetic signals) or from distinct distributions (GIFT-Eval), across multiple context lengths $(128, 256, 512)$ 
and forecasting horizons $(32, 64, 96, \dots, 512)$, yielding approximately $115{,}000$ distinct signals with our pipeline.
During autoregressive inference, the 0.5 quantile (i.e., the median) is used as the point forecast. 
To preserve uncertainty estimates throughout the rollout over long forecasting horizons, we adopt the quantile-decoding strategy proposed in \citet{moirai2} (see Appendix~\ref{inferenceloop}).

\subsection{Training Convergence}

\begin{figure}[h]
    \centering
    \begin{minipage}[t]{0.49\linewidth} 
        \centering
        \includegraphics[width=1\linewidth]{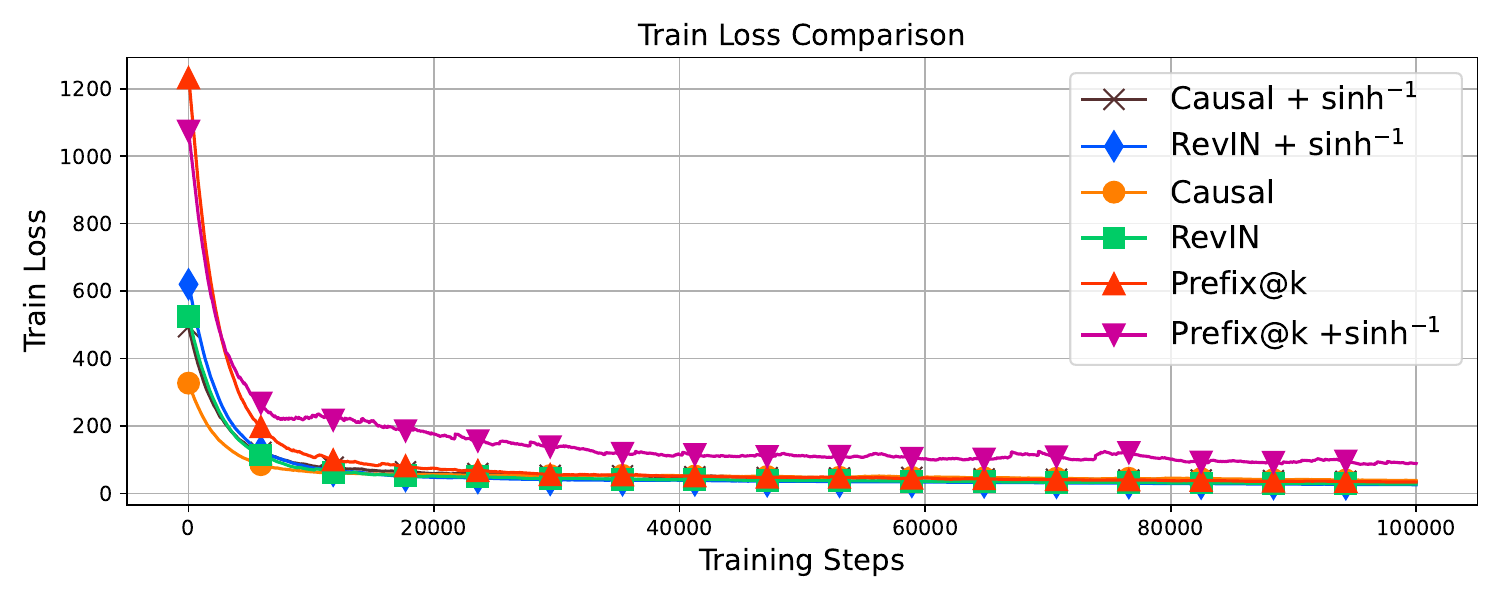}
        \caption{Training loss for first 100K steps.}
        \label{fig:lossall}
    \end{minipage}
    \begin{minipage}[t]{0.49\linewidth} 
        \centering
        \includegraphics[width=1\linewidth]{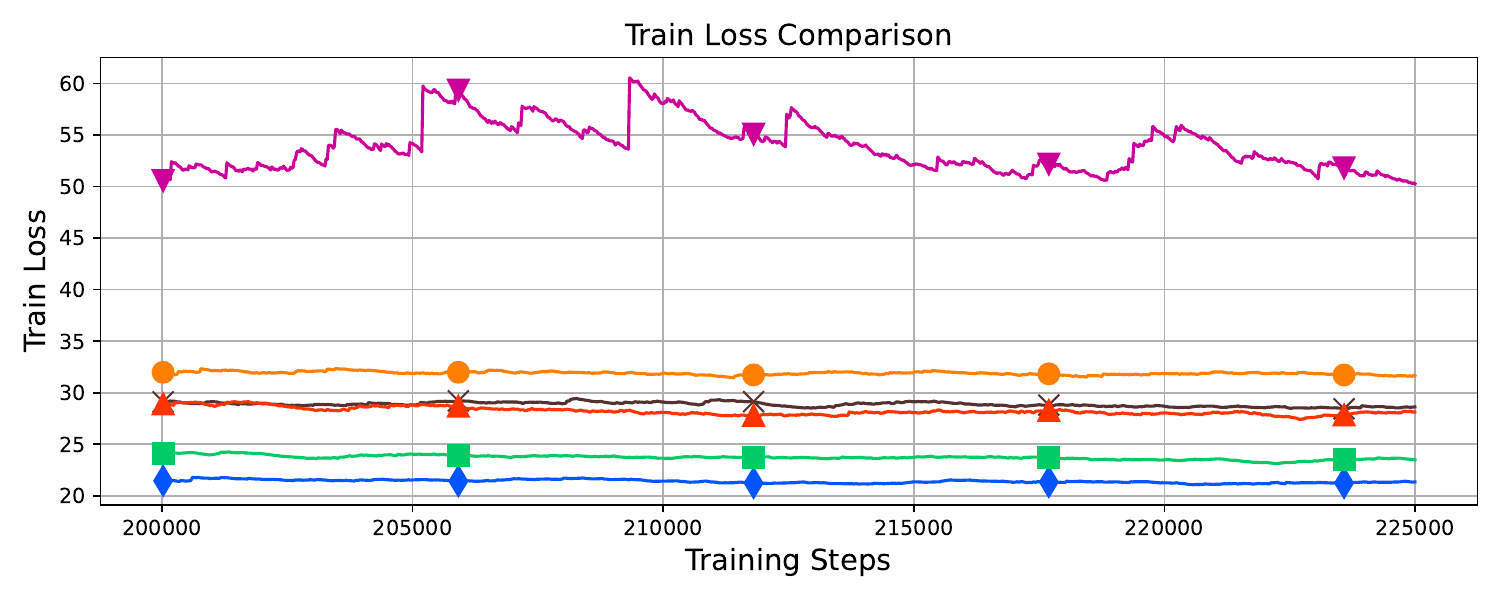}
        \caption{Training loss for 200K-225K steps.}
        \label{fig:lossend}
    \end{minipage}%
\end{figure}

Figure~\ref{fig:lossall} presents the training loss trajectories. At initialization, \causalrevin{} and \causalrevinasinh{} achieve the lowest losses, followed by \revin{} and \revinasinh{}, whereas \wurevin{} and \wurevinasinh{} exhibit substantially higher initial losses.  
Forecasts produced by causal-based strategies better match the local scale of each ground-truth patch, as they rely on patch-specific statistics, which likely accounts for their lower initial loss.  
In contrast, prefix-based strategies are more sensitive to non-stationarities in the training data: their forecasts are scaled according to the first \(k\) patches, which may differ from the scale of subsequent patches, leading to higher initial losses.  
Vanilla strategies occupy an intermediate position, with initial losses higher than those of causal-based strategies but lower than those of prefix-based strategies. 
As training progresses, the ranking changes: \revinasinh{} and \revin{} reach the lowest minima as it can be seen in Figure~\ref{fig:lossend}.
The \causalrevinasinh{} and \wurevin{} strategies converge to a comparable secondary plateau. 
Finally, \causalrevin{} stabilizes at a higher loss, while \wurevinasinh{} exhibits both the highest loss and the greatest variability (see discussion in Appendix~\ref{intuition}).

\subsection{Forecasting Performance}

Across all test signals, context lengths, and forecasting horizons, 
two strategies seem to outperform the others: \causalrevinasinh{} and \wurevin{}, in terms of MAE, RMSE, MASE (Eq.~\ref{mase}) and SQL (Eq.~\ref{sql}) as reported in Table~\ref{tab:results}.
With respect to context length \(\rm{w}>128\), \causalrevinasinh{} achieves the strongest overall performance, with \wurevin{} ranking second.
In contrast, \wurevin{} attains the best results for context length \(\rm{w}=128\), followed by \revinasinh{} and \causalrevinasinh{}.
The \wurevinasinh{} strategy systematically underperforms across all metrics and context lengths. 
\revin{}, \revinasinh{}, and \causalrevin{} exhibit intermediate performance. 
Notably, the inclusion of the $\sinh^{-1}$ transformation improves the performance of \causalrevin{} but degrades that of \wurevin{}. 
Additional results, including pairwise skill scores and evaluations across other context lengths, are reported in Appendix~\ref{moreresults} and confirm our findings considering the leaders.

\begin{table}[h]
\centering
\mytabsize
\setlength{\tabcolsep}{3pt} 
\caption{Critical difference diagram \citep{cddiag} results across context sizes \(\rm{w}\) for each normalization strategy aggregated by datasets and forecasting horizons. The best overall mean rank for each metric and context size is shown in \textbf{bold}, and the second-best is \underline{underlined} (lower ranks indicate better performance). 
Strategies connected within the same group (\(\text{rank}^{\text a}\)) are not statistically distinguishable according to the Nemenyi test with $\alpha=0.05$. Best group is a, second-best group is b, and so on. 
Results are aggregated over all test signals and forecasting horizons.}
\label{tab:results}
\begin{tabular}{@{}c|ccc|ccc|ccc|ccc|ccc|ccc@{}}
\midrule
     & \multicolumn{3}{c|}{\wurevin}
     & \multicolumn{3}{c|}{\wurevinasinh}
     & \multicolumn{3}{c|}{\revin}
     & \multicolumn{3}{c|}{\revinasinh}
     & \multicolumn{3}{c|}{\causalrevin}
     & \multicolumn{3}{c}{\causalrevinasinh} \\
     \midrule
    \(\rm{w}\) & 128 & 256 & 512 
     & 128 & 256 & 512
     & 128 & 256 & 512
     & 128 & 256 & 512
     & 128 & 256 & 512
     & 128 & 256 & 512 \\
\midrule

MAE  & \textbf{2.7}\textsuperscript{\tiny  a}    & \underline{2.8}\textsuperscript{\tiny  ab}    & \underline{2.9}\textsuperscript{\tiny  b} & 4.1\textsuperscript{\tiny  bc}  & 4.6\textsuperscript{\tiny  c}       & 5.2\textsuperscript{\tiny  c}    & 3.1\textsuperscript{\tiny  ab}     & 4.1\textsuperscript{\tiny  c}     & 4.0\textsuperscript{\tiny  b} & \underline{3.0}\textsuperscript{\tiny  ab}           & 4.1\textsuperscript{\tiny  c}    & 3.7\textsuperscript{\tiny  b}  & 4.5\textsuperscript{\tiny  c}       & 3.7\textsuperscript{\tiny  bc}       & 3.5\textsuperscript{\tiny  b}   & 3.8\textsuperscript{\tiny  abc} & \textbf{1.7}\textsuperscript{\tiny  a} & \textbf{1.6}\textsuperscript{\tiny  a} \\
RMSE  & \textbf{2.3}\textsuperscript{\tiny  a} & \underline{2.7}\textsuperscript{\tiny  ab}    & \underline{2.9}\textsuperscript{\tiny  b} & 4.1\textsuperscript{\tiny  b}       & 4.6\textsuperscript{\tiny  c}       & 5.3\textsuperscript{\tiny  d}   & 3.3\textsuperscript{\tiny  ab}     & 4.2\textsuperscript{\tiny  c}     &  4.3\textsuperscript{cd} &  \underline{3.2}\textsuperscript{\tiny  ab}     & 4.1\textsuperscript{\tiny  c}    & 3.7\textsuperscript{\tiny  bc} & 4.3\textsuperscript{\tiny  b}       & 3.7\textsuperscript{bc}       & 3.5\textsuperscript{\tiny  bc}  & 3.9\textsuperscript{\tiny  b}          & \textbf{1.6}\textsuperscript{\tiny  a} & \textbf{1.4}\textsuperscript{\tiny  a} \\
MASE & \textbf{2.0}\textsuperscript{\tiny  a} & \underline{1.6}\textsuperscript{\tiny  a} & \textbf{2.1}\textsuperscript{\tiny  a} & 3.9\textsuperscript{\tiny  b}      & 4.5\textsuperscript{\tiny  c}       & 4.6\textsuperscript{\tiny  c}  & 4.1\textsuperscript{\tiny  b}     & 5.2\textsuperscript{\tiny  c}     & 5.0\textsuperscript{\tiny  c} & 4.3\textsuperscript{\tiny  b}    & 5.1\textsuperscript{\tiny  c}    & 4.0\textsuperscript{\tiny  bc} & 4.2\textsuperscript{\tiny  b}     & 3.2\textsuperscript{\tiny  b}       & 3.1\textsuperscript{\tiny  ab}  & \underline{2.5}\textsuperscript{\tiny  a}   & \textbf{1.4}\textsuperscript{\tiny  a}    & \underline{2.1}\textsuperscript{\tiny  a}    \\
SQL & \textbf{2.1}\textsuperscript{\tiny  a} & \underline{1.8}\textsuperscript{\tiny  a} & \textbf{2.2}\textsuperscript{\tiny  a} & 4.1\textsuperscript{\tiny  b}      & 3.9\textsuperscript{\tiny  b}       & 4.2\textsuperscript{\tiny  b}   & 4.2\textsuperscript{\tiny  b}     & 5.4\textsuperscript{\tiny  c}     & 4.7\textsuperscript{\tiny  b}    & 4.6\textsuperscript{\tiny  b}  & 5.4\textsuperscript{\tiny  c}    & 4.5\textsuperscript{\tiny  b}  & 3.7\textsuperscript{\tiny  b}     & 3.1\textsuperscript{\tiny  b}       & 3.0\textsuperscript{\tiny  a} & \underline{2.3}\textsuperscript{\tiny  a}   & \textbf{1.4}\textsuperscript{\tiny  a}    & \underline{2.4}\textsuperscript{\tiny  a}    \\
\midrule
\end{tabular}
\end{table}

\section{Conclusion}

In this work, we propose a categorization of normalization strategies for large time-series models, highlighting the trade-off between causal integrity and forecasting accuracy. Our empirical analysis suggests two main insights for model design. First, we observe 
that although statistical leakage during training in \revin{} and \revinasinh{} can lead to lower training loss, this advantage does 
not systematically translate into improved downstream performance. This indicates that preserving causal integrity may contribute 
to better forecasting accuracy.
Second, our results indicate that \causalrevinasinh{} and \wurevin{} are promising candidates for large-scale training. While they 
reach comparable asymptotic training losses, they exhibit complementary context-size-dependent behavior: \causalrevinasinh{} tends to 
perform better at larger context sizes, whereas \wurevin{} often achieves stronger results for shorter context lengths. Together, these observations provide 
practical guidance for choosing normalization strategies that accommodate the diverse statistical characteristics of heterogeneous 
time-series data while preserving efficient causal training.

\bibliography{iclr2026_conference}
\bibliographystyle{iclr2026_conference}

\appendix
\section{Appendix}

The code is available at \href{https://github.com/vilhess/normalizer}{https://github.com/vilhess/normalizer}.

\subsection{Formalization of the Problem Setting}\label{formalization}

We consider the task of univariate time-series forecasting using patch-based representations.  
Let a time-series be partitioned into a sequence of patches
\begin{equation}
\mathbf{x} =(\mathbf{x}_{\rm P _1}, \mathbf{x}_{\rm P_ 2}, \dots, \mathbf{x}_{\rm P_ {\rm{N}}}), 
\end{equation}
where each patch \(\mathbf{x}_{\rm P_ i} \in \mathbb{R}^{\rm{L}}\) corresponds to a contiguous segment of length \(\rm{L}\).
Given the first \(n\) patches, the objective is to predict the next patch \(\mathbf{x}_{{\rm P}_ {n+1}}\).

We denote by \(f_\theta\) a causal forecasting model parameterized by \(\theta\).  
The prediction of the next patch is defined as
\begin{equation}
\hat{\mathbf{x}}_{{\rm P}_ {n+1}} = f_{\theta, n}(\mathbf{x}_{\rm P_ 1}, \dots, \mathbf{x}_{{\rm P}_ n}),
\end{equation}
where causality means that hidden representations of each patch depend only on past ones.

During training, the model is optimized to minimize a loss function \(\mathcal{L}\) over all patches in the sequence:
\begin{equation}
\min_{\theta} \; 
\mathbb{E}_{\mathbf{x} =(\mathbf{x}_{{\rm P}_ 1}, \dots, \mathbf{x}_{{\rm P}_ {\rm{N}}}) \sim \mathcal{D}}
\left[
\frac{1}{\rm{N}-1} \sum_{n=1}^{\rm{N}-1} 
\mathcal{L}\big(f_{\theta, n}(\mathbf{x}_{{\rm P}_ 1}, \dots, \mathbf{x}_{{\rm P}_ n}), \mathbf{x}_{{\rm P}_ {n+1}}\big)
\right],
\end{equation}
where \(\mathcal{D}\) denotes the training distribution.

\paragraph{Causal Self-Attention Model.}
For clarity, we instantiate \(f_\theta\) as a single-head self-attention mechanism with causal masking.
A similar analysis can be done for LSTM-based models, where the hidden state at each step depends only on past inputs, and the same normalization issues arise.
The prediction for the next patch \(\hat{\mathbf{x}}_{{\rm P}_ {n+1}}\) is computed as
\begin{equation}
f_{\theta, n}(\mathbf{x}_{{\rm P}_ 1}, \dots, \mathbf{x}_{{\rm P}_ n})
= \sum_{i=1}^{n} \alpha_{n,i} \, W_V \mathbf{x}_{{\rm P}_ i},
\end{equation}
where the attention weights are defined as
\begin{equation}
\alpha_{n,i} = 
\frac{\exp\big((W_Q \mathbf{x}_{{\rm P}_ n})^\top (W_K \mathbf{x}_{{\rm P}_ i})\big)}
{\sum_{j=1}^{n} \exp\big((W_Q \mathbf{x}_{{\rm P}_ n})^\top (W_K \mathbf{x}_{{\rm P}_ j})\big)}.
\end{equation}
Here \(W_Q, W_K, W_V \in \mathbb{R}^{\rm{L} \times \rm{L}}\) are the query, key, and value projection matrices, and \(\theta = \{W_Q, W_K, W_V\}\).

Then, when forecasting the subsequent patch \(\hat{\mathbf{x}}_{\rm P_ {n+2}}\), the model computes
\begin{equation}
f_{\theta, n+1}(\mathbf{x}_{{\rm P}_ 1}, \dots, \mathbf{x}_{{\rm P}_ {n+1}})
= \sum_{i=1}^{n+1} \alpha_{n+1,i} \, W_V \mathbf{x}_{{\rm P}_ i},
\end{equation}
with updated attention weights
\begin{equation}
\alpha_{n+1,i} = 
\frac{\exp\big((W_Q \mathbf{x}_{{\rm P}_ {n+1}})^\top (W_K \mathbf{x}_{{\rm P}_ i})\big)}
{\sum_{j=1}^{n+1} \exp\big((W_Q \mathbf{x}_{{\rm P}_ {n+1}})^\top (W_K \mathbf{x}_{{\rm P}_ j})\big)}.
\end{equation}
Notably, the previously computed projections
\begin{equation}(W_K \mathbf{x}_{{\rm P}_ i}, \, W_V \mathbf{x}_{{\rm P}_ i}) \quad \text{for } i = 1, \dots, n\end{equation}
can be reused, enabling efficient incremental computation.

Furthermore, transformers architectures can perform all predictions during efficient causal training in parallel using matrix operations and causal masking during training, 
making it very efficient during training.
{\small
\begin{equation*}
\underbrace{
\begin{pmatrix}
\hat{\mathbf{x}}_2 \\
\hat{\mathbf{x}}_3 \\

\vdots \\
\hat{\mathbf{x}}_{n+1}
\end{pmatrix}
}_{\in \mathbb{R}^{n \times \rm{L}}}
=
\mathrm{RowSoftmax}\!\left(
\underbrace{
\begin{pmatrix}
W_Q \mathbf{x}_1 \\
W_Q \mathbf{x}_2 \\
\vdots \\
W_Q \mathbf{x}_n
\end{pmatrix}
}_{\mathbb{R}^{n \times \rm{L}}}
\;
\underbrace{
\begin{pmatrix}
W_K \mathbf{x}_1 \\
W_K \mathbf{x}_2 \\
\vdots \\
W_K \mathbf{x}_n
\end{pmatrix}^{\!\top}
}_{\mathbb{R}^{\rm{L} \times n}}
\;\odot\;
\left(
-\infty \cdot
\underbrace{
\begin{pmatrix}
1 & 0 & \cdots & 0 \\
1 & 1 & \cdots & 0 \\
\vdots & \vdots & \ddots & \vdots \\
1 & 1 & \cdots & 1
\end{pmatrix}
}_{\mathbb{R}^{n \times n}}
\right)
\right)
\;
\underbrace{
\begin{pmatrix}
W_V \mathbf{x}_1 \\
W_V \mathbf{x}_2 \\
\vdots \\
W_V \mathbf{x}_n
\end{pmatrix}
}_{\mathbb{R}^{n \times \rm{L}}}
\end{equation*}
}
\noindent
RowSoftmax applies the softmax operation independently to each row.
The strictly lower-triangular mask enforces causality by preventing each query
from attending to future patches.

This formulation naturally extends to multi-head attention and deeper Transformer architectures.

\paragraph{Normalization in the Patch Space.}
In practice, forecasting is commonly performed in the normalized space.
Given the first \(n\) patches, we define the empirical mean and variance as
\begin{equation}
\mu_n = \frac{1}{n} \sum_{i=1}^{n} \bigg( \frac{1}{{\rm L}} \sum_{j=1}^{\rm{L}} \mathbf{x}_{{\rm P}_{i,j}} \bigg),
\quad
\sigma_n^2 = \frac{1}{n} \sum_{i=1}^{n} \bigg( \frac{1}{{\rm L}} \sum_{j=1}^{\rm{L}} (\mathbf{x}_{{\rm P}_{i,j}} - \mu_n)^2 \bigg).
\end{equation}
Each patch is then normalized as
\begin{equation}
\tilde{{\bf x}}_{{\rm P}_ i}^{(n)} = \frac{{\bf x}_{{\rm P}_ i} - \mu_n}{\sqrt{\sigma_n^2 + \epsilon}},
\end{equation}
where \(\epsilon > 0\) ensures numerical stability.

The model predicts the next normalized patch:
\begin{equation}
\widehat{\tilde{\bf x}}_{{\rm P}_ {n+1}} = f_{\theta, n}(\tilde{\mathbf{x}}_{{\rm P}_ 1}^{(n)}, \dots, \tilde{\mathbf{x}}_{{\rm P}_ n}^{(n)}),
\end{equation}
and the prediction is mapped back to the original space via
\begin{equation}
\hat{\mathbf{x}}_{{\rm P}_ {n+1}} = \widehat{\tilde{\bf x}}_{{\rm P}_ {n+1}} \cdot \sqrt{\sigma_n^2 + \epsilon} + \mu_n.
\end{equation}

\paragraph{Impact of Normalization on Causal Computation.}
When using normalized inputs, the attention computation becomes
\begin{equation}
f_{\theta, n}(\tilde{\mathbf{x}}_{{\rm P}_ 1}^{(n)}, \dots, \tilde{\mathbf{x}}_{{\rm P}_ n}^{(n)})
= \sum_{i=1}^{n} \tilde{\alpha}_{n,i} \, W_V {\tilde{\mathbf{x}}}_{{\rm P}_ i}^{(n)},
\end{equation}
with
\begin{equation}
\tilde{\alpha}_{n,i} = 
\frac{\exp\big((W_Q \tilde{\mathbf{x}}_{{\rm P}_ n}^{(n)})^\top (W_K \tilde{\mathbf{x}}_{{\rm P}_ i}^{(n)})\big)}
{\sum_{j=1}^{n} \exp\big((W_Q \tilde{{\bf x}}_{{\rm P}_ n}^{(n)})^\top (W_K \tilde{\mathbf{\bf x}}_{{\rm P}_ j}^{(n)})\big)}.
\end{equation}

Crucially, when a new patch \(\mathbf{x}_{{\rm P}_ {n+1}}\) becomes available, the normalization statistics change from \((\mu_n, \sigma_n^2)\) to \((\mu_{n+1}, \sigma_{n+1}^2)\).
As a consequence, all previously normalized patches must be recomputed:
\begin{equation}
\tilde{\mathbf{x}}_{{\rm P}_ i}^{(n+1)} \neq \tilde{\mathbf{x}}_{{\rm P}_ i}^{(n)}, \quad \forall i \leq n.
\end{equation}
Therefore, previously computed key and value projections
\begin{equation}
(W_K \tilde{\mathbf{x}}_{{\rm P}_ i}^{(n)}, \, W_V \tilde{\mathbf{x}}_{{\rm P}_ i}^{(n)}) \quad \text{for } i = 1, \dots, n
\end{equation}
cannot be reused.

This dependency breaks the efficient incremental computation property during training with Transformers.
Hence, standard normalization schemes are fundamentally incompatible with causal pretraining unless one accepts either statistical leakage or modified normalization strategies.

This observation motivates the normalization strategies studied in the main paper.

\subsection{Prefix Normalization Discussion}\label{warmupmoreover}

For Prefix normalization (\wurevin{} and \wurevinasinh{}), we evaluate a single prefix length of \(k=8\) patches. 
This value is chosen as a practical compromise between obtaining sufficiently stable normalization statistics and limiting the prefix window to reduce the risk of distribution shifts within the context. 
Nevertheless, the choice of \(k\) remains an important hyperparameter, and a systematic investigation of its impact on convergence and forecasting performance is left for future work.

During inference, normalization statistics can be computed either using the first \(k\) patches, in order to remain consistent with the training procedure, or using the full context (vanilla strategy). 
In the main experiments, we adopt the \(k\)-patch statistics (allowing the use of key--value caching) and additionally evaluate the use of full-context statistics at inference time. 
The results indicate that this alternative does not lead to performance improvements for either \wurevin{} or \wurevinasinh{}. 
Specifically, \wurevin{} achieves the best overall performance, followed by its full-context variant \wurevinabl{}, then \wurevinasinh{} and its corresponding full-context variant \wurevinasinhabl{}, as reported in Table~\ref{tab:results_warmup}.

\begin{table}[h]
\centering
\mytabsize
\setlength{\tabcolsep}{3pt} 
\caption{Critical difference diagram \citep{cddiag} results across context sizes \(\rm{w}\) for prefix normalization strategies aggregated by datasets and forecasting horizons. The best overall mean rank for each metric and context size is shown in \textbf{bold}, and the second-best is \underline{underlined} (lower ranks indicate better performance). 
Strategies connected within the same group (\(\text{rank}^{\text a}\)) are not statistically distinguishable according to the Nemenyi test with $\alpha=0.05$. Best group is a, second-best group is b, and so on. 
Results are aggregated over all test signals and forecasting horizons.}
\label{tab:results_warmup}
\begin{tabular}{@{}c|ccccc|ccccc|ccccc|ccccc@{}}
\midrule
     & \multicolumn{5}{c|}{\wurevin}
     & \multicolumn{5}{c|}{\wurevinabl}
     & \multicolumn{5}{c|}{\wurevinasinh}
     & \multicolumn{5}{c|}{\wurevinasinhabl} \\
     \midrule
     \(\rm{w}\) & 64 & 128 & 256 & 512 & 1024 
     & 64 & 128 & 256 & 512 & 1024
     & 64 & 128 & 256 & 512 & 1024
     & 64 & 128 & 256 & 512 & 1024 \\
\midrule
    
MAE & \textbf{1.4}\textsuperscript{\tiny  a} & \underline{1.7}\textsuperscript{\tiny  a}   & \textbf{1.5}\textsuperscript{\tiny  a}    & \textbf{1.2}\textsuperscript{\tiny  a} & \textbf{1.3}\textsuperscript{\tiny  a} & \underline{1.6}\textsuperscript{\tiny  a}  & \textbf{1.3}\textsuperscript{\tiny  a}       & \underline{1.7}\textsuperscript{\tiny  a}       & \underline{2.1}\textsuperscript{\tiny  b} & \underline{2.0}\textsuperscript{\tiny  a}       & 3.7\textsuperscript{\tiny  b}  & 3.3\textsuperscript{\tiny  b}     & 3.4\textsuperscript{\tiny  b}     & 3.2\textsuperscript{\tiny  c} & 3.4\textsuperscript{\tiny  b}     & 3.3\textsuperscript{\tiny  b}  & 3.7\textsuperscript{\tiny  b}      & 3.5\textsuperscript{\tiny  b}    & 3.6\textsuperscript{\tiny  c} & 3.2\textsuperscript{\tiny  b} \\
RMSE & \textbf{1.4}\textsuperscript{\tiny  a} & \underline{1.6}\textsuperscript{\tiny  a} & \textbf{1.5}\textsuperscript{\tiny  a}    & \textbf{1.1}\textsuperscript{\tiny  a} & \textbf{1.3}\textsuperscript{\tiny  a}  & \underline{1.7}\textsuperscript{\tiny  a}   & \textbf{1.4}\textsuperscript{\tiny  a}       & \underline{2.0}\textsuperscript{\tiny  a}       & \underline{2.3}\textsuperscript{\tiny  b} & \underline{2.1}\textsuperscript{\tiny  b}      & 3.7\textsuperscript{\tiny  b}   & 3.3\textsuperscript{\tiny  b}    & 3.3\textsuperscript{\tiny  b}     & 3.1\textsuperscript{\tiny  c}  & 3.2\textsuperscript{\tiny  c}    & 3.3\textsuperscript{\tiny  b} & 3.7\textsuperscript{\tiny  b}     & 3.2\textsuperscript{\tiny  b}    & 3.4\textsuperscript{\tiny  c} & 3.4\textsuperscript{\tiny  c} \\
MASE & \textbf{1.2}\textsuperscript{\tiny  a} & \textbf{1.4}\textsuperscript{\tiny  a} & \textbf{1.2}\textsuperscript{\tiny  a} & \textbf{1.7}\textsuperscript{\tiny  a} & \textbf{1.3}\textsuperscript{\tiny  a}  &       \underline{1.8}\textsuperscript{\tiny  a}   & \underline{1.6}\textsuperscript{\tiny  a}       & \underline{2.2}\textsuperscript{\tiny  b}       & \underline{2.0}\textsuperscript{\tiny  a}   & \underline{2.1}\textsuperscript{\tiny  b} & 3.7\textsuperscript{\tiny  b}   & 3.3\textsuperscript{\tiny  b}     & 2.8\textsuperscript{\tiny  b}     & 3.0\textsuperscript{\tiny  b} & 2.9\textsuperscript{\tiny  c}    & 3.3\textsuperscript{\tiny  b}  & 3.7\textsuperscript{\tiny  b}           & 3.9\textsuperscript{\tiny  c}    & 3.3\textsuperscript{\tiny  b}  & 3.7\textsuperscript{\tiny  d} \\
SQL & \textbf{2.3}\textsuperscript{\tiny  a} & \textbf{1.4}\textsuperscript{\tiny  a} & \textbf{1.2}\textsuperscript{\tiny  a} & \textbf{1.7}\textsuperscript{\tiny  a} & \textbf{1.4}\textsuperscript{\tiny  a} &         3.0\textsuperscript{\tiny  a}   & \underline{1.7}\textsuperscript{\tiny  a}       & \underline{2.2}\textsuperscript{\tiny  b}       & \underline{2.1}\textsuperscript{\tiny  a}   & \underline{2.2}\textsuperscript{\tiny  b} & \underline{2.3}\textsuperscript{\tiny  a}   & 3.1\textsuperscript{\tiny  b}     & 2.8\textsuperscript{\tiny  b}     & 3.0\textsuperscript{\tiny  b} & 2.8\textsuperscript{\tiny  b}    & 2.4\textsuperscript{\tiny  b}  & 3.8\textsuperscript{\tiny  c}           & 3.8\textsuperscript{\tiny  c}    & 3.2\textsuperscript{\tiny  b}  & 3.6\textsuperscript{\tiny  c} \\
\midrule
\end{tabular}
\end{table}

\subsection{Computational Overhead}\label{computation}

\begin{figure}[h]
    \centering
    
    \begin{subfigure}[t]{0.49\linewidth}
        \centering
        \includegraphics[width=\linewidth]{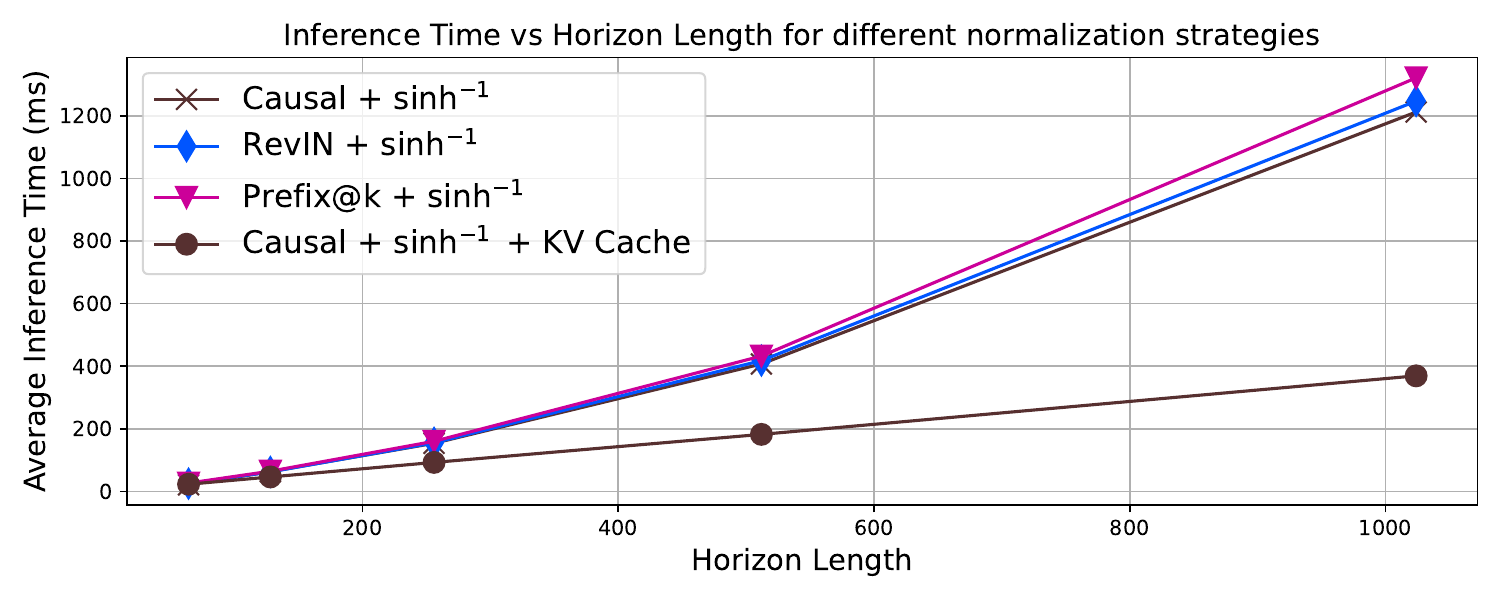}
        \caption{Context length $128$}
    \end{subfigure}
    \hfill
    \begin{subfigure}[t]{0.49\linewidth}
        \centering
        \includegraphics[width=\linewidth]{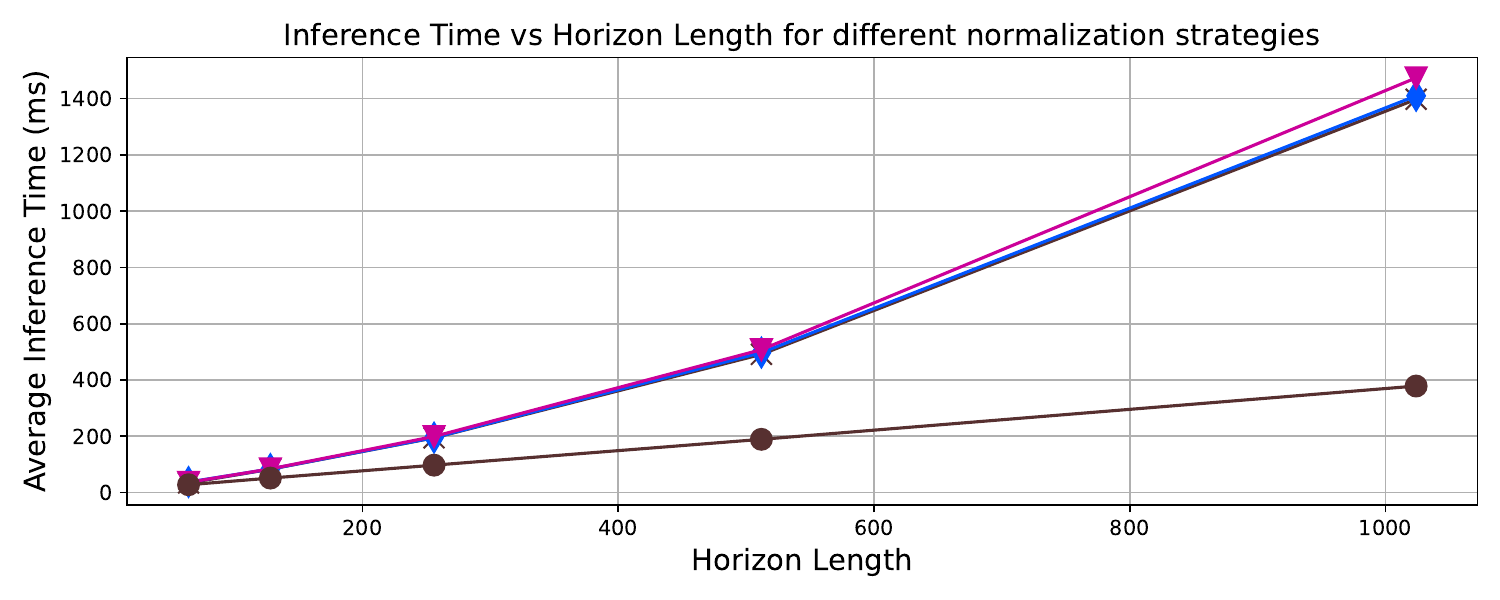}
        \caption{Context length $256$}
    \end{subfigure}

    \begin{subfigure}[t]{0.49\linewidth}
        \centering
        \includegraphics[width=\linewidth]{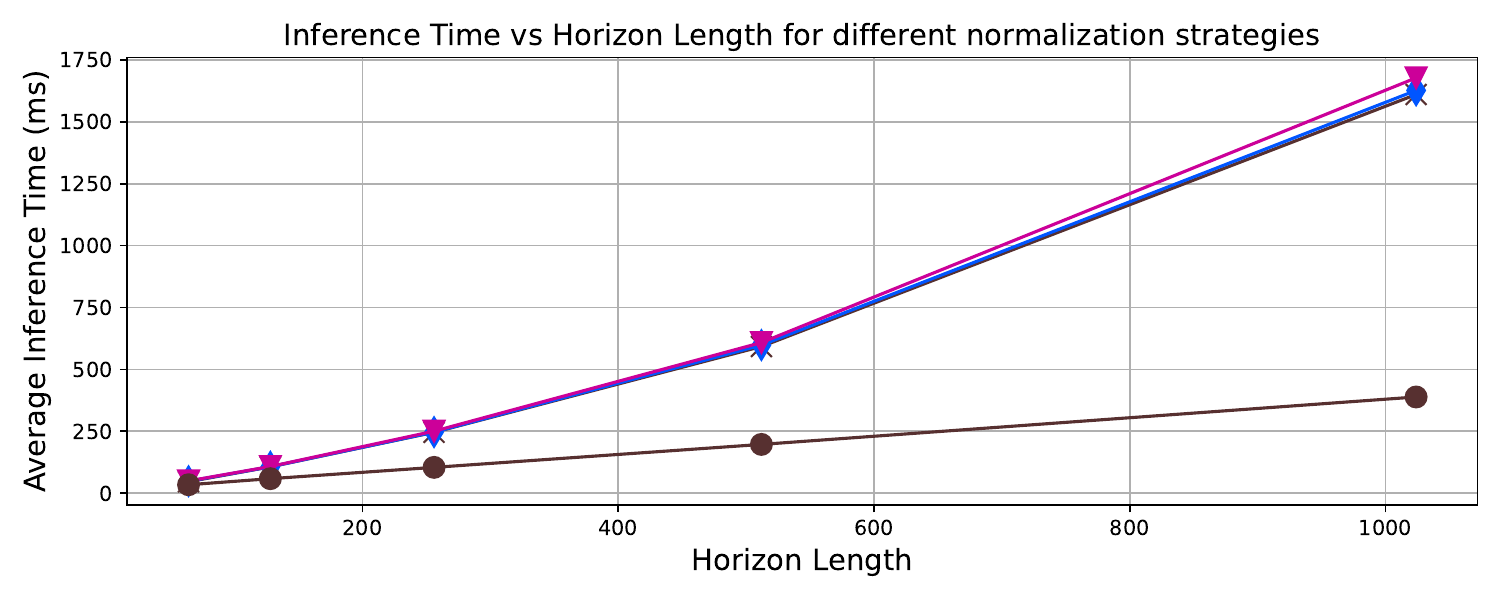}
        \caption{Context length $512$}
    \end{subfigure}
    \hfill
    \begin{subfigure}[t]{0.49\linewidth}
        \centering
        \includegraphics[width=\linewidth]{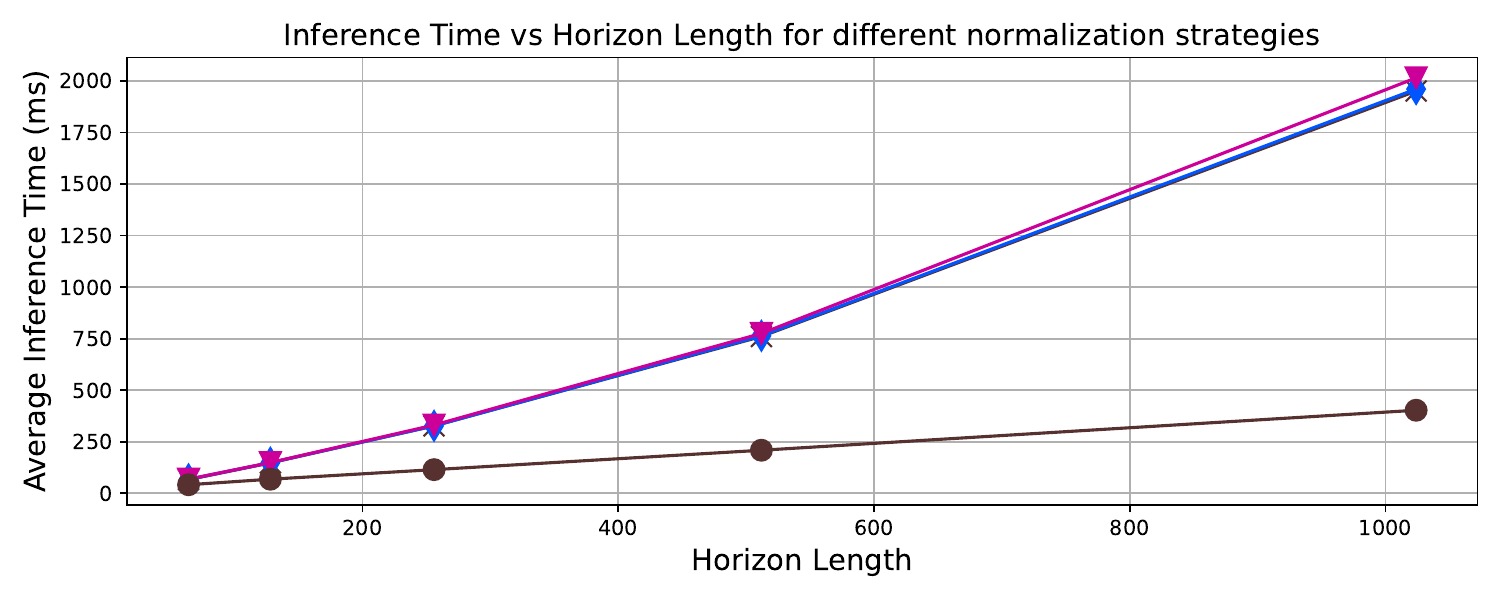}
        \caption{Context length $1024$}
    \end{subfigure}
\caption{Inference time versus horizon length $h$, comparing normalization strategies and the use of key–value caching in \causalrevinasinh{} across different context lengths. 
Similar behavior is observed without the $\sinh^{-1}$ transformation.}    
\label{fig:kvcomputation}
\end{figure}

The computational cost introduced by the different normalization strategies remains negligible relative to the overall model complexity. 
Given \(\rm N\) patches in the context, \causalrevin{} and \causalrevinasinh{} require the computation of \(\rm N\) means and variances, 
whereas \wurevin{}, \wurevinasinh{}, \revin{}, and \revinasinh{} rely on a single mean and variance estimated over the entire context. 
Importantly, \causalrevin{} and \causalrevinasinh{} enable the use of the key--value caching mechanism at inference time without additional assumptions, 
thereby avoiding the recomputation of key and value projections for past patches and significantly reducing inference latency, 
as illustrated in Figure~\ref{fig:kvcomputation}. 
Vanilla-based strategies, for each new patch appended to the context, require recomputing the normalization statistics and re-normalizing all past patches, 
which prevents the use of key--value caching. However, it is possible to consider some relaxations for using key--value caching with these strategies, 
such as editing the normalization statistics not at every step but only after a certain number of new patches, or using the statistics from the last steps for a certain number of new patches, which would introduce some approximation but could still yield significant computational savings.

\subsection{Model Architecture}\label{architecture}

We employ a Transformer architecture optimized for patch-based time-series forecasting. 
The model is trained to generate probabilistic forecasts, outputting the median alongside a set of predictive quantiles to characterize uncertainty.

The core architecture consists of stacked layers of multi-head self-attention and feed-forward networks. 
We use SwiGLU activation functions within the FFN blocks \citep{swiglu} and incorporate Rotary Positional Embeddings \citep{rope} to inject relative temporal information into the attention mechanism. 
Standard residual connections are used throughout the stack, and temporal causality is strictly enforced via a triangular attention mask.

\begin{table}[h]
\centering
\caption{Model configuration}
\label{tab:model_details}
\begin{tabular}{l|l}
\hline 
\textbf{Component} & \textbf{Value} \\
\hline
Patch size \({\rm L}\) & 32 observations \\
Max context \({\rm N}\) & 32 patch (1024 observations) \\
Forecast horizon & 1 patch (32 observations) \\
Quantiles $\mathcal{Q}$ & $\{0.1,0.2,0.3,0.4,0.5,0.6,0.7,0.8,0.9\}$ \\
Model dimension & 2048 \\
Number of layers & 6 \\
Attention heads per layer & 64 \\
Head dimension & 32 \\
Parameters & $\approx 300$M \\
\end{tabular}
\end{table}

\subsection{Training}\label{training}

\subsubsection{Training Data}\label{datasets}
Training such a model requires a large and heterogeneous corpus of time-series, enabling it to learn representations that generalize across signals rather than overfitting to individual instances. 
We therefore combine synthetic datasets, spanning diverse scales, dynamics, sampling rates, seasonalities, noise levels, and distribution shifts, with real-world data including industrial sensors, electricity load, and weather measurements.

\paragraph{Artificial Dataset.}
We synthesize approximately 1.4M time-series from basic families such as sinusoidal, linear, polynomial, and logarithmic functions, as well as their mixtures, using TSMixup \citep{chronos} to increase pattern diversity. 
In addition, we adopt the KernelSynth strategy from \citet{chronos} to sample Gaussian-process signals with mixtures of kernels (e.g., RBF, periodic, linear), sweeping hyperparameters (amplitudes, frequencies/periods, length scales, noise) to cover a broad range of dynamics.

\paragraph{Real-World Dataset.}
For real-world signals, we leverage two datasets. 
The first is the Unified time-series Dataset (UTSD) \citep{utsd}, spanning seven domains: Energy, IoT, Nature, Web, Health, Transport, and Environment. 
UTSD is organized hierarchically to provide multiple benchmark subsets, where each smaller dataset is a strict subset of a larger one, enabling scalable training and evaluation. 
It aggregates publicly available time-series curated from diverse research groups and providers. 
For training, we use UTSD-12G, which yields approximately 19M signals under our preprocessing pipeline.

The second dataset is the GIFT-Eval pretraining corpus \citep{gift}, aligned with the GIFT-Eval benchmark but constructed to avoid data leakage with respect to the benchmark train/test split, enabling a fair evaluation of large models on GIFT-Eval. 
The dataset contains approximately 71 univariate and 17 multivariate time-series datasets from various domains and frequencies. 
After preprocessing, this yields approximately 1.2M univariate time-series.

\subsubsection{Training Objective - Multi-Quantile (Pinball) Loss}

The model is trained not only to forecast the next patch but also to estimate predictive quantiles \citep{tirex, moirai2, chronos2}. 
We output quantiles
\( \mathcal{Q}=\{0.1, 0.2, 0.3, 0.4, 0.5, 0.6, 0.7, 0.8, 0.9\} \), where the median (\(q{=}0.5\)) serves as the point forecast,
and the remaining quantiles provide uncertainty bands.

During training, the model takes as input a sequence of \( \rm{N} \) patches \(\mathbf{x}_{{\rm P}_1 },\ldots,\mathbf{x}_{  {\rm P}_{\rm N}}\). For each position \(i \in \{1,\ldots,\rm{N}\}\),
it outputs quantile predictions for the next patch \(\mathbf{x}_{{\rm P}_ {i+1}}\), which is conditioned on the past patches \(\mathbf{x}_{{\rm P}_ 1},\ldots,\mathbf{x}_{{\rm P}_ i}\). 
We denote the prediction at quantile \(q\) by \(\hat{\mathbf{x}}_{{\rm P}_ {i+1}}^{(q)} \in \R^{\rm{L}}\).

Then, the training objective is the multi-quantile (pinball) loss. Aggregated over positions, 
patch elements, and quantiles, it is defined as:
\begin{equation}
\mathcal{L}
= \frac{1}{\rm N \,{\rm L}\,|\mathcal{Q}|}
\sum_{i=1}^{\rm N}\;\sum_{t=1}^{\rm{L}}\;\sum_{q \in \mathcal{Q}}
\rho_q\!\left(\mathbf{x}_{{\rm P}_ {i+1},t} - \hat{\mathbf{x}}_{{\rm P}_ {i+1},t}^{(q)}\right).
\end{equation}
where the pinball loss \(\rho_q\) is defined as:
\begin{equation}\label{rho}
\rho_q(u)=
\begin{cases}
q\,u, & u \ge 0,\\
(q-1)\,u, & u < 0.
\end{cases}
\end{equation}

\subsubsection{Setup}
Training was performed on 4 NVIDIA V100 GPUs. The implementation uses PyTorch Lightning 
with Distributed Data Parallel. We have a global batch size of 1024 and train for 225k steps in total.
We use the AdamW optimizer, a learning rate starting at \(10^{-5}\) with a linear warm-up over 10k steps until \(5.10^{-4}\), followed by cosine decay to \(10^{-5}\) until 150k steps, 
and then a constant learning rate until the 225k steps.

\subsection{Inference Loop}\label{inferenceloop}

During autoregressive inference, the models generate forecasted values patch by patch. At each step, the predicted patch is fed back into the model 
as input for the next step. This iterative process continues until the desired target horizon is reached.

When performing quantile forecasting, the situation becomes more complex. Instead of producing a single patch per step, the models output 
multiple patches corresponding to different quantiles. Since the models expect a single patch for the next step, it is not straightforward 
to feed all quantile predictions back into the model simultaneously. 

A common workaround is to feed only the median prediction back into the model at each step. While this approach preserves the 
autoregressive structure, it discards the uncertainty information captured by the other quantiles.

An alternative approach is autoregressive multi-quantile decoding, as proposed in \citet{moirai2}. This method enables consistent 
autoregressive generation while preserving the full uncertainty provided by the quantile predictions. However, it is computationally 
more expensive than the median-only approach as it requires duplicating the context for each quantile.

Considering the normalization strategies used during inference, they are handled as follows in our main experiments.

For \revin{} and \revinasinh{}, normalization statistics are computed over the entire current context. 
After appending a newly predicted patch, the context is updated and the statistics are recomputed before the next prediction.

For \wurevin{} and \wurevinasinh{}, statistics are computed over the first \(k\) patches of the context. 
When new patches are appended, the statistics remain unchanged.

For \causalrevin{} and \causalrevinasinh{}, statistics are computed in a causal manner for each patch. 
When a new patch is added to the context, its normalization statistics are computed using only preceding patches.

\subsection{Extended Discussion on Training Dynamics}\label{intuition}

\begin{figure}[h]
    \centering

    \begin{minipage}[t]{0.32\linewidth}
        \centering
        \includegraphics[width=1\linewidth]{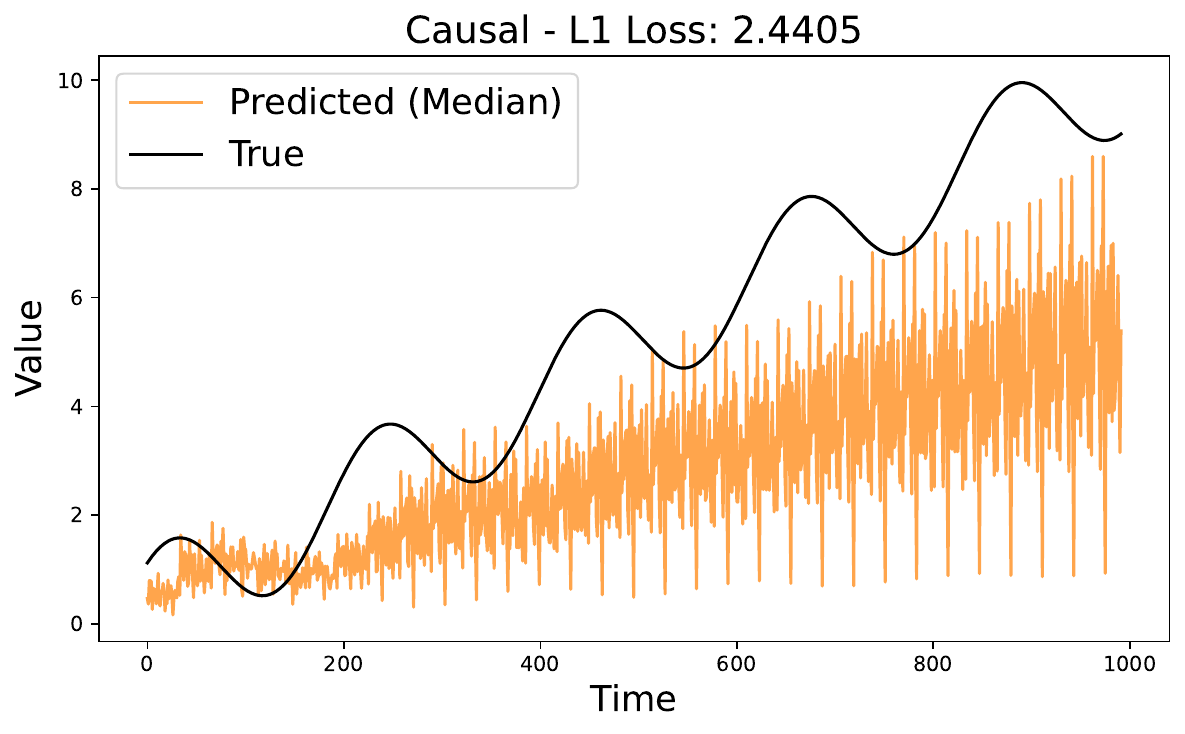}
        \small \\ (a) Causal
    \end{minipage}
    \hfill
    \begin{minipage}[t]{0.32\linewidth}
        \centering
        \includegraphics[width=1\linewidth]{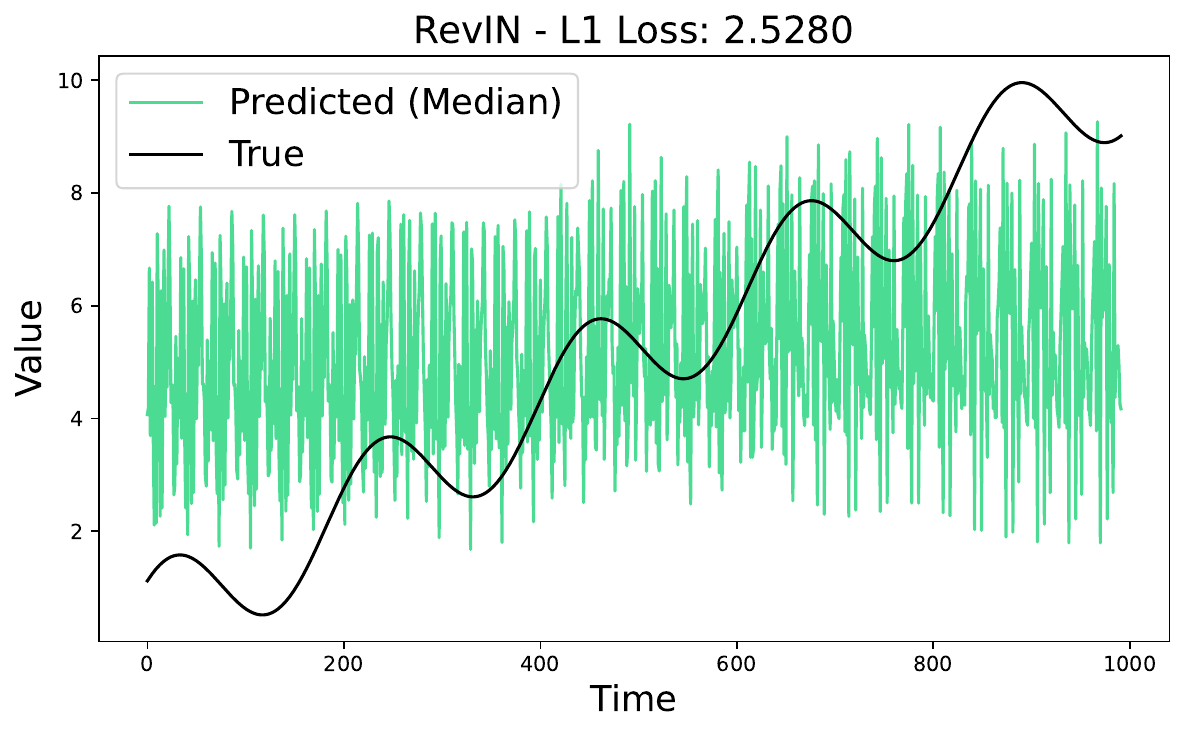}
        \small \\ (b) Vanilla
    \end{minipage}
    \hfill
    \begin{minipage}[t]{0.32\linewidth}
        \centering
        \includegraphics[width=1\linewidth]{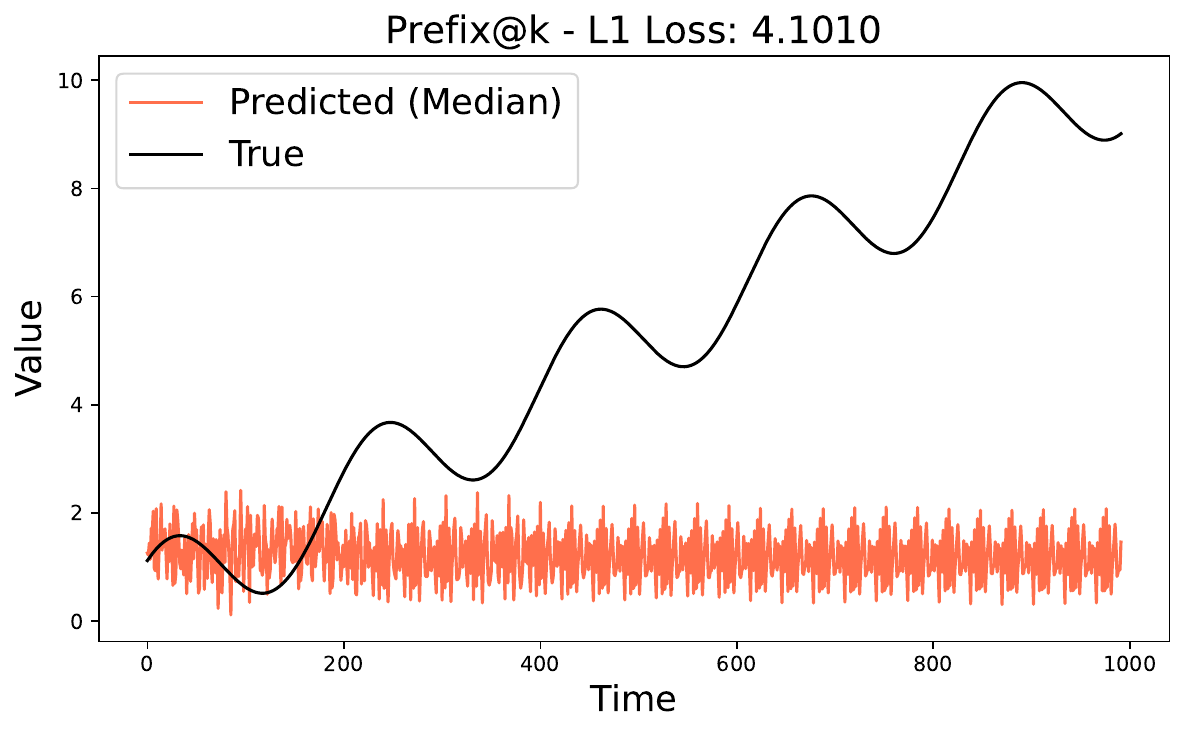}
        \small \\ (c) Prefix@$k$
    \end{minipage}
    \caption{Forecast for each normalization strategy on a synthetic non-stationary sinusoidal signal at initialization.}
    \label{fig:init_norm_comparison}
\end{figure}

Causal normalization appears to exhibit faster initial convergence, as suggested by Figure~\ref{fig:init_norm_comparison}(a). 
At initialization, the model is untrained and its parameters are randomly initialized, leading to essentially random patch predictions. 
Based on patch-specific statistics, the denormalization step in \causalrevin{} and \causalrevinasinh{} may help map these random patch predictions back to the original scale of the local data, which could explain the lower initial loss.
As training progresses, however, the model must learn to cope with the non-smoothness introduced by patch-wise normalization, which may limit its asymptotic performance.

Vanilla normalization tends to begin with a higher loss, as this strategy is not specifically adapted to local statistics (Figure~\ref{fig:init_norm_comparison}(b)).  
At initialization, the model's random patch predictions are denormalized using statistics computed over the entire context, which may not accurately reflect the local scale of each patch.
Nevertheless, this strategy appears to achieve the lowest asymptotic loss in our experiments.

Prefix-based strategies generally start with the highest initial loss.
At initialization, the model's random patch predictions are denormalized using statistics computed from a fixed prefix of the context. 
These statistics may not be representative of the local scale of subsequent patches, particularly when the signal exhibits non-stationarity or trends (Figure~\ref{fig:init_norm_comparison}(c)).
As a result, the denormalized predictions are expressed in the scale of the prefix, which can differ substantially from the scale of the local patch, potentially leading to a significant initial error.

Furthermore, prefix-based strategies appear to exhibit instability when combined with \(\sinh^{-1}\).
If the signal exhibits non-stationarity or a trend shift after the prefix window, later patches may become out-of-distribution relative to the prefix statistics.
The \(\sinh^{-1}\) transformation can compress these large values, resulting in a loss of feature resolution.
Consequently, the model may struggle to map these compressed representations back to the original scale during denormalization, which can lead to increased forecasting error.

\subsection{More results}\label{moreresults}

\subsubsection{Metrics definition}

In this section, we provide additional experimental results. But first, we clarify the computation of the MASE metric in our experimental setting.

Given a context signal $\mathbf{x} \in \mathbb{R}^{\rm{w}}$, we define the future horizon of interest as the next $h$ values, which we denote as the 
ground-truth observations as the target sequence $\mathbf{x}_{\text{tar}} 
\in \mathbb{R}^h$. The model's corresponding $q$-quantile estimate for this forecast horizon is denoted by 
$\hat{\mathbf{x}}^{(q)}_{\text{tar}} \in \mathbb{R}^{h}$.

The Mean Absolute Scaled Error (MASE) is computed as follows:
\begin{equation}\label{mase}
\text{MASE} = \frac{\frac{1}{h} \sum_{i=1}^{h} |\mathbf{x}_{\text{tar},i} - \hat{\mathbf{x}}^{(0.5)}_{\text{tar},i}|}{\frac{1}{\rm{w}-1} \sum_{j=2}^{\rm{w}} |\mathbf{x}_{j} - \mathbf{x}_{j-1}|}.
\end{equation}
The Scaled Quantile Loss (SQL) is computed as follows:
\begin{equation}\label{sql}
\text{SQL} = \frac{\frac{1}{h} \sum_{i=1}^{h} \sum_{q \in \mathcal{Q}} \rho_q(\mathbf{x}_{\text{tar},i} -  \hat{\mathbf{x}}^{(q)}_{\text{tar}, i} ) }{\frac{1}{\rm{w}-1} \sum_{j=2}^{\rm{w}} |\mathbf{x}_{j} - \mathbf{x}_{j-1}|}.
\end{equation}
With \(\rho_q\) defined in Eq.\ref{rho}.

Since the scaling factors in the denominators are based on one-step differences within the context, whereas the numerators aggregate error over a potentially long forecast horizon, 
MASE and SQL naturally tend to exceed one in most cases, even when the model exhibits strong predictive performance.

\subsubsection{Skill scores}

We complement the critical difference diagrams presented in the main paper with pairwise skill scores~\citep{fevbench}. 
The skill score of a given strategy \(s\) with respect to a reference strategy \(r\) is defined as
\begin{equation}
\text{SkillScore}(s, r) = 1 - \sqrt[D]{\prod_{d=1}^D \text{clip}\bigg( \frac{\text{Err}_{s, d}}{\text{Err}_{r, d}}, \, l, \, u \bigg) },
\end{equation}
where \(\text{Err}_{s, d}\) denotes the error of strategy \(s\) on dataset \(d\), \(D\) is the total number of datasets, and \(\text{clip}(\cdot, l, u)\) constrains the ratio to the interval \([l, u]\) in order to prevent extreme values from disproportionately influencing the aggregated score.
Following~\citet{fevbench}, we set \(l = 0.01\) and \(u = 100\).

The skill score takes values in \((-\infty, 1]\): a score of \(1\) corresponds to perfect performance (i.e., zero error), a score of \(0\) indicates performance equivalent to the reference strategy, and negative values indicate worse performance than the reference.
Thus, \(\text{SkillScore}(s, r)\) quantifies the average relative error reduction achieved by strategy \(s\) compared to the reference strategy \(r\) across datasets.

\begin{figure}[h]
    \centering
    \includegraphics[width=\linewidth]{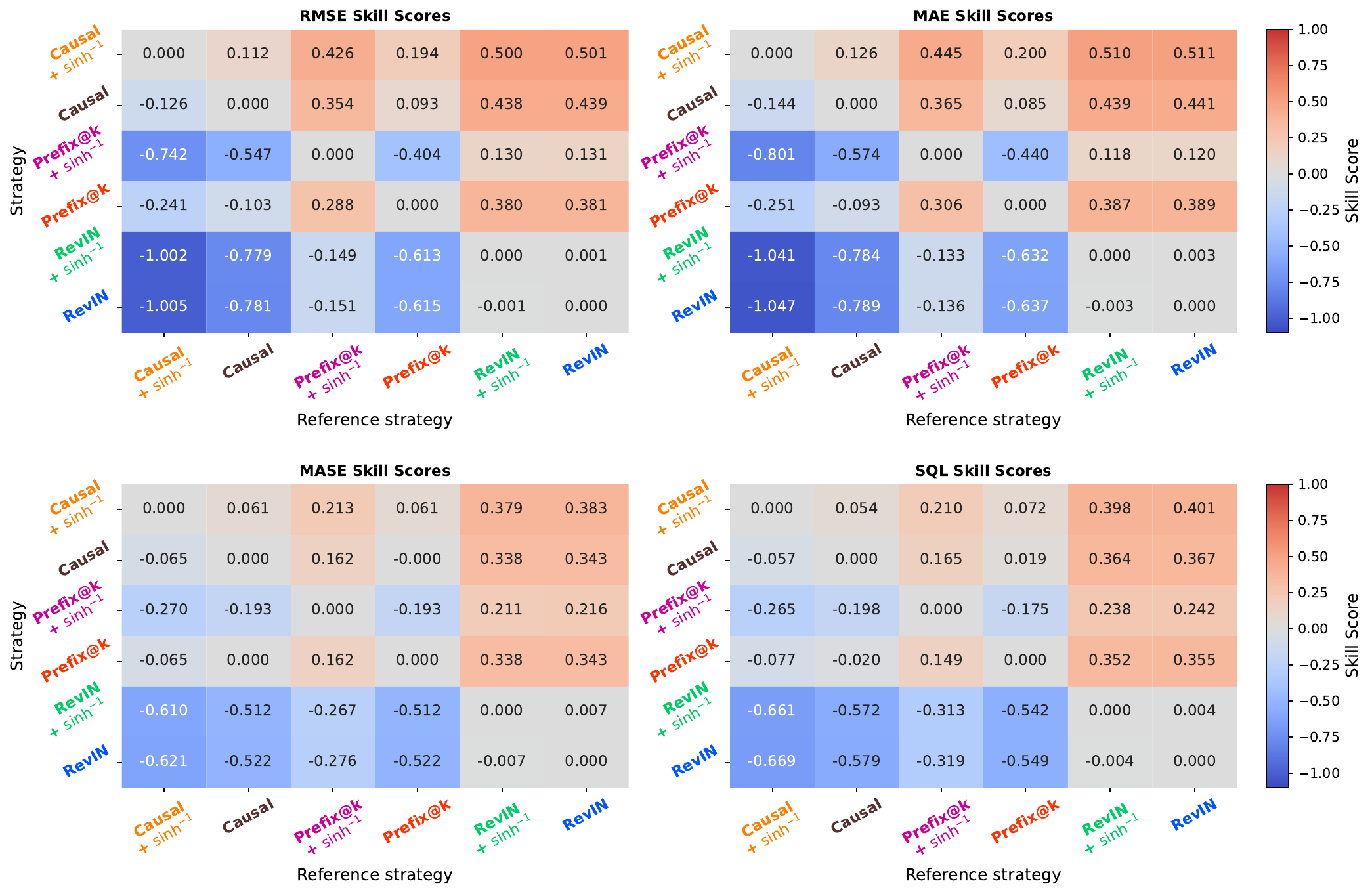}
    \caption{Pairwise skill scores for each normalization strategy, aggregated over context lengths \((128, 256, 512)\), datasets, and forecasting horizons \((32, 64, 96, \dots, 512)\).}
    \label{fig:skillscores}
\end{figure}

As shown in Figure~\ref{fig:skillscores} and consistent with the conclusions drawn from Table~\ref{tab:results}, \causalrevinasinh{} achieves the best overall performances across all metrics.
When considering scaled metrics (MASE and SQL), \causalrevinasinh{} still outperforms all other strategies, however, the gap with \wurevin{} and \causalrevin{} is reduced, which is consistent with the critical difference diagram results for these metrics in Table~\ref{tab:results} when considering \wurevin{}.
While \revin{} and \revinasinh{} previously exhibited intermediate performance, they now rank among the lowest-performing strategies under this aggregated skill score analysis.

\subsubsection{Context lengths 64 and 1024}

Secondly, we provide additional results in Table~\ref{tab:results_641024}, again using the critical difference diagram methodology presented in Table~\ref{tab:results}, but focusing specifically on context lengths of \(64\) and \(1024\).

During training, the maximum context length is set to 32 patches (=1024 observations). At evaluation time, when using a context length of 1024 observations and forecasting beyond the next 32 values, we do not truncate the signal when appending predictions to the existing context.
Consequently, the effective context length increases with the prediction horizon and may exceed 1024.
This experimental setting enables us to analyze model behavior when the effective context length surpasses the maximum sequence length encountered during training.

\begin{table}[h]
\centering
\mytabsize
\setlength{\tabcolsep}{3pt} 
\caption{Critical difference diagram \citep{cddiag} results across context sizes \(\rm{w} \in \{64, 1024\}\) for each normalization strategy aggregated by datasets and forecasting horizons. The best overall mean rank for each metric and context size is shown in \textbf{bold}, and the second-best is \underline{underlined} (lower ranks indicate better performance). 
Strategies connected within the same group (\(\text{rank}^{\text a}\)) are not statistically distinguishable according to the Nemenyi test with $\alpha=0.05$. Best group is a, second-best group is b, and so on. 
Results are aggregated over all test signals and forecasting horizons.}
\label{tab:results_641024}
\begin{tabular}{@{}c|cc|cc|cc|cc|cc|cc@{}}
\midrule
     & \multicolumn{2}{c|}{\wurevin}
     & \multicolumn{2}{c|}{\wurevinasinh}
     & \multicolumn{2}{c|}{\revin}
     & \multicolumn{2}{c|}{\revinasinh}
     & \multicolumn{2}{c|}{\causalrevin}
     & \multicolumn{2}{c}{\causalrevinasinh} \\
     \midrule
    \(\rm{w}\) & 64 & 1024
     & 64 & 1024
     & 64 & 1024
     & 64 & 1024
     & 64 & 1024
     & 64 & 1024 \\
\midrule

MAE  & 3.0\textsuperscript{\tiny  ab}    & 3.6\textsuperscript{b}  & 4.2\textsuperscript{\tiny  c} & 5.7\textsuperscript{\tiny  c}& \textbf{2.9}\textsuperscript{\tiny  a}   & \underline{3.3}\textsuperscript{\tiny  b} & \underline{2.9}\textsuperscript{\tiny  a}     & 3.5\textsuperscript{\tiny  b}   & 4.0\textsuperscript{\tiny  bc} &  3.7\textsuperscript{\tiny  b} & 4.0\textsuperscript{bc}    & \textbf{1.4}\textsuperscript{\tiny  a}     \\
RMSE  & \textbf{2.3}\textsuperscript{\tiny  a} & \underline{2.8}\textsuperscript{\tiny  b}  & 4.4\textsuperscript{\tiny  c} & 5.1\textsuperscript{\tiny  d}    & 3.4\textsuperscript{\tiny  abc}   & 4.4\textsuperscript{\tiny  d}    & \underline{3.1}\textsuperscript{\tiny  ab}     & 4.2\textsuperscript{\tiny  cd}   &  3.8\textsuperscript{bc} & 3.2\textsuperscript{\tiny  bc}   & 4.1\textsuperscript{\tiny  bc}    & \textbf{1.3}\textsuperscript{\tiny  a}      \\
MASE & \textbf{2.1}\textsuperscript{\tiny  a} & 3.2\textsuperscript{\tiny  b} & 3.7\textsuperscript{\tiny  bc} & 5.9\textsuperscript{\tiny  c}    & \underline{3.0}\textsuperscript{\tiny  ab}   & \underline{2.8}\textsuperscript{\tiny  ab}   & 4.0\textsuperscript{\tiny  bc}     & 3.8\textsuperscript{\tiny  b}  & 4.6\textsuperscript{\tiny  c}  & 3.4\textsuperscript{\tiny  b} & 3.5\textsuperscript{\tiny  b}    & \textbf{1.9}\textsuperscript{\tiny  a}     \\
SQL & \underline{3.0}\textsuperscript{\tiny  a} & 3.3\textsuperscript{\tiny  bc}  & \textbf{2.9}\textsuperscript{\tiny  a} & 5.5\textsuperscript{\tiny  d}    & 3.3\textsuperscript{\tiny  a}   & \underline{3.0}\textsuperscript{\tiny  b}  & 4.5\textsuperscript{\tiny  b}     & 4.2\textsuperscript{\tiny  c}  & 4.1\textsuperscript{\tiny  ab}  & 3.1\textsuperscript{\tiny  bc}  & 3.2\textsuperscript{\tiny  a}    & \textbf{1.9}\textsuperscript{\tiny  a}     \\
\midrule
\end{tabular}
\end{table}

For a context length of \(64\), \wurevin{} appears to be the most effective strategy, followed by \revinasinh{}. In contrast, the causal-based strategies exhibit comparatively weaker performance in this setting. 
However, when the context length increases to \(1024\), \causalrevinasinh{} becomes the top-performing method, with \wurevin{} ranking second by a substantial margin.
This result confirms that the effectiveness of normalization strategies depends on the context length. Specifically, prefix-based methods perform well for short context lengths, whereas causal normalization strategies achieve superior performance for longer contexts but exhibit degraded performance when the context length is limited.

\subsubsection{Forecasting performance across horizons}

Finally, we report complementary comparisons by plotting forecasting performance across multiple prediction horizons, context lengths, and evaluation metrics.

\begin{figure}[h]
    \centering
    \begin{subfigure}[t]{0.32\linewidth}
        \centering
        \includegraphics[width=\linewidth]{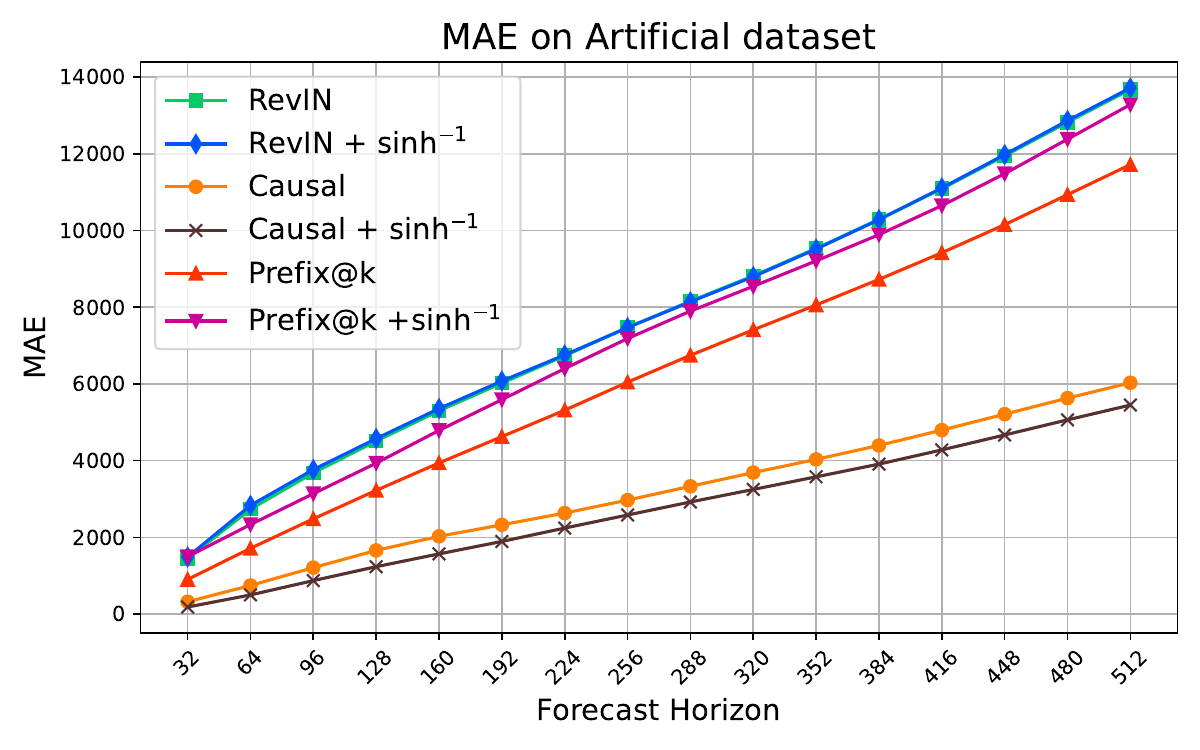}
        \caption{MAE 128}
        \label{fig:horizon_mae_128_artificial}
    \end{subfigure}
    \begin{subfigure}[t]{0.32\linewidth}
        \centering
        \includegraphics[width=\linewidth]{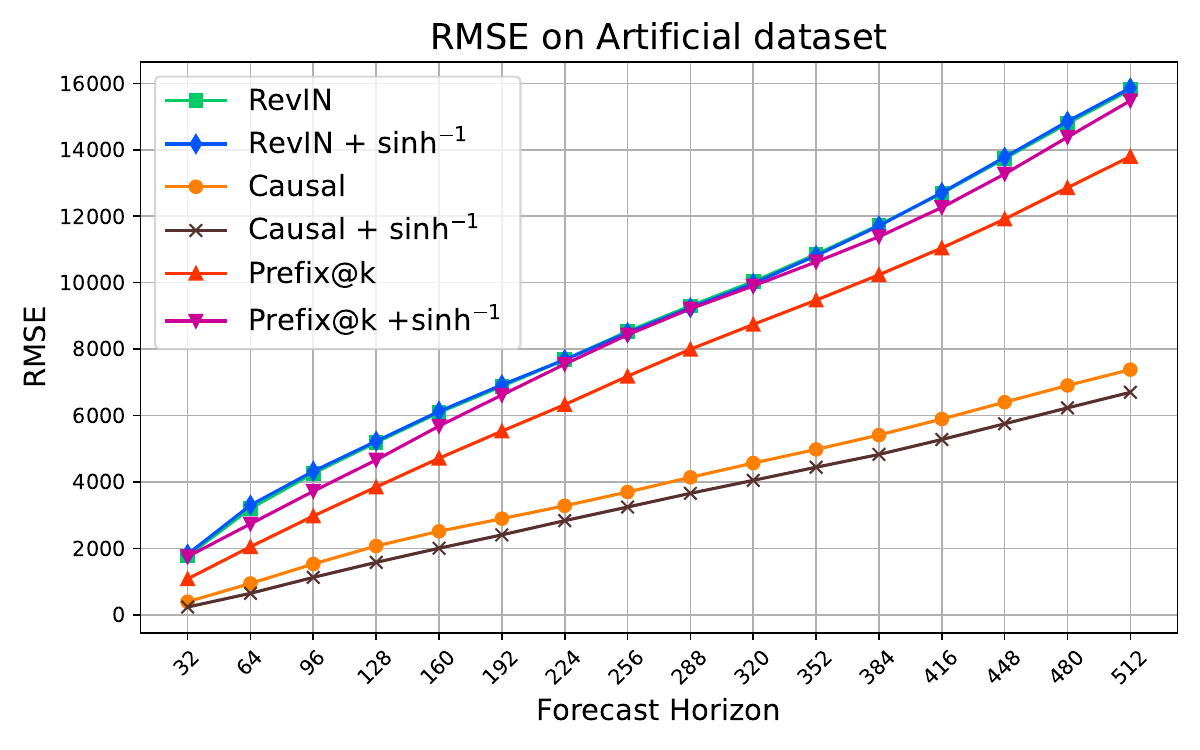}
        \caption{RMSE 128}
        \label{fig:horizon_rmse_128_artificial}
    \end{subfigure}
    \begin{subfigure}[t]{0.32\linewidth}
        \centering
        \includegraphics[width=\linewidth]{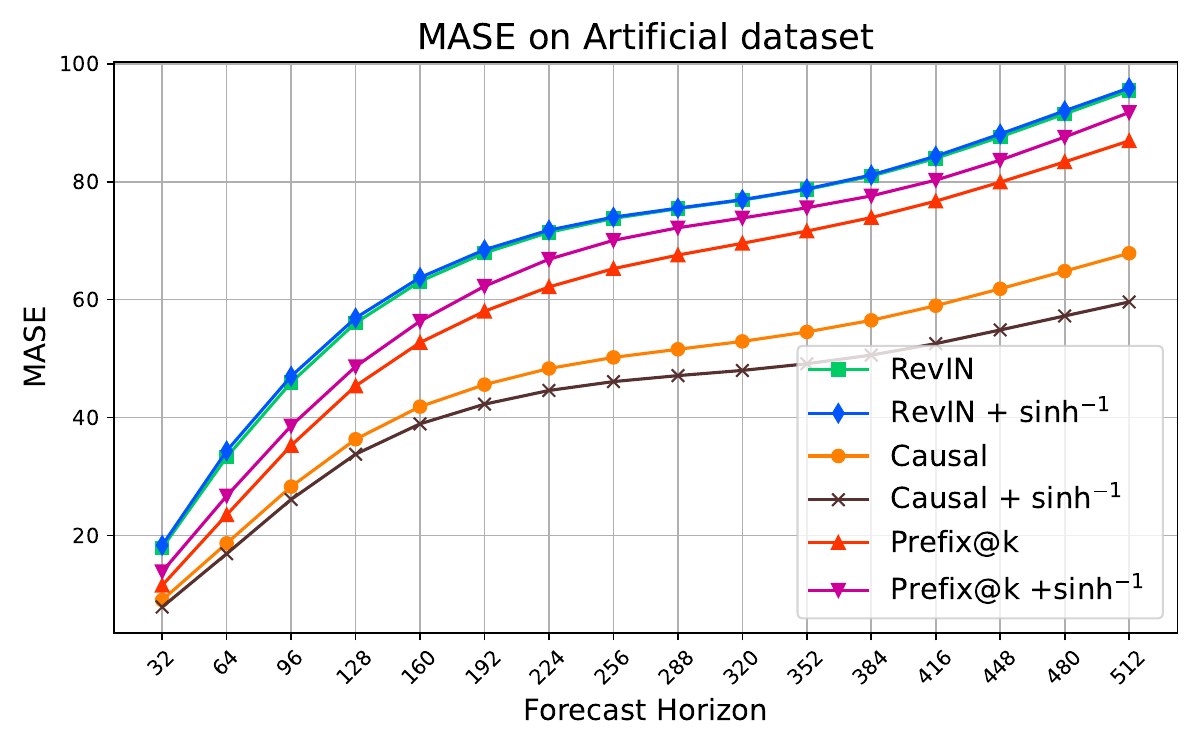}
        \caption{MASE 128}
        \label{fig:horizon_mase_128_artificial}
    \end{subfigure}

    \begin{subfigure}[t]{0.32\linewidth}
        \centering
        \includegraphics[width=\linewidth]{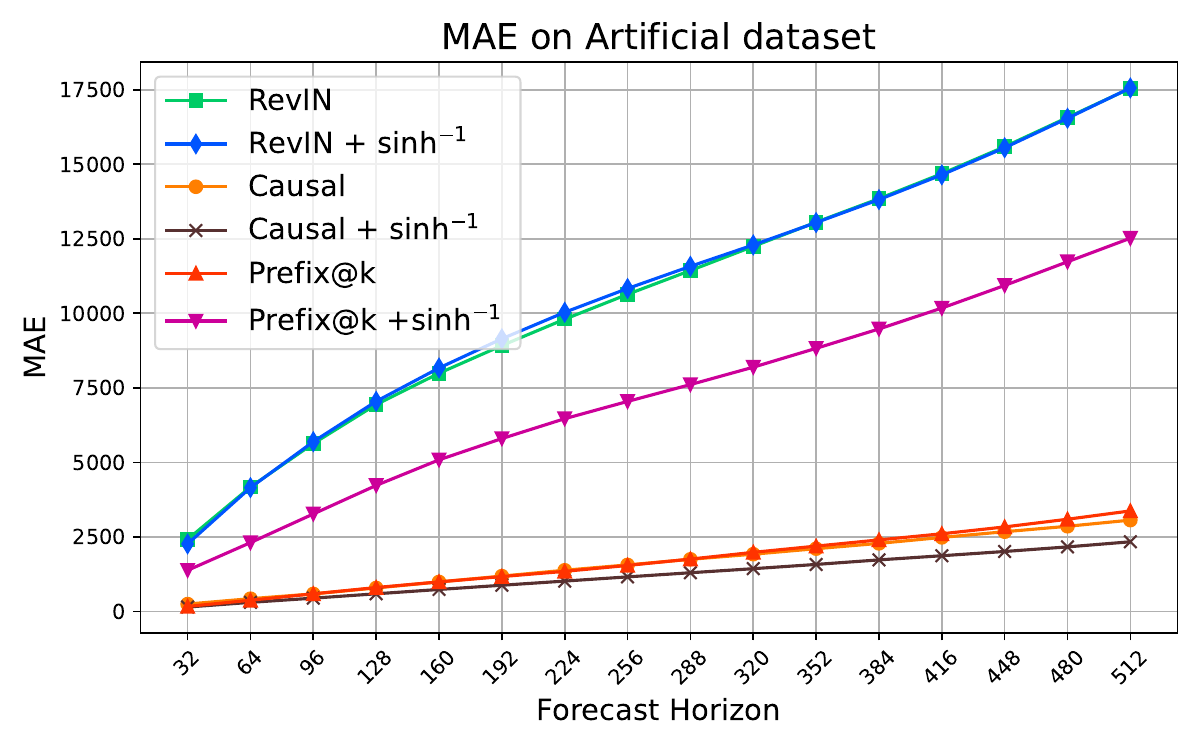}
        \caption{MAE 256}
        \label{fig:horizon_mae_256_artificial}
    \end{subfigure}
    \begin{subfigure}[t]{0.32\linewidth}
        \centering
        \includegraphics[width=\linewidth]{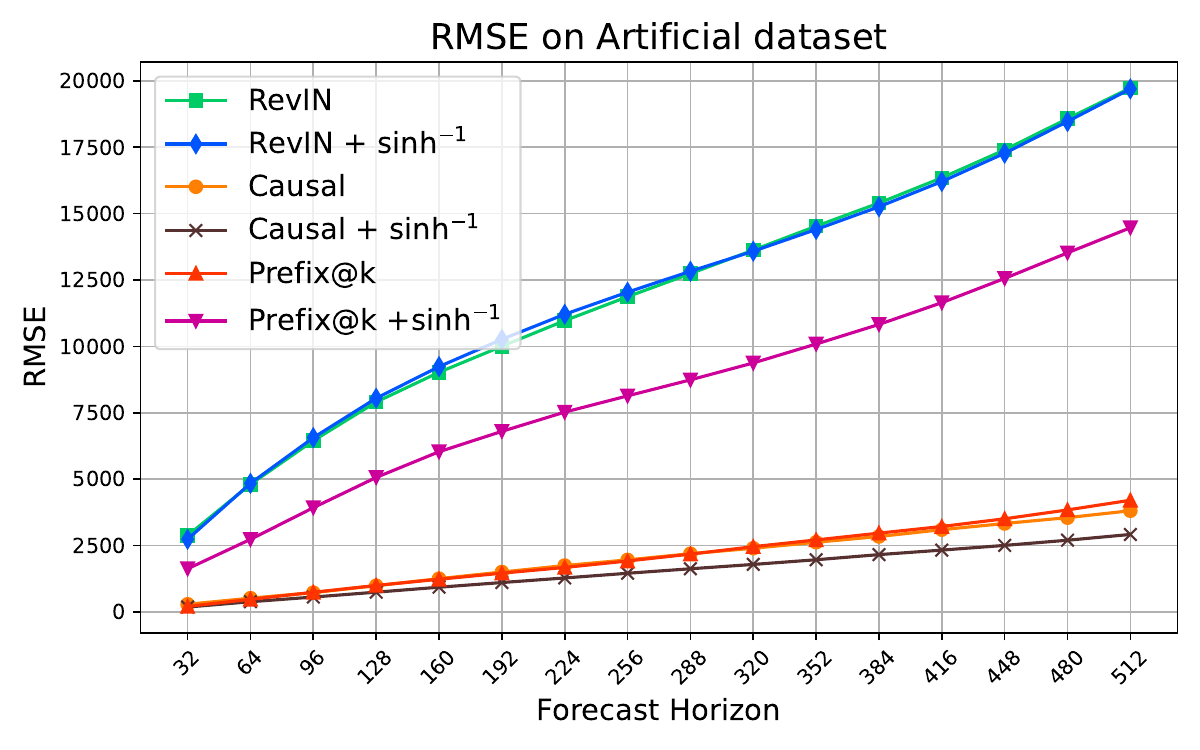}
        \caption{RMSE 256}
        \label{fig:horizon_rmse_256_artificial}
    \end{subfigure}
    \begin{subfigure}[t]{0.32\linewidth}
        \centering
        \includegraphics[width=\linewidth]{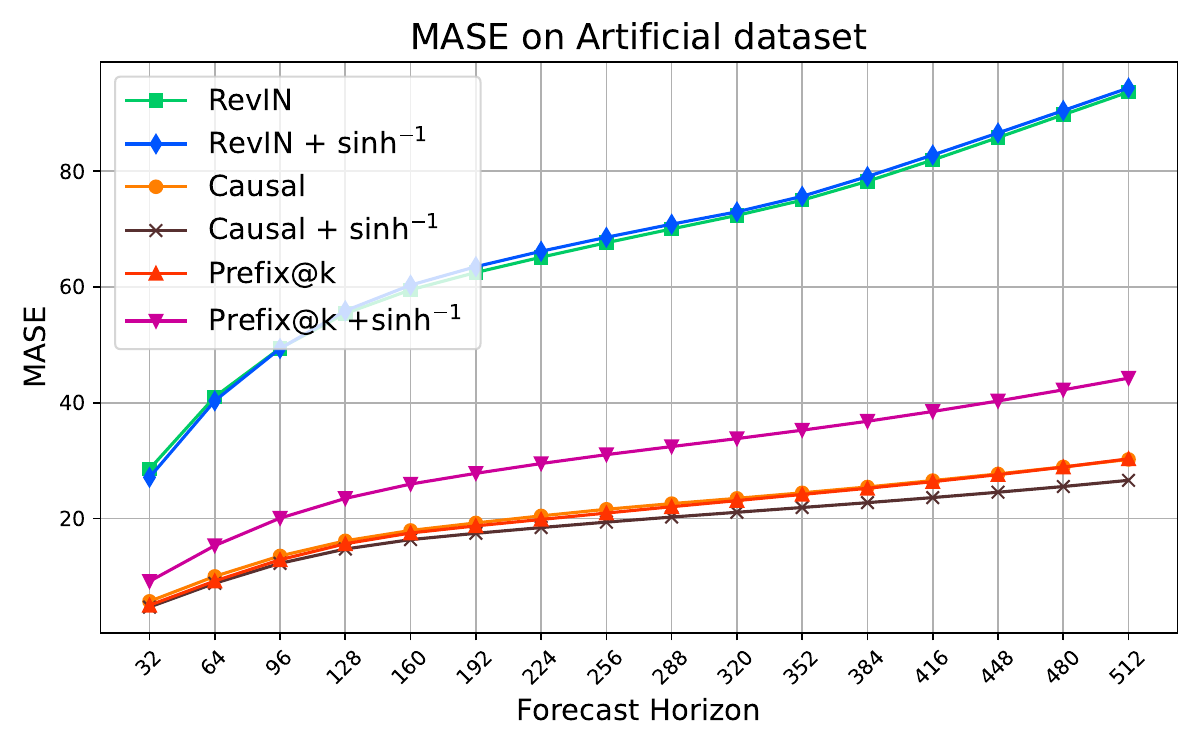}
        \caption{MASE 256}
        \label{fig:horizon_mase_256_artificial}
    \end{subfigure}

    \begin{subfigure}[t]{0.32\linewidth}
        \centering
        \includegraphics[width=\linewidth]{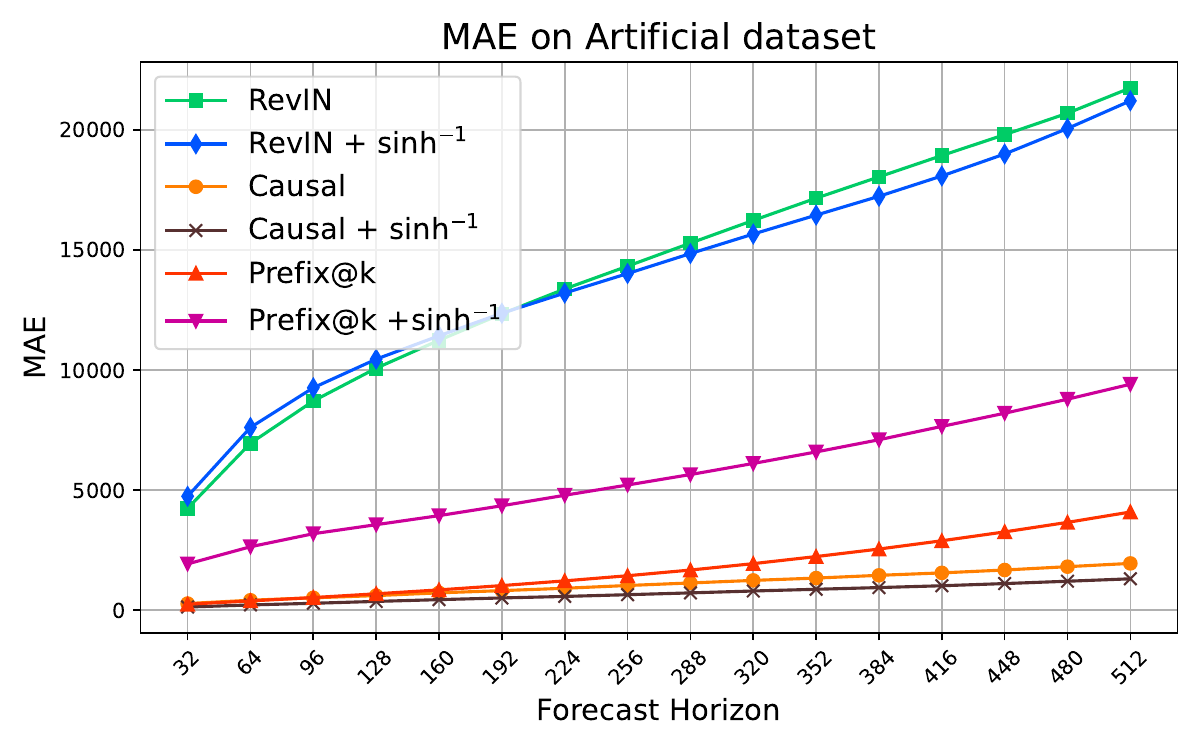}
        \caption{MAE 512}
        \label{fig:horizon_mae_512_artificial}
    \end{subfigure}
    \begin{subfigure}[t]{0.32\linewidth}
        \centering
        \includegraphics[width=\linewidth]{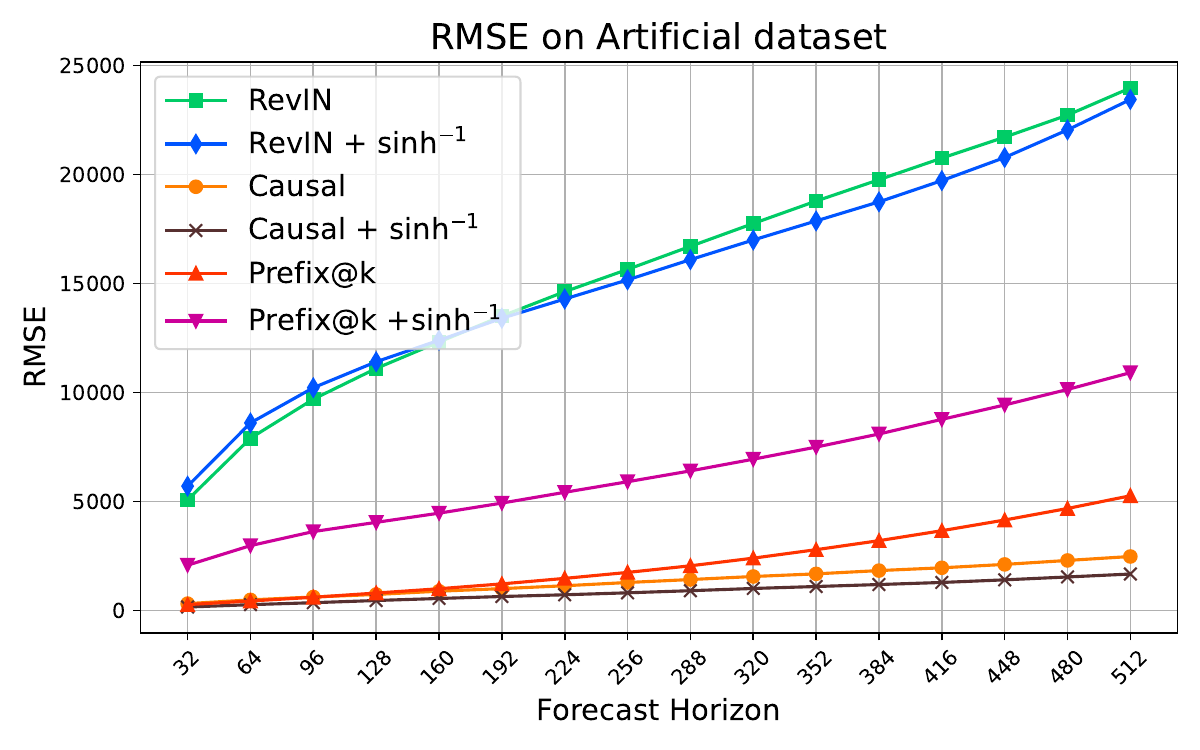}
        \caption{RMSE 512}
        \label{fig:horizon_rmse_512_artificial}
    \end{subfigure}
    \begin{subfigure}[t]{0.32\linewidth}
        \centering
        \includegraphics[width=\linewidth]{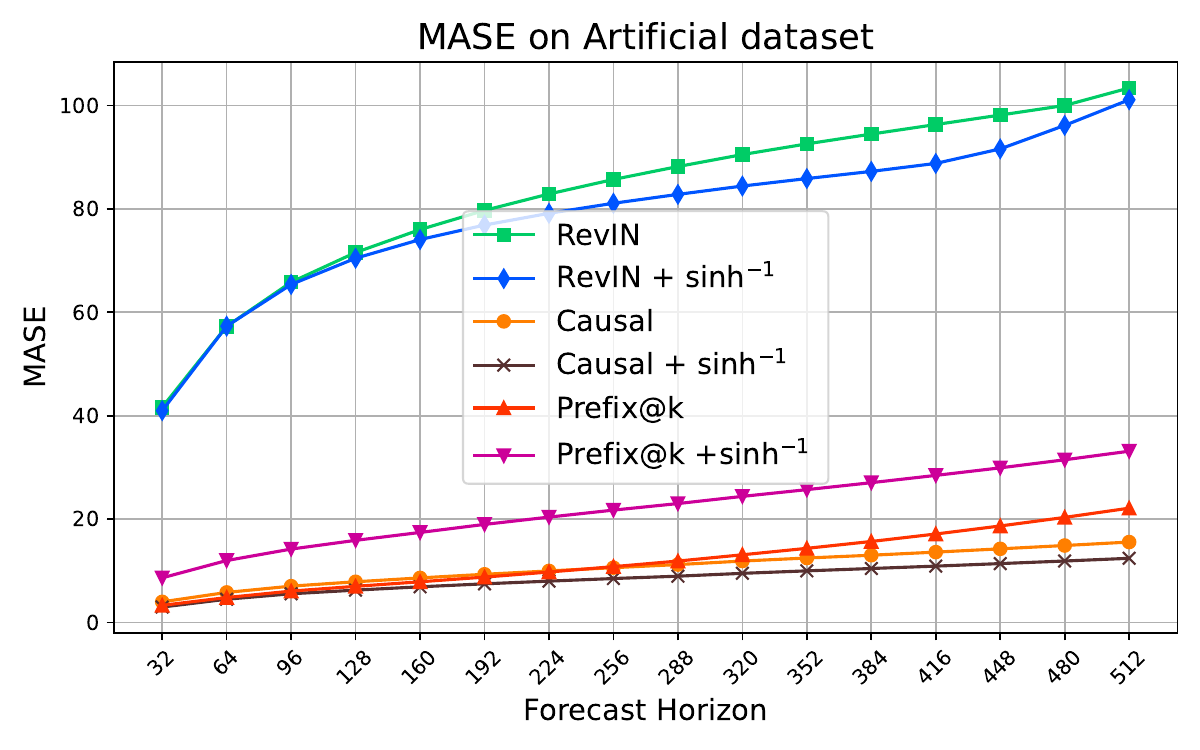}
        \caption{MASE 512}
        \label{fig:horizon_mase_512_artificial}
    \end{subfigure}

    \begin{subfigure}[t]{0.32\linewidth}
        \centering
        \includegraphics[width=\linewidth]{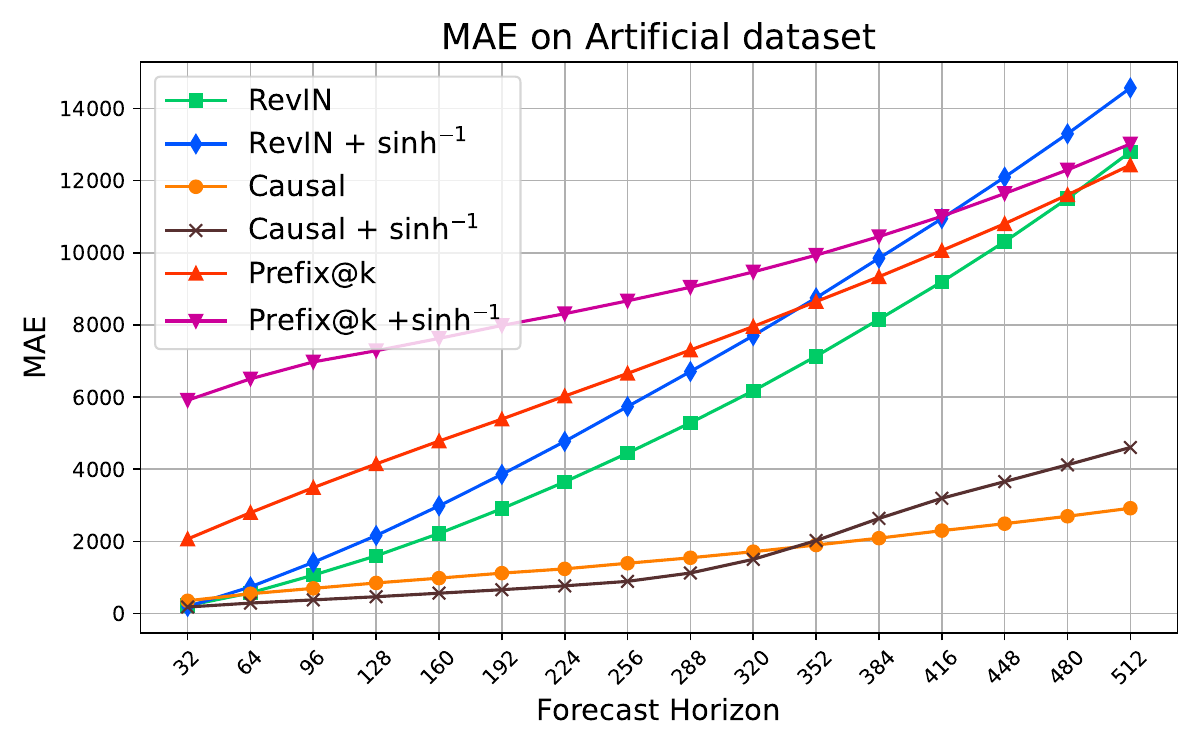}
        \caption{MAE 1024}
        \label{fig:horizon_mae_1024_artificial}
    \end{subfigure}
    \begin{subfigure}[t]{0.32\linewidth}
        \centering
        \includegraphics[width=\linewidth]{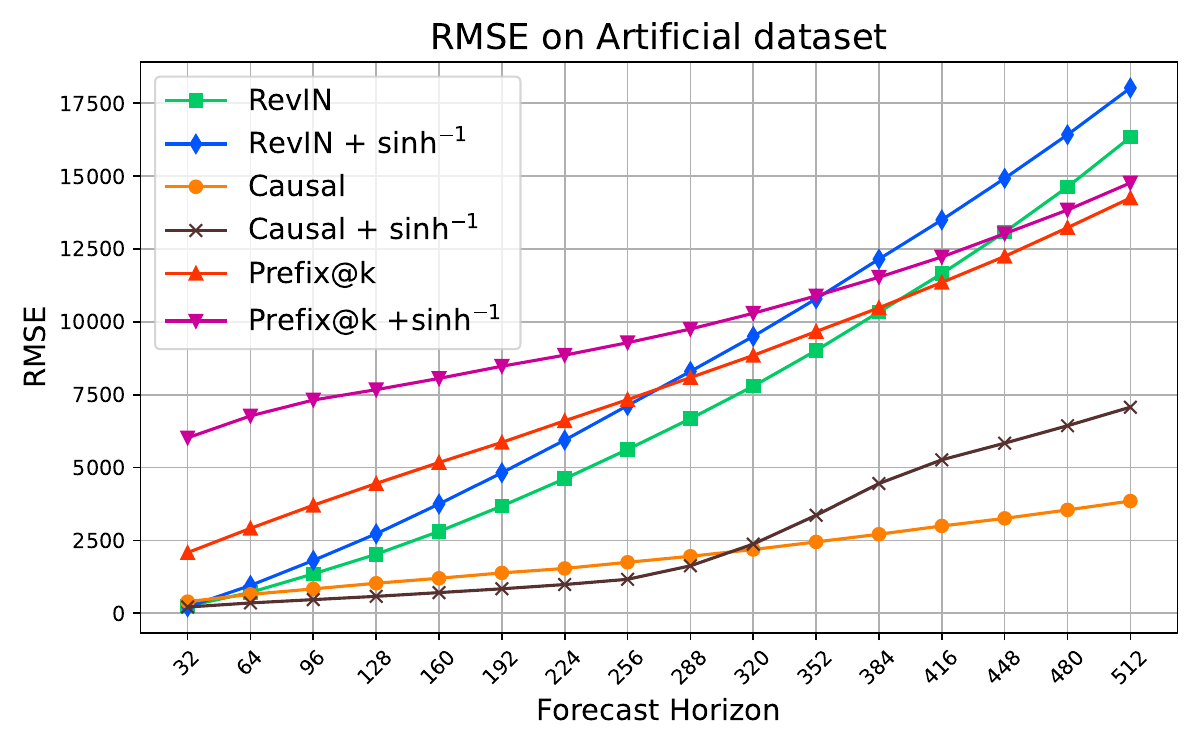}
        \caption{RMSE 1024}
        \label{fig:horizon_rmse_1024_artificial}
    \end{subfigure}
    \begin{subfigure}[t]{0.32\linewidth}
        \centering
        \includegraphics[width=\linewidth]{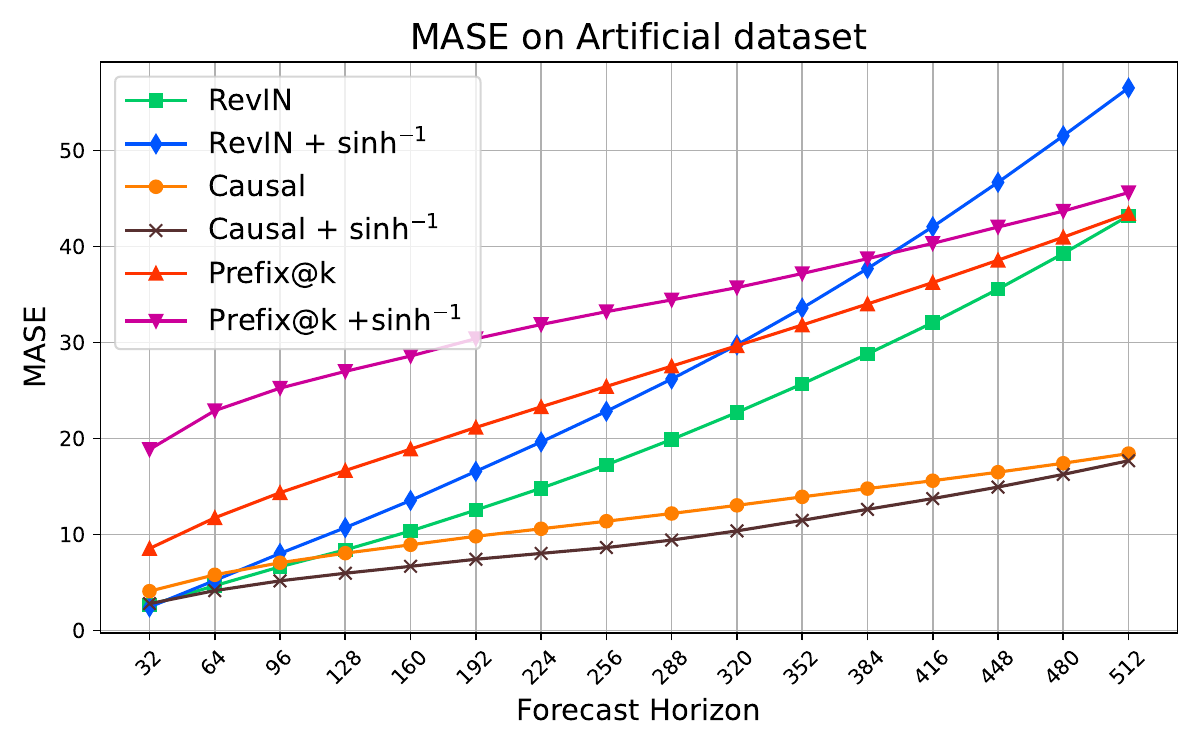}
        \caption{MASE 1024}
        \label{fig:horizon_mase_1024_artificial}
    \end{subfigure}

    \caption{Forecasting performance across prediction horizons on the Synthetic dataset for four context lengths (128, 256, 512, 1024).}
    \label{fig:horizon_synthetic_comparison}
\end{figure}

On the synthetic benchmark (Figure~\ref{fig:horizon_synthetic_comparison}), \causalrevinasinh{} and \causalrevin{} achieve the strongest overall performances across all context lengths and evaluation metrics.
For context lengths of 256 and 512, \wurevin{} performs competitively with \causalrevinasinh{} and \causalrevin{}.
For context lengths below 1024, \revin{} and \revinasinh{} yield the weakest performance across all metrics.
However, when the effective context length exceeds 1024 during inference, \revin{} and \revinasinh{} achieve performance comparable to \wurevin{} and \wurevinasinh{}.

\begin{figure}[h]
    \centering
    \begin{subfigure}[t]{0.32\linewidth}
        \centering
        \includegraphics[width=\linewidth]{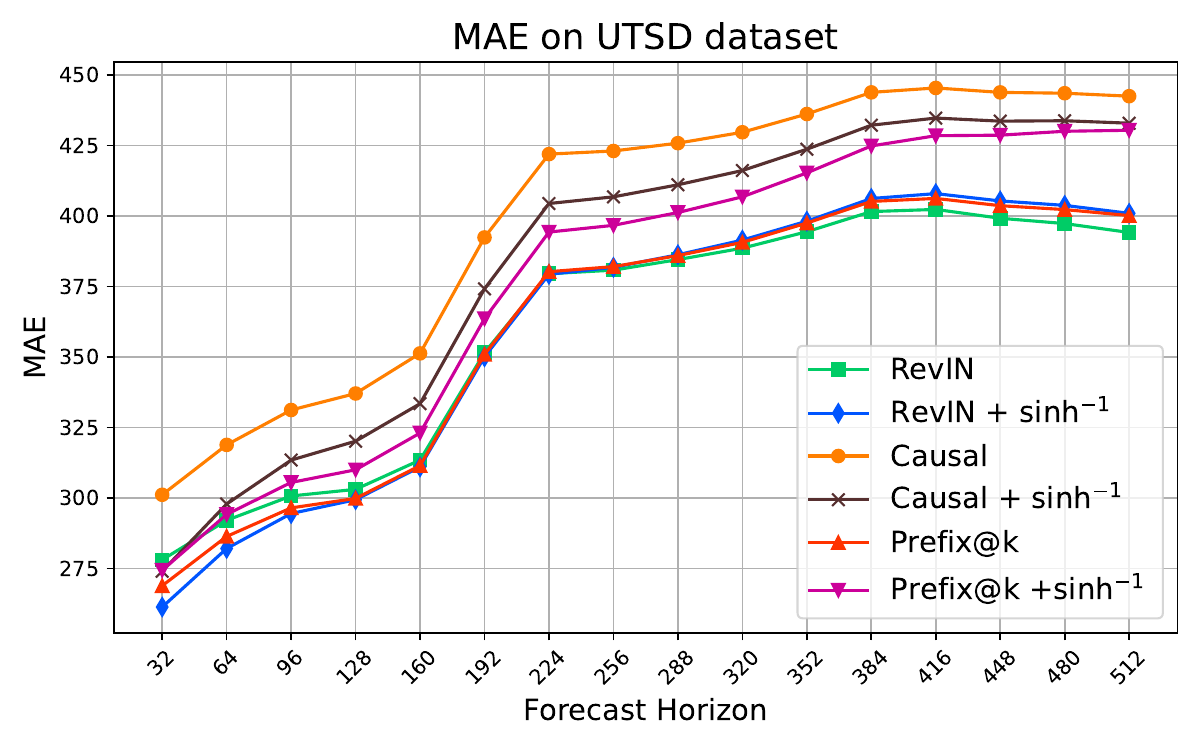}
        \caption{MAE 128}
        \label{fig:horizon_mae_128_utsd}
    \end{subfigure}
    \begin{subfigure}[t]{0.32\linewidth}
        \centering
        \includegraphics[width=\linewidth]{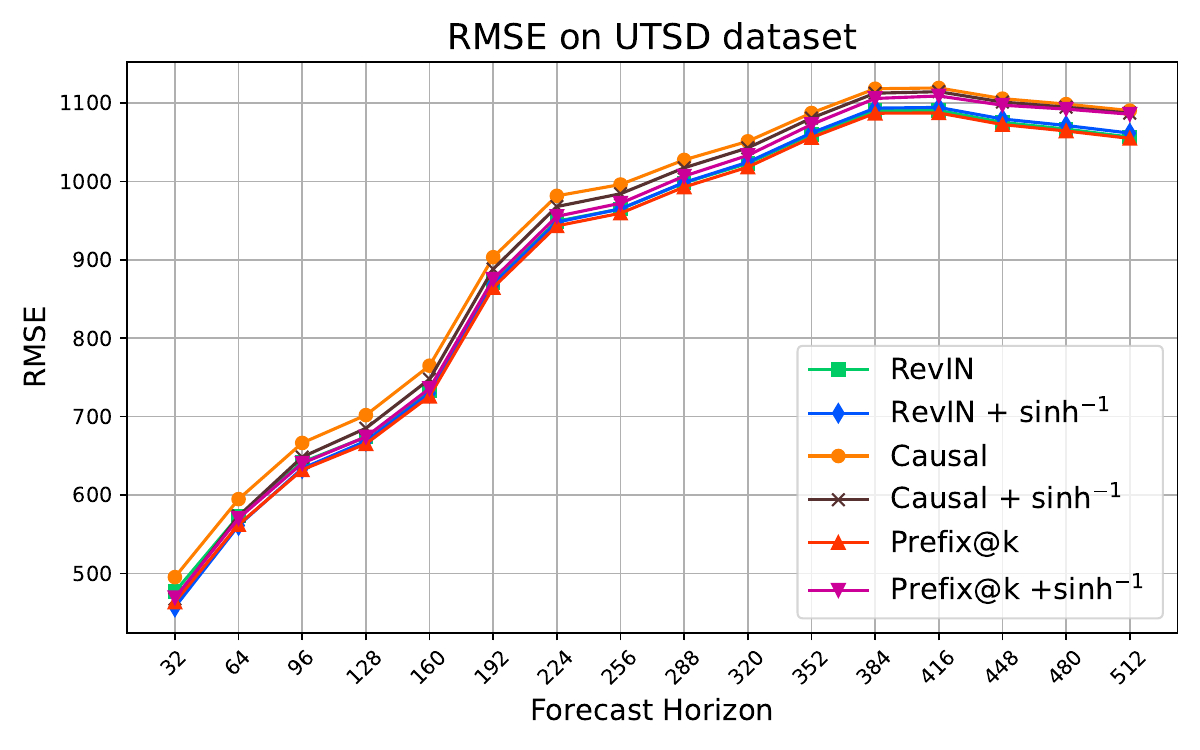}
        \caption{RMSE 128}
        \label{fig:horizon_rmse_128_utsd}
    \end{subfigure}
    \begin{subfigure}[t]{0.32\linewidth}
        \centering
        \includegraphics[width=\linewidth]{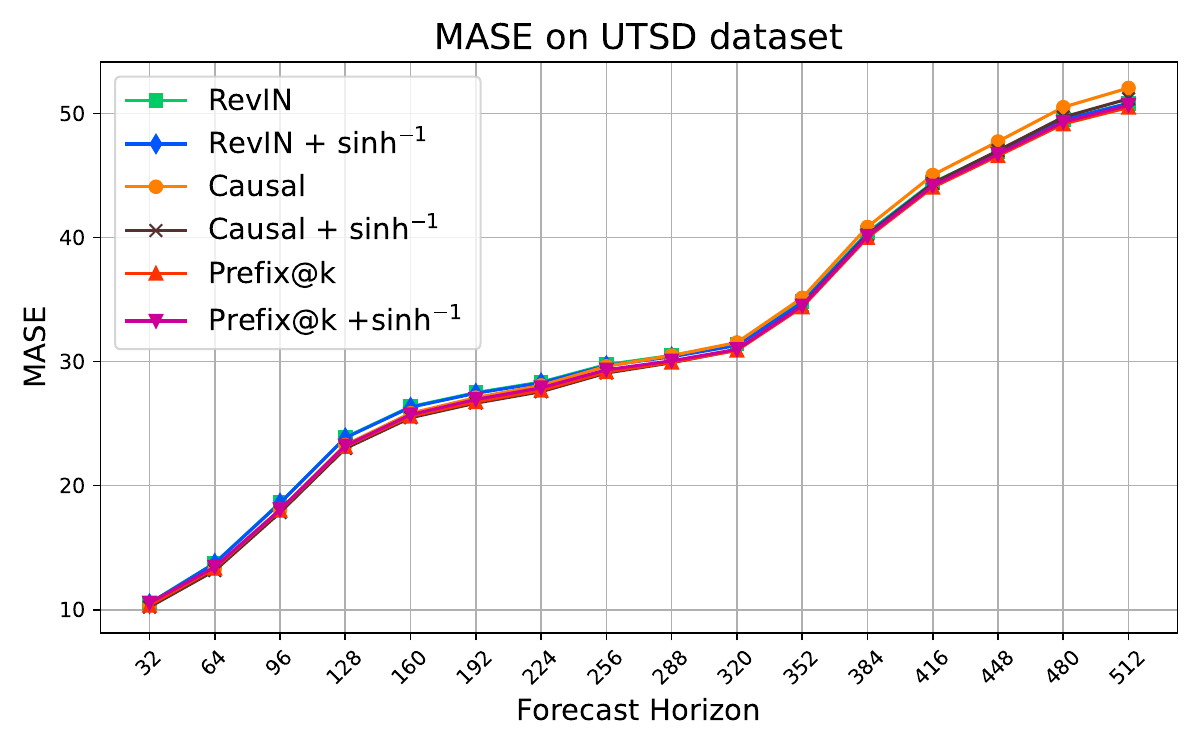}
        \caption{MASE 128}
        \label{fig:horizon_mase_128_utsd}
    \end{subfigure}

    \begin{subfigure}[t]{0.32\linewidth}
        \centering
        \includegraphics[width=\linewidth]{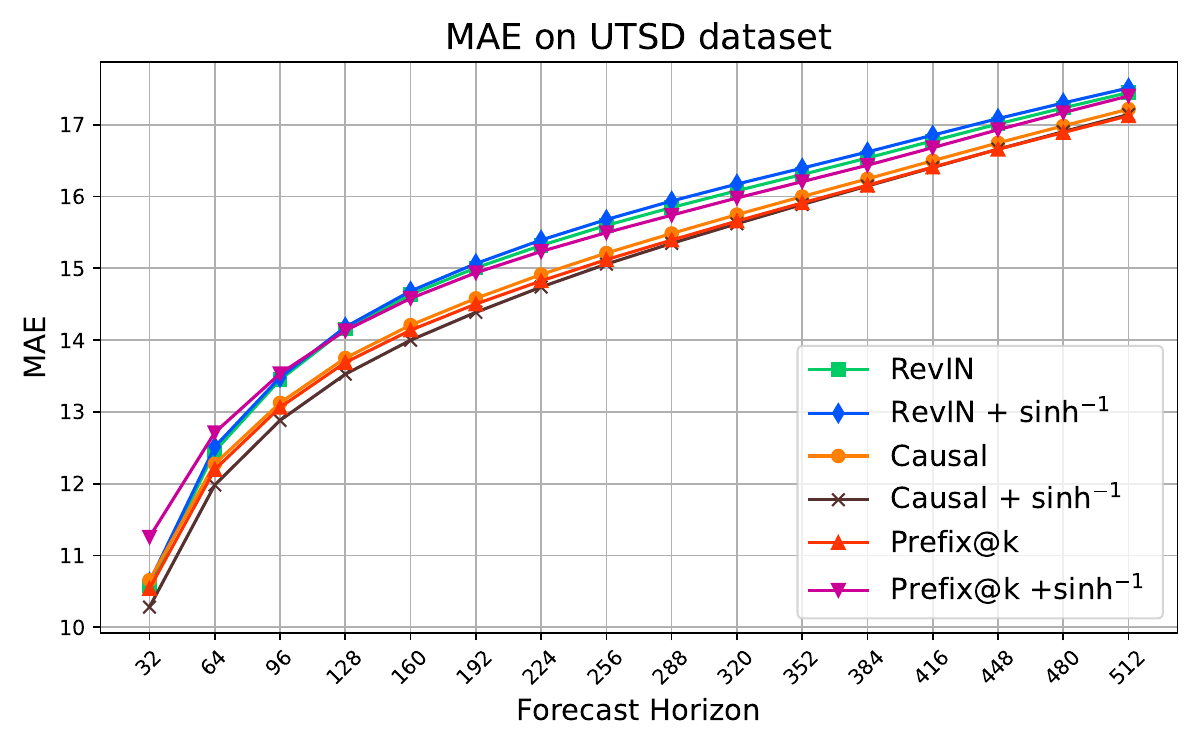}
        \caption{MAE 256}
        \label{fig:horizon_mae_256_utsd}
    \end{subfigure}
    \begin{subfigure}[t]{0.32\linewidth}
        \centering
        \includegraphics[width=\linewidth]{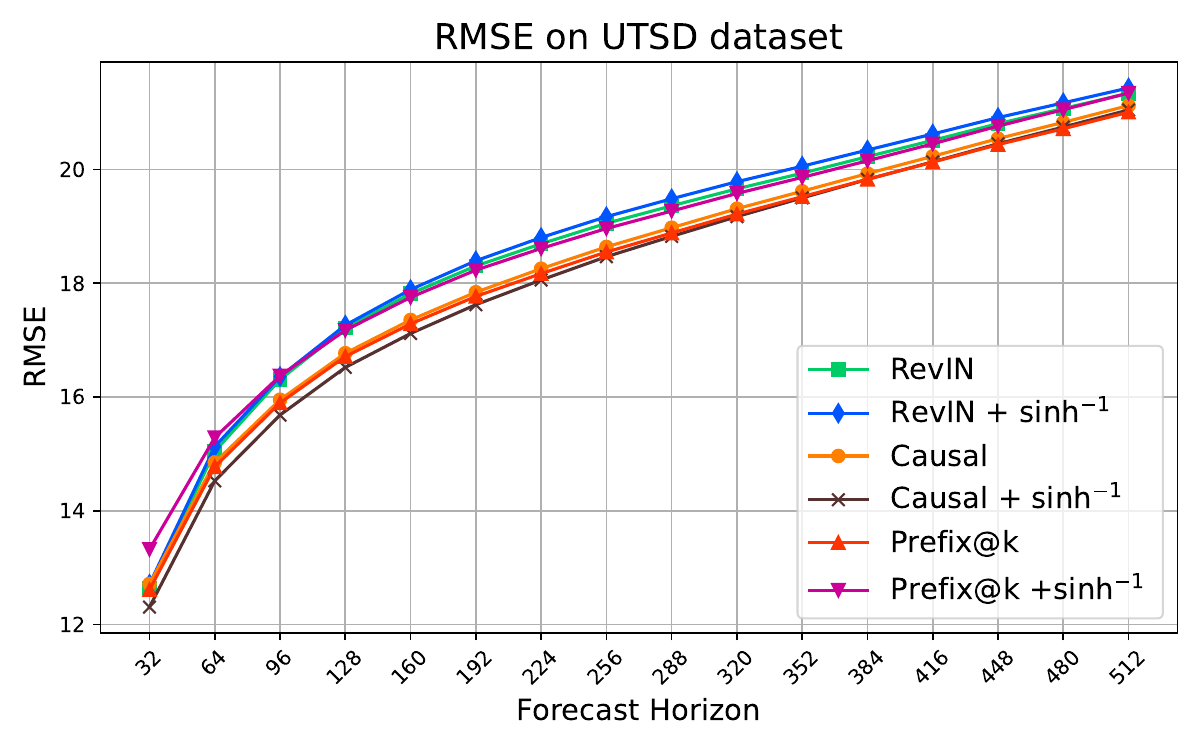}
        \caption{RMSE 256}
        \label{fig:horizon_rmse_256_utsd}
    \end{subfigure}
    \begin{subfigure}[t]{0.32\linewidth}
        \centering
        \includegraphics[width=\linewidth]{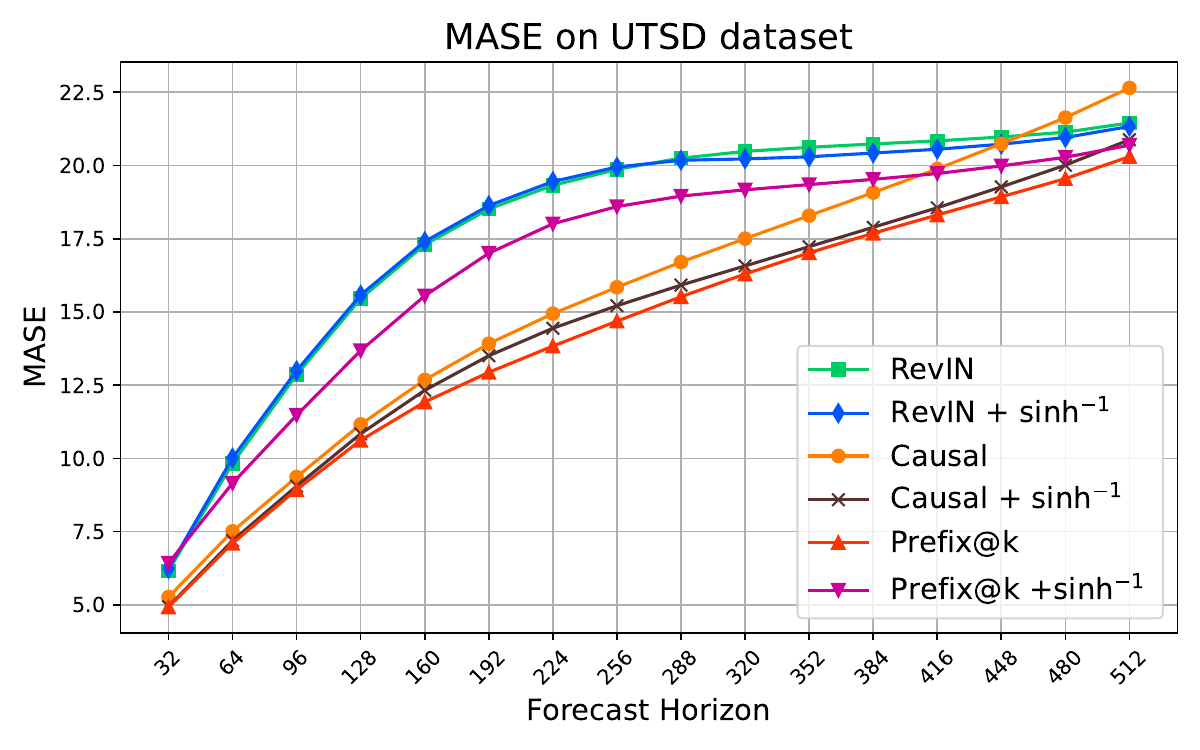}
        \caption{MASE 256}
        \label{fig:horizon_mase_256_utsd}
    \end{subfigure}

    \begin{subfigure}[t]{0.32\linewidth}
        \centering
        \includegraphics[width=\linewidth]{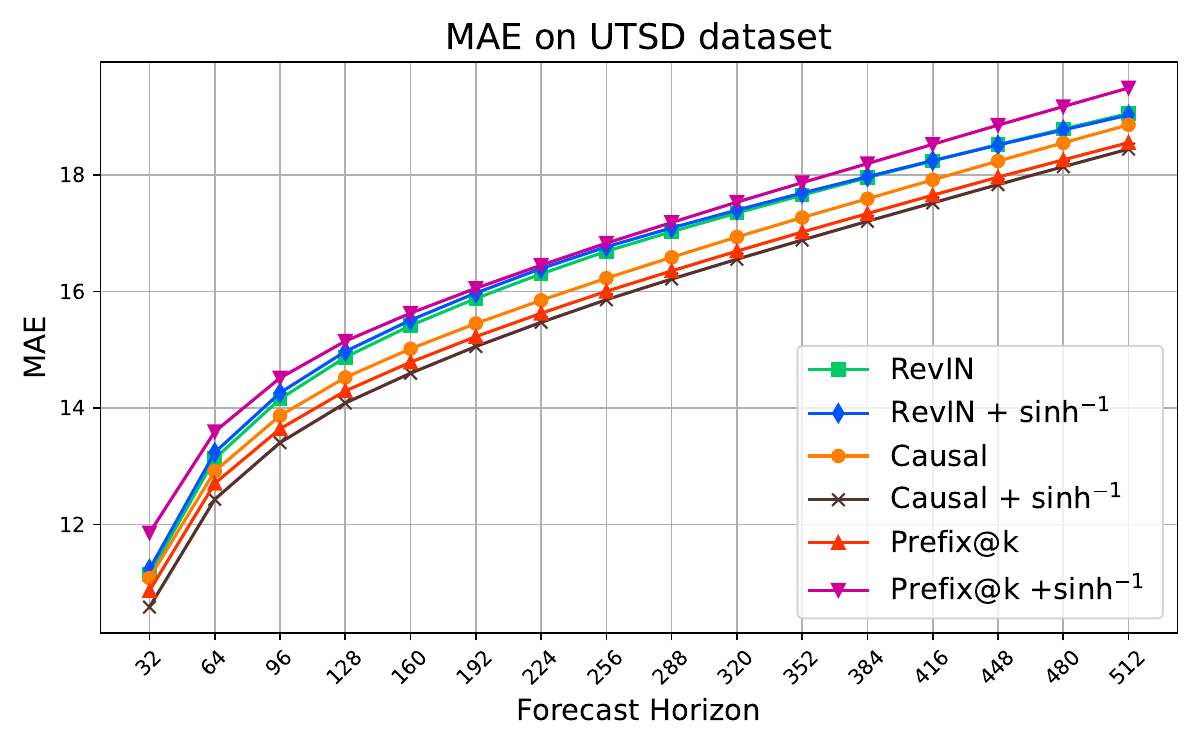}
        \caption{MAE 512}
        \label{fig:horizon_mae_512_utsd}
    \end{subfigure}
    \begin{subfigure}[t]{0.32\linewidth}
        \centering
        \includegraphics[width=\linewidth]{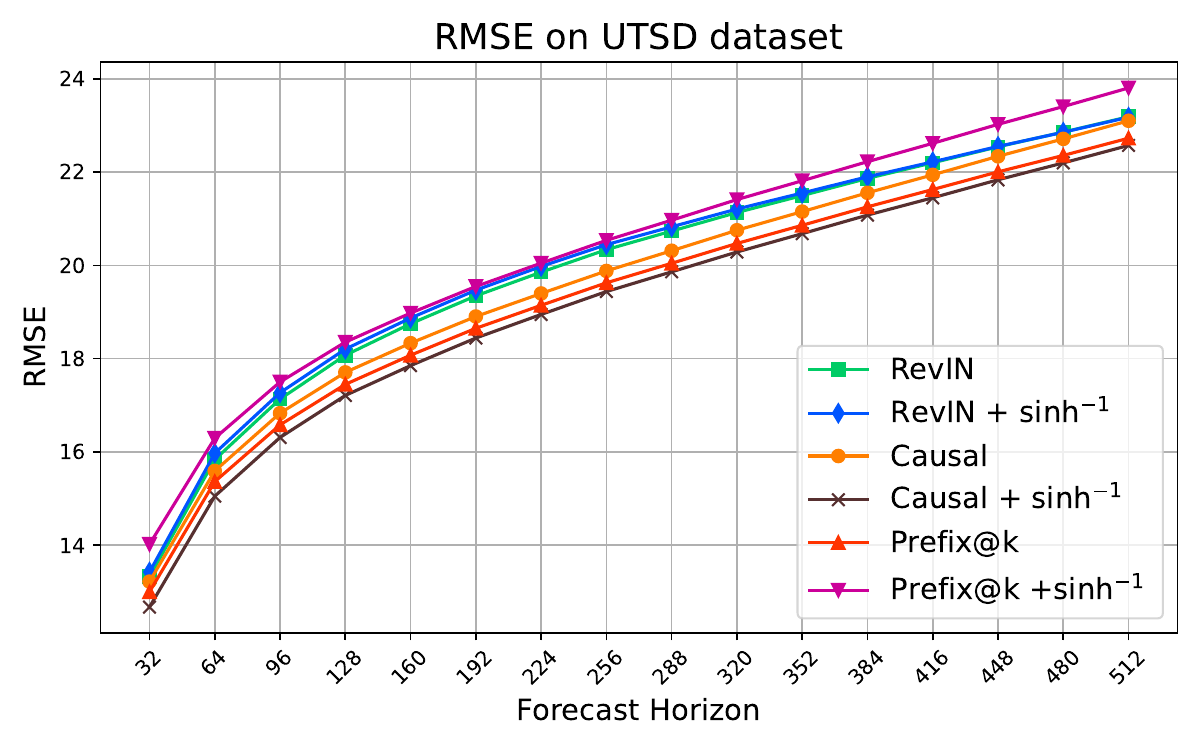}
        \caption{RMSE 512}
        \label{fig:horizon_rmse_512_utsd}
    \end{subfigure}
    \begin{subfigure}[t]{0.32\linewidth}
        \centering
        \includegraphics[width=\linewidth]{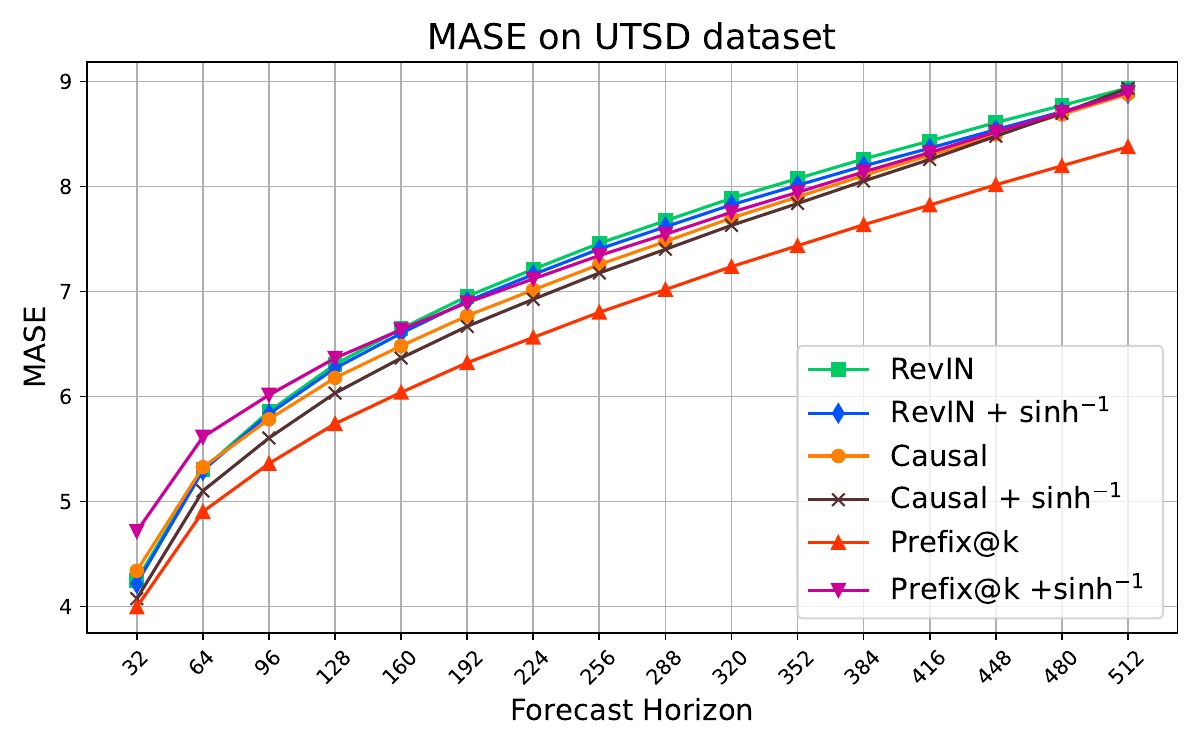}
        \caption{MASE 512}
        \label{fig:horizon_mase_512_utsd}
    \end{subfigure}

    \begin{subfigure}[t]{0.32\linewidth}
        \centering
        \includegraphics[width=\linewidth]{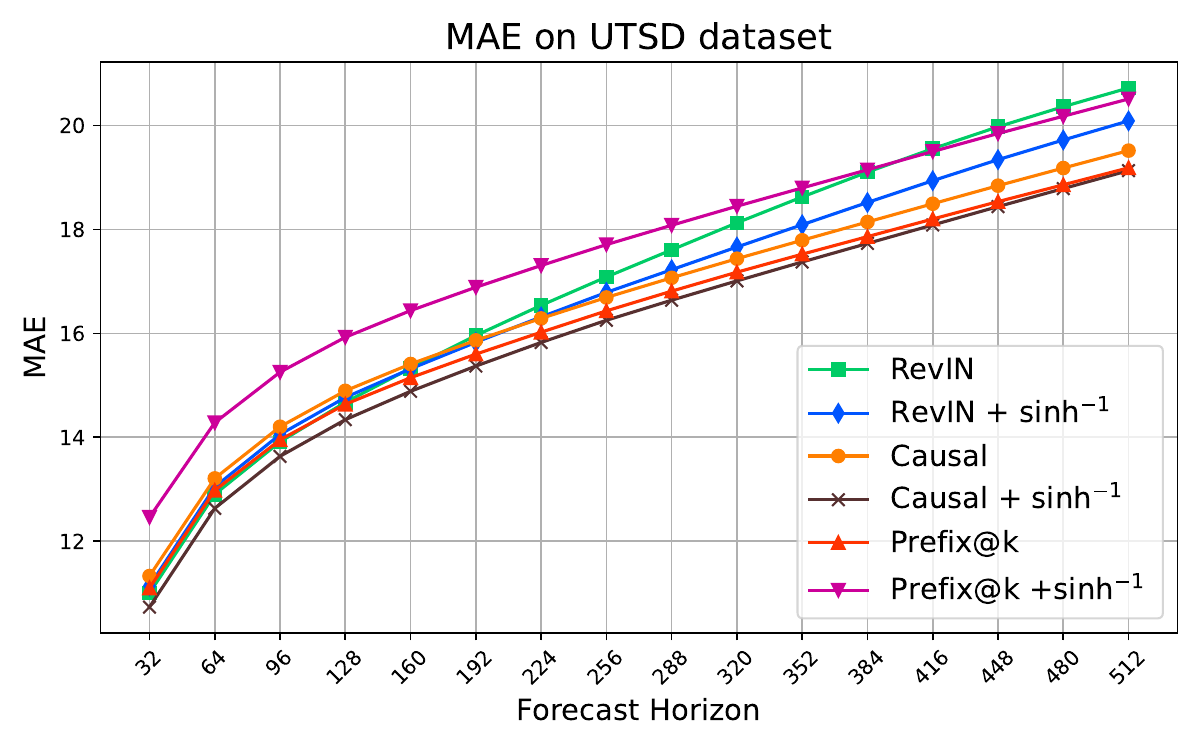}
        \caption{MAE 1024}
        \label{fig:horizon_mae_1024_utsd}
    \end{subfigure}
    \begin{subfigure}[t]{0.32\linewidth}
        \centering
        \includegraphics[width=\linewidth]{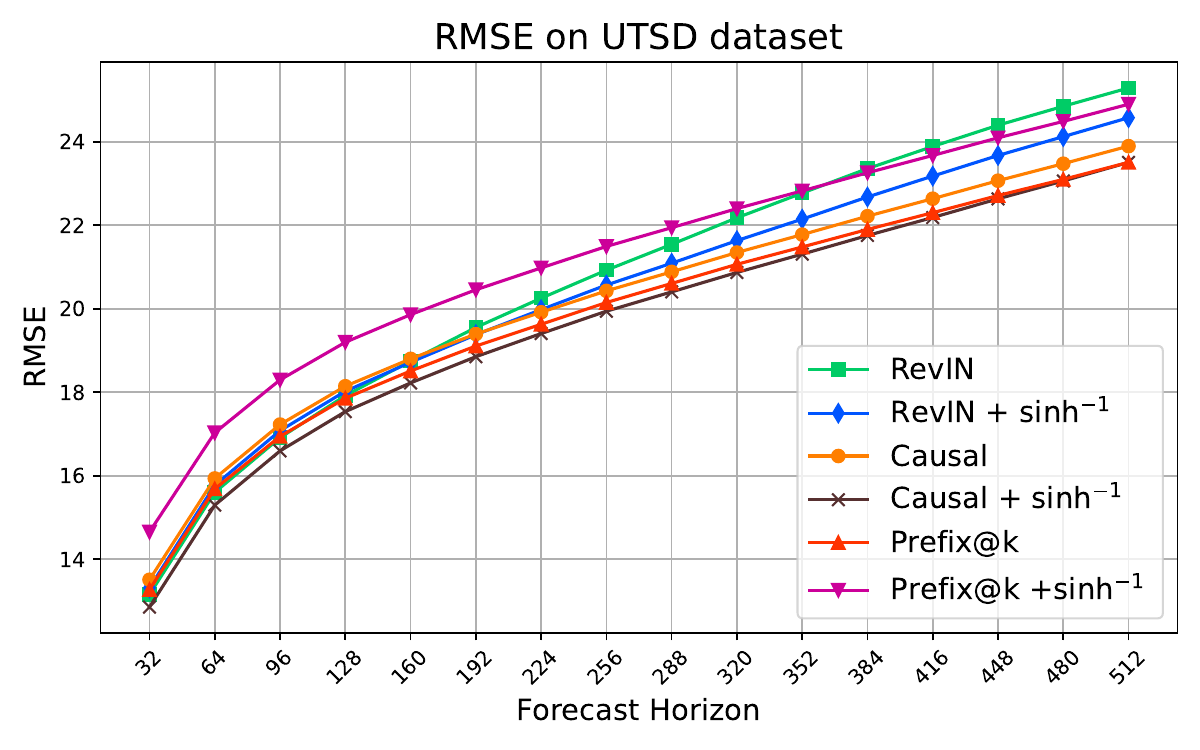}
        \caption{RMSE 1024}
        \label{fig:horizon_rmse_1024_utsd}
    \end{subfigure}
    \begin{subfigure}[t]{0.32\linewidth}
        \centering
        \includegraphics[width=\linewidth]{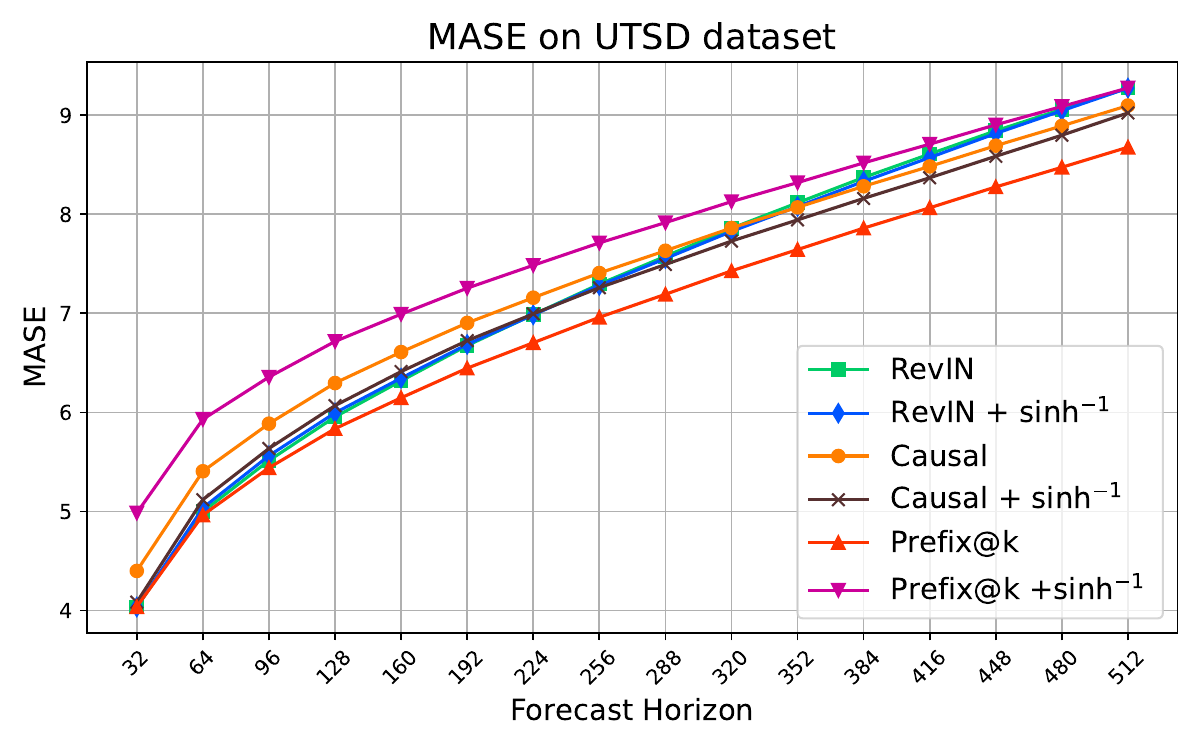}
        \caption{MASE 1024}
        \label{fig:horizon_mase_1024_utsd}
    \end{subfigure}

    \caption{Forecasting performance across prediction horizons on the UTSD-12G dataset for four context lengths (128, 256, 512, 1024).}
    \label{fig:horizon_utsd_comparison}
\end{figure}

Results on the UTSD-12G benchmark (Figure~\ref{fig:horizon_utsd_comparison}) reveal a more nuanced performance profile.
In terms of MAE and RMSE and for context length of 128, \revin{}, \revinasinh{} and \wurevin{} rank among the strongest methods. For MASE and context length of 128, all methods perform comparably.
For context lengths of 256 and 512, \revin{} and \revinasinh{} exhibit degraded performance, while \wurevin{}, \causalrevin{} and \causalrevinasinh{} rank among the best methods.
Finally, for context length of 1024, \wurevin{} seems to be the best-performing method for MASE, while for MAE and RMSE \causalrevinasinh{} and \wurevin{} show competitive performance.
\wurevinasinh{} generally exhibits poor performance across all settings.

\begin{figure}[h]
    \centering
    \begin{subfigure}[t]{0.32\linewidth}
        \centering
        \includegraphics[width=\linewidth]{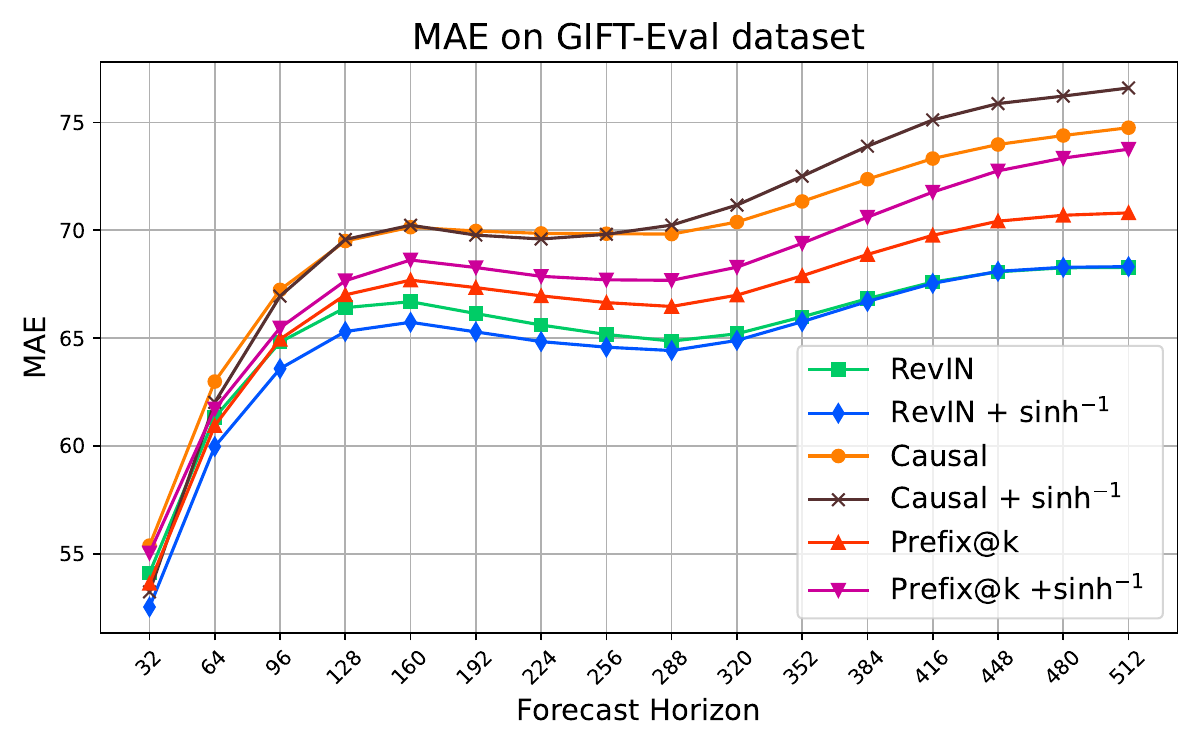}
        \caption{MAE 128}
        \label{fig:horizon_mae_128_gift_eval}
    \end{subfigure}
    \begin{subfigure}[t]{0.32\linewidth}
        \centering
        \includegraphics[width=\linewidth]{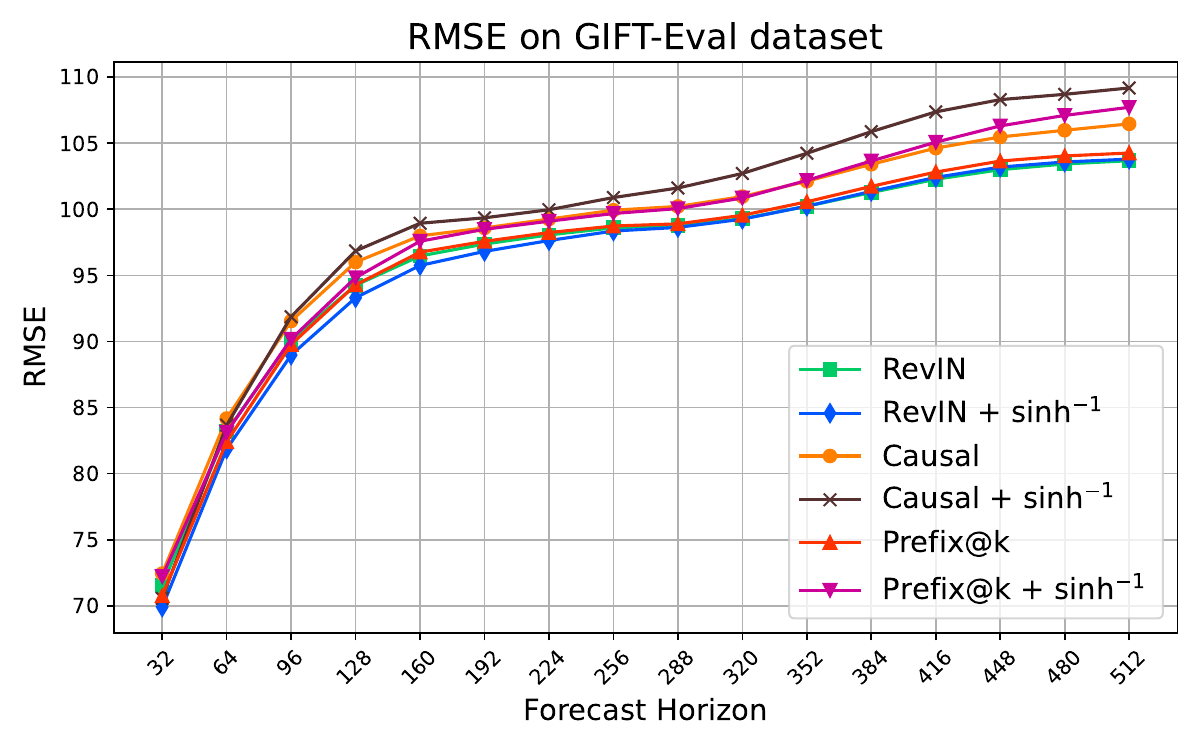}
        \caption{RMSE 128}
        \label{fig:horizon_rmse_128_gift_eval}
    \end{subfigure}
    \begin{subfigure}[t]{0.32\linewidth}
        \centering
        \includegraphics[width=\linewidth]{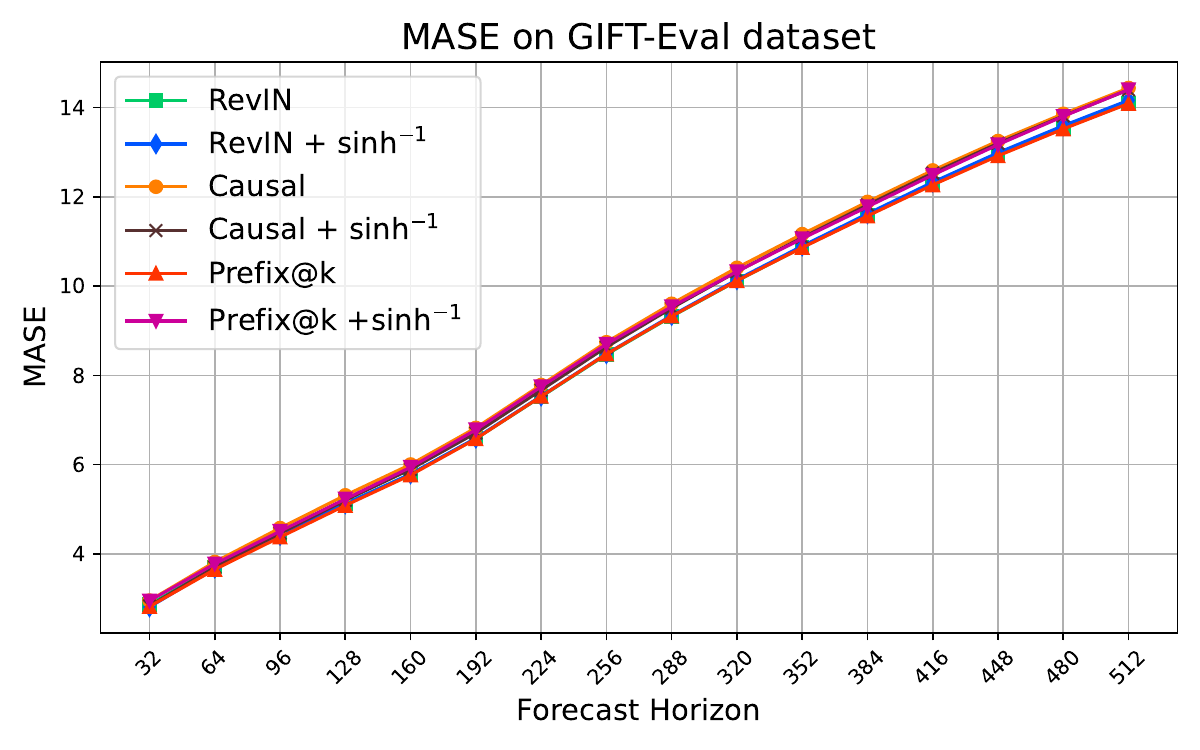}
        \caption{MASE 128}
        \label{fig:horizon_mase_128_gift_eval}
    \end{subfigure}

    \begin{subfigure}[t]{0.32\linewidth}
        \centering
        \includegraphics[width=\linewidth]{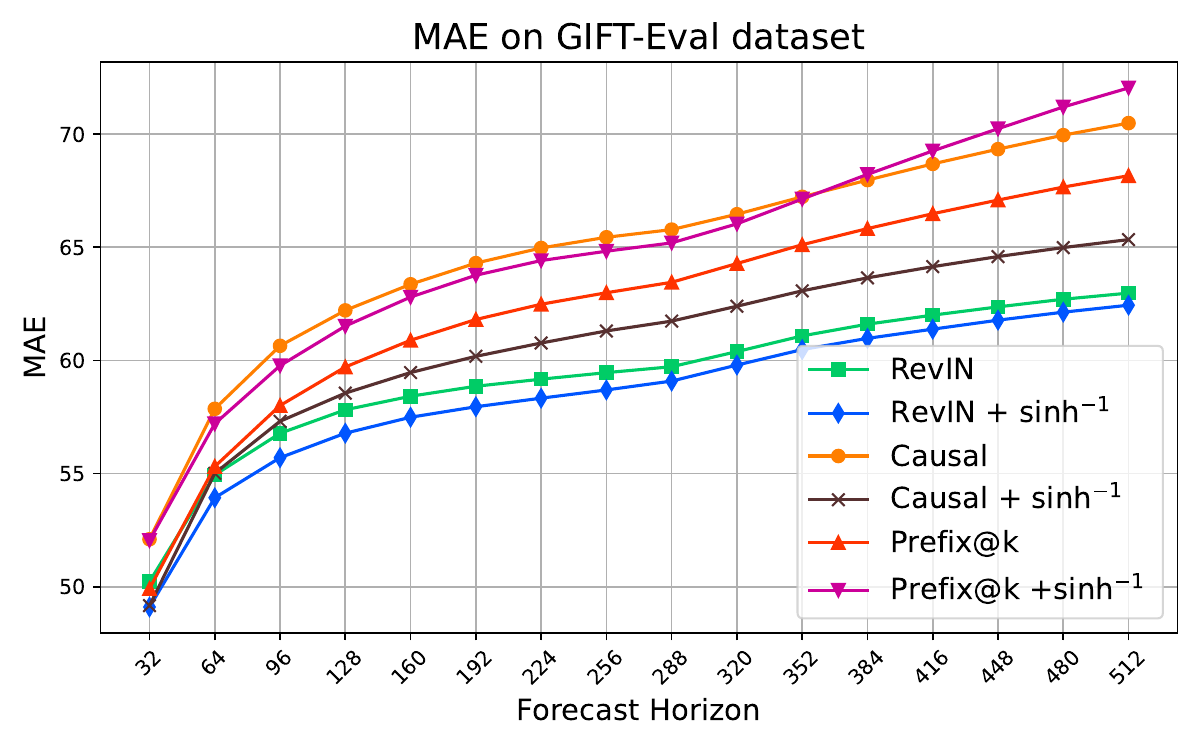}
        \caption{MAE 256}
        \label{fig:horizon_mae_256_gift_eval}
    \end{subfigure}
    \begin{subfigure}[t]{0.32\linewidth}
        \centering
        \includegraphics[width=\linewidth]{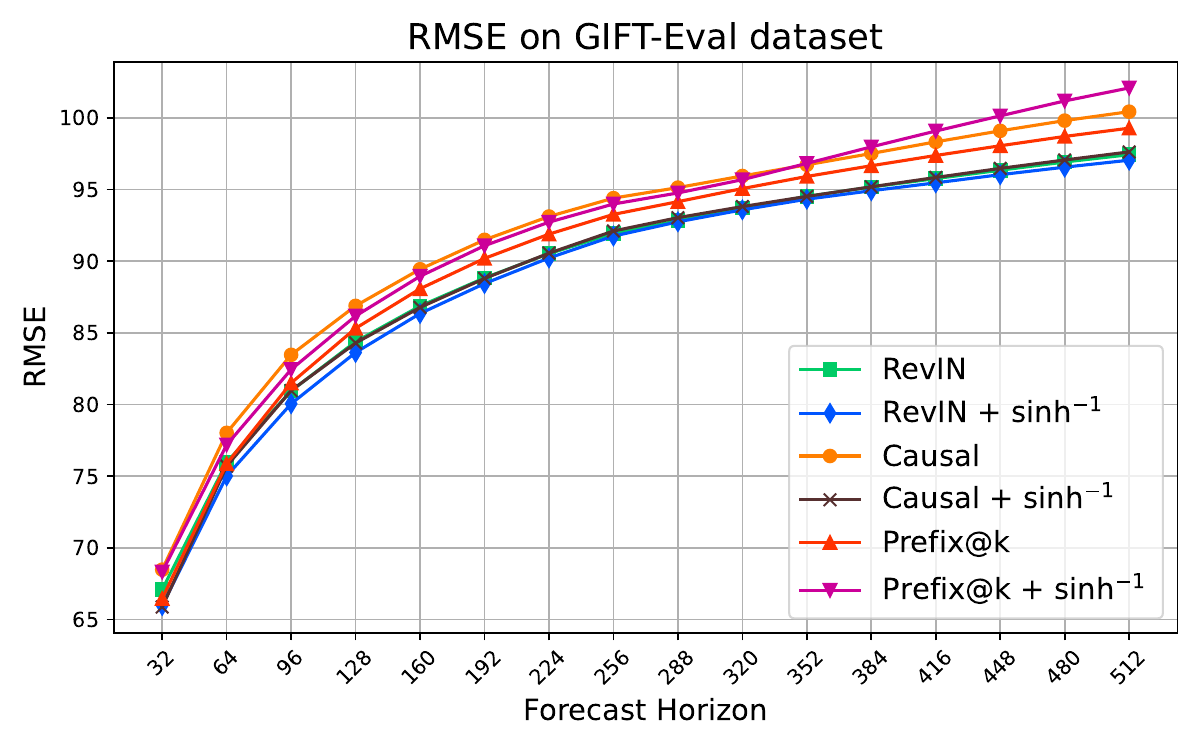}
        \caption{RMSE 256}
        \label{fig:horizon_rmse_256_gift_eval}
    \end{subfigure}
    \begin{subfigure}[t]{0.32\linewidth}
        \centering
        \includegraphics[width=\linewidth]{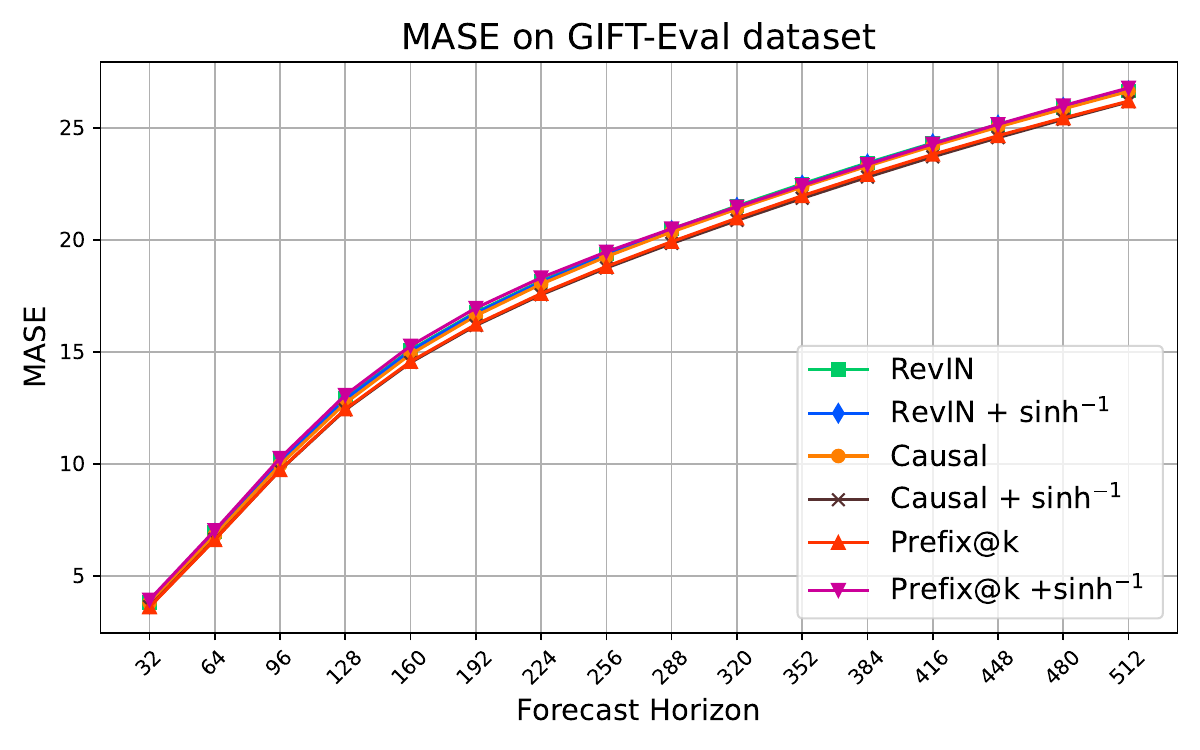}
        \caption{MASE 256}
        \label{fig:horizon_mase_256_gift_eval}
    \end{subfigure}

    \begin{subfigure}[t]{0.32\linewidth}
        \centering
        \includegraphics[width=\linewidth]{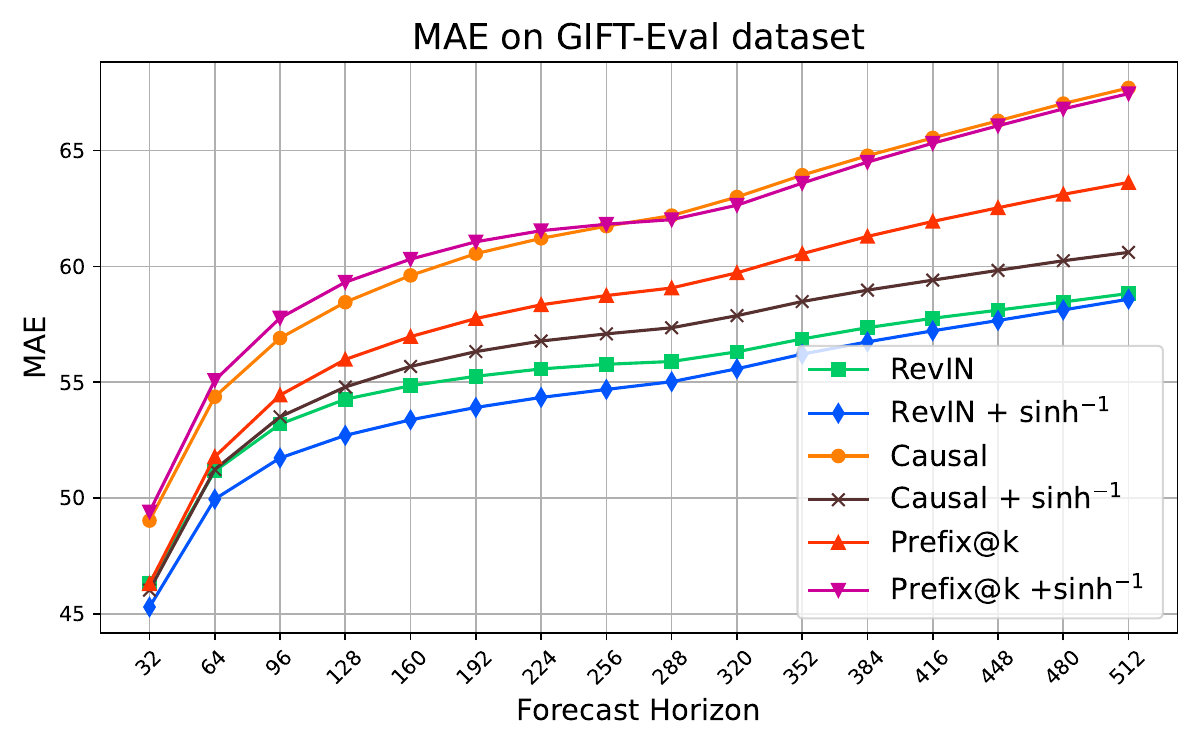}
        \caption{MAE 512}
        \label{fig:horizon_mae_512_gift_eval}
    \end{subfigure}
    \begin{subfigure}[t]{0.32\linewidth}
        \centering
        \includegraphics[width=\linewidth]{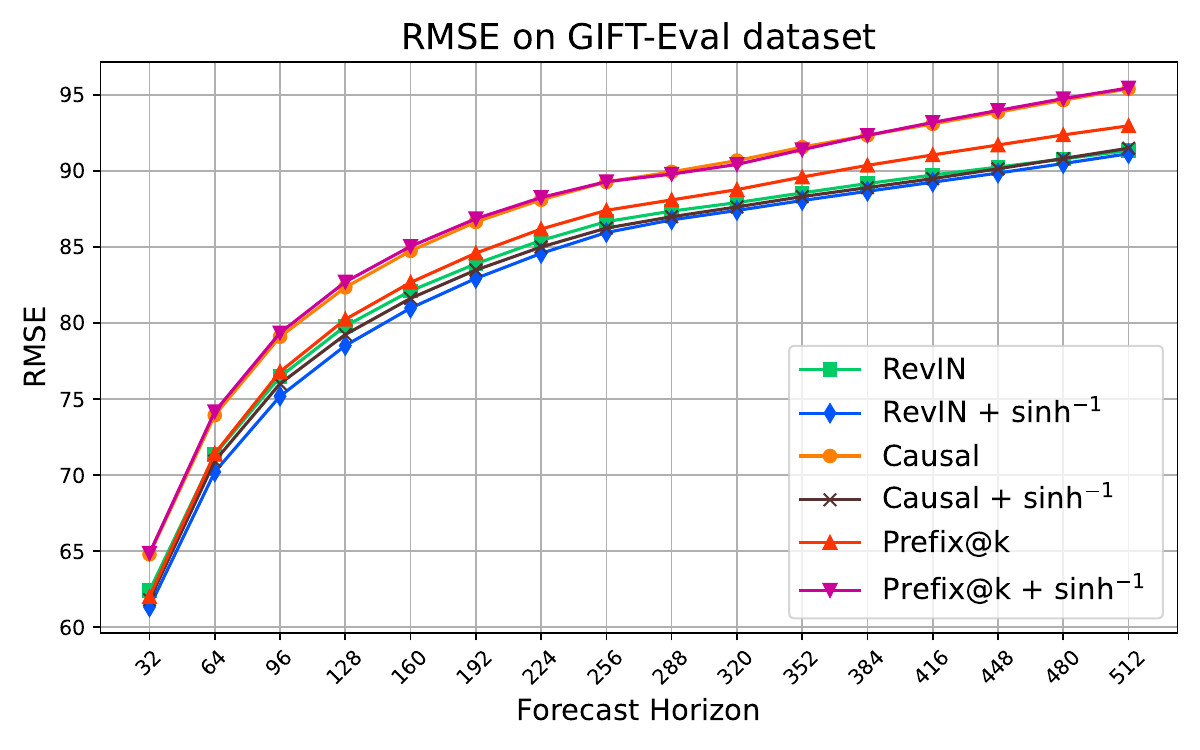}
        \caption{RMSE 512}
        \label{fig:horizon_rmse_512_gift_eval}
    \end{subfigure}
    \begin{subfigure}[t]{0.32\linewidth}
        \centering
        \includegraphics[width=\linewidth]{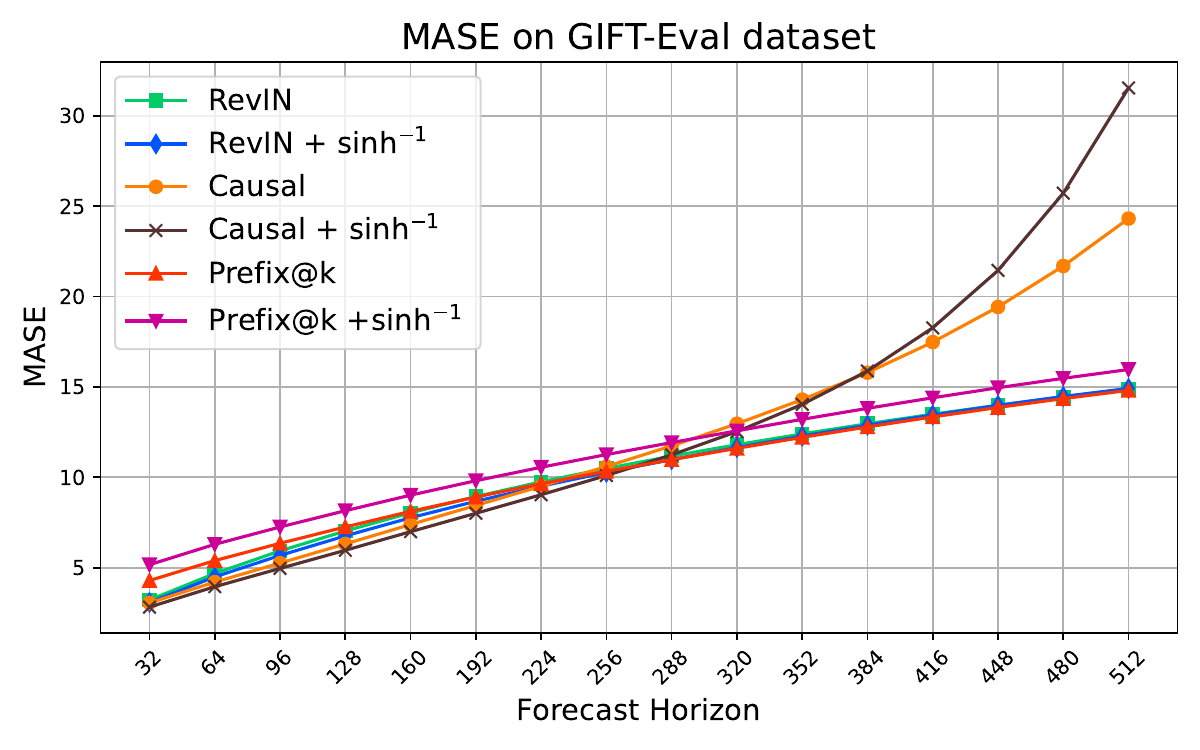}
        \caption{MASE 512}
        \label{fig:horizon_mase_512_gift_eval}
    \end{subfigure}

    \begin{subfigure}[t]{0.32\linewidth}
        \centering
        \includegraphics[width=\linewidth]{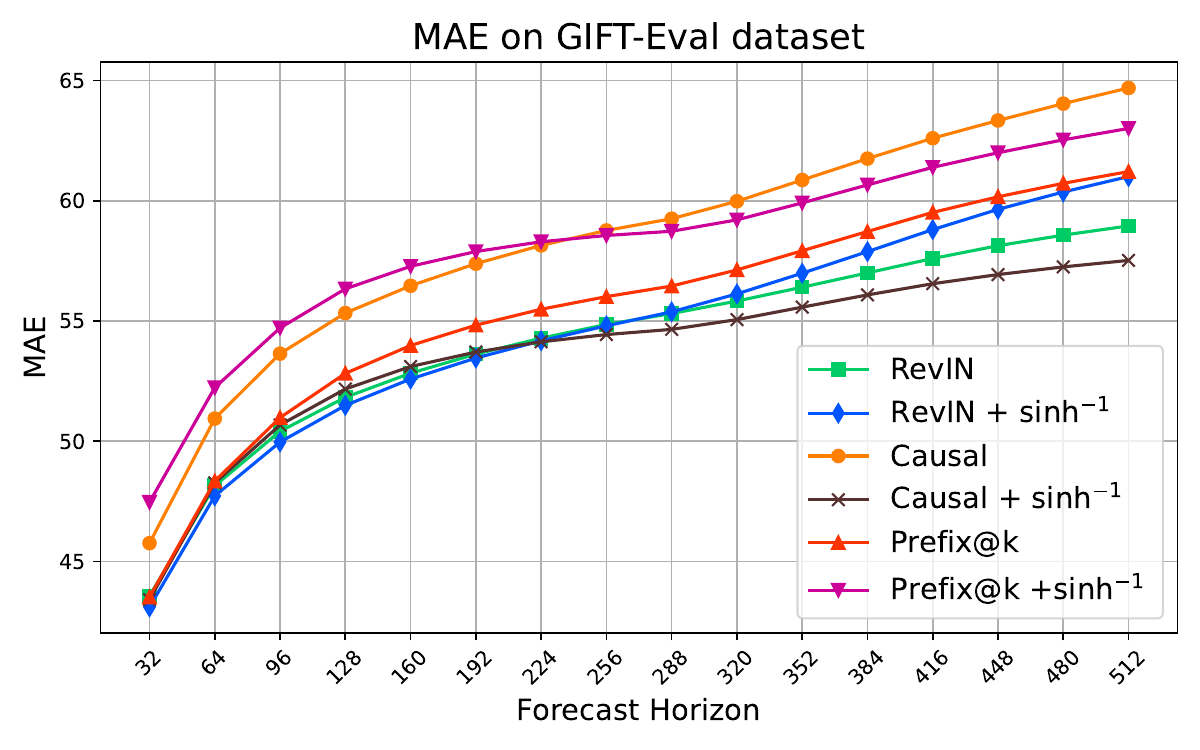}
        \caption{MAE 1024}
        \label{fig:horizon_mae_1024_gift_eval}
    \end{subfigure}
    \begin{subfigure}[t]{0.32\linewidth}
        \centering
        \includegraphics[width=\linewidth]{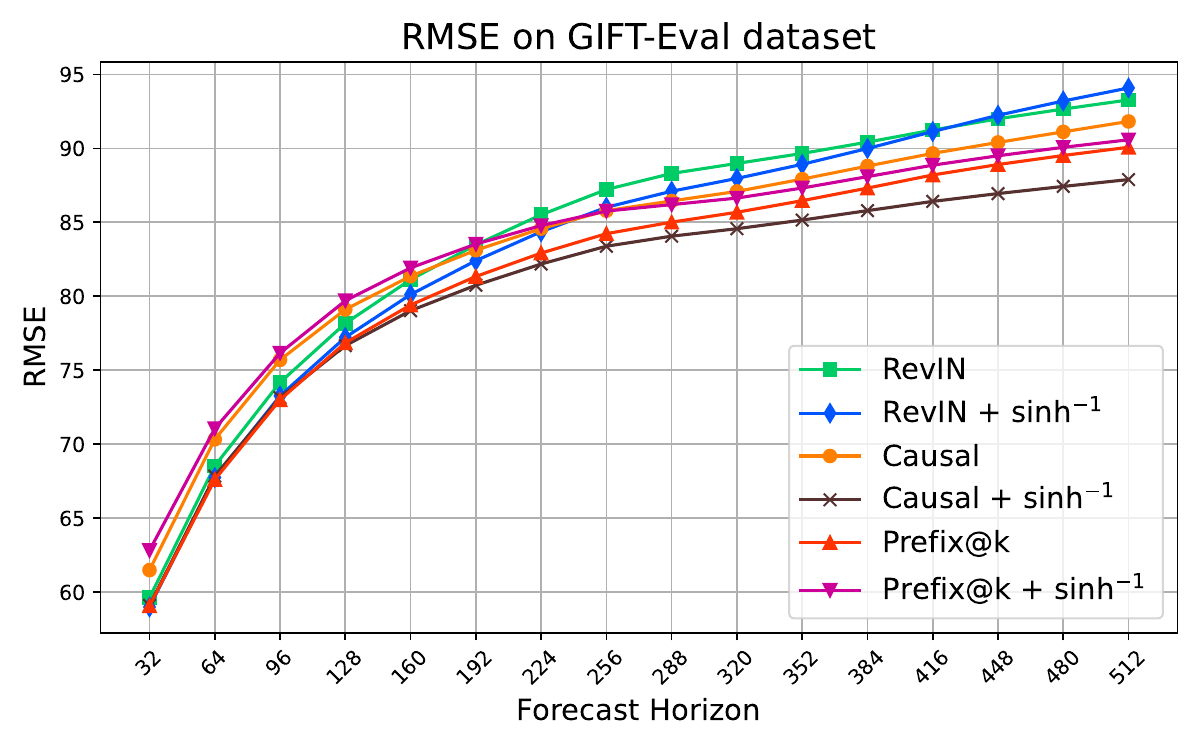}
        \caption{RMSE 1024}
        \label{fig:horizon_rmse_1024_gift_eval}
    \end{subfigure}
    \begin{subfigure}[t]{0.32\linewidth}
        \centering
        \includegraphics[width=\linewidth]{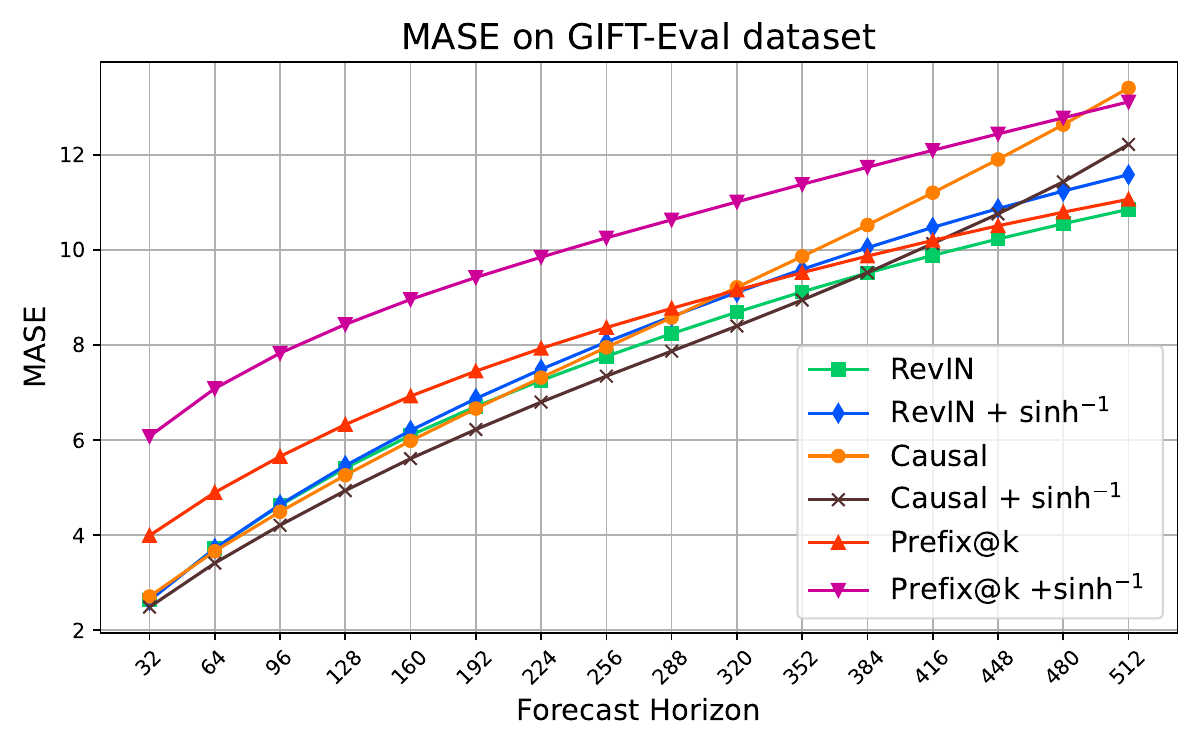}
        \caption{MASE 1024}
        \label{fig:horizon_mase_1024_gift_eval}
    \end{subfigure}

    \caption{Forecasting performance across prediction horizons on the GIFT-Eval dataset for four context lengths (128, 256, 512, 1024).}
    \label{fig:horizon_gift_eval_comparison}
\end{figure}

The GIFT-Eval benchmark (Figure~\ref{fig:horizon_gift_eval_comparison}) also allows for a more nuanced interpretation of the results.
For MAE and RMSE, the non-causal configurations \revinasinh{} and \revin{} achieve the strongest performance up to a context length of 1024. 
The performance of \causalrevinasinh{} improves as the context length increases, eventually surpassing both \revinasinh{} and \revin{} at a context length of 1024.
Under the MASE metric, all models perform comparably up to a context length of 512. 
However, \causalrevinasinh{} and \causalrevin{} exhibit a pronounced scaling failure at a context length of 512 for forecasting horizons beyond 288, despite remaining competitive in terms of MAE and RMSE.

\begin{figure}[h]
    \centering
    \begin{subfigure}[t]{0.32\linewidth}
        \centering
        \includegraphics[width=\linewidth]{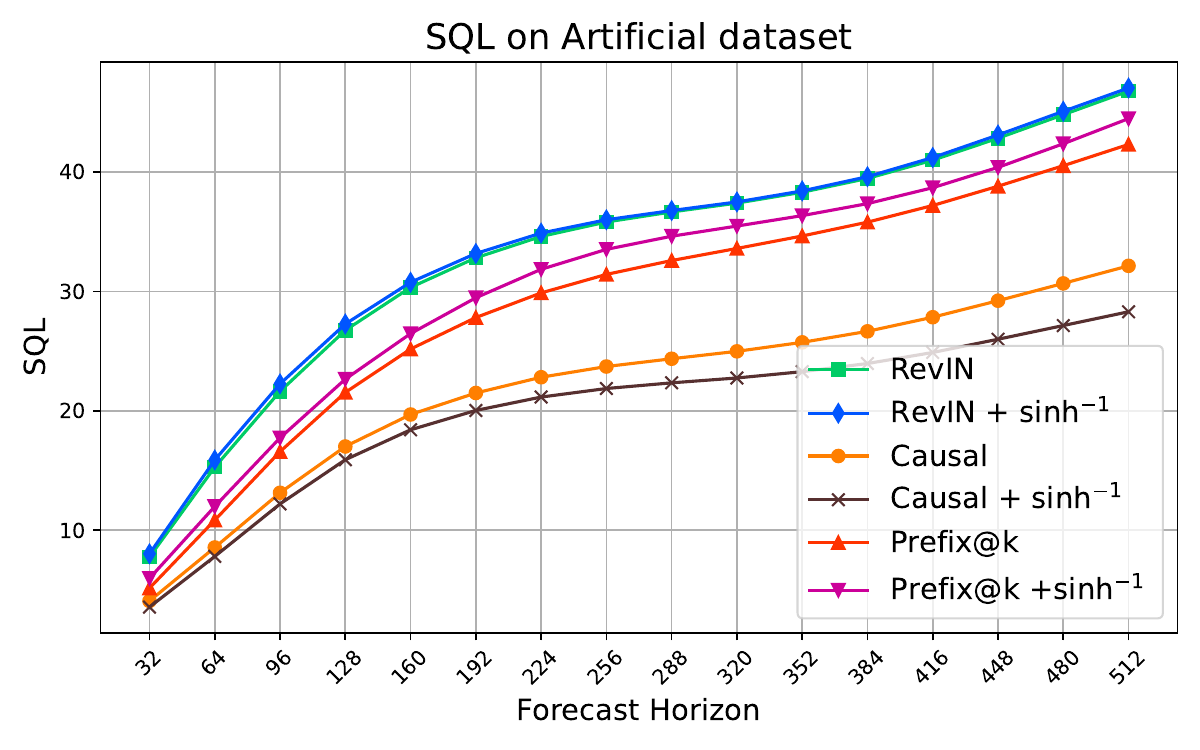}
        \caption{Synthetic 128}
        \label{fig:horizon_sql_128_artificial}
    \end{subfigure}
    \begin{subfigure}[t]{0.32\linewidth}
        \centering
        \includegraphics[width=\linewidth]{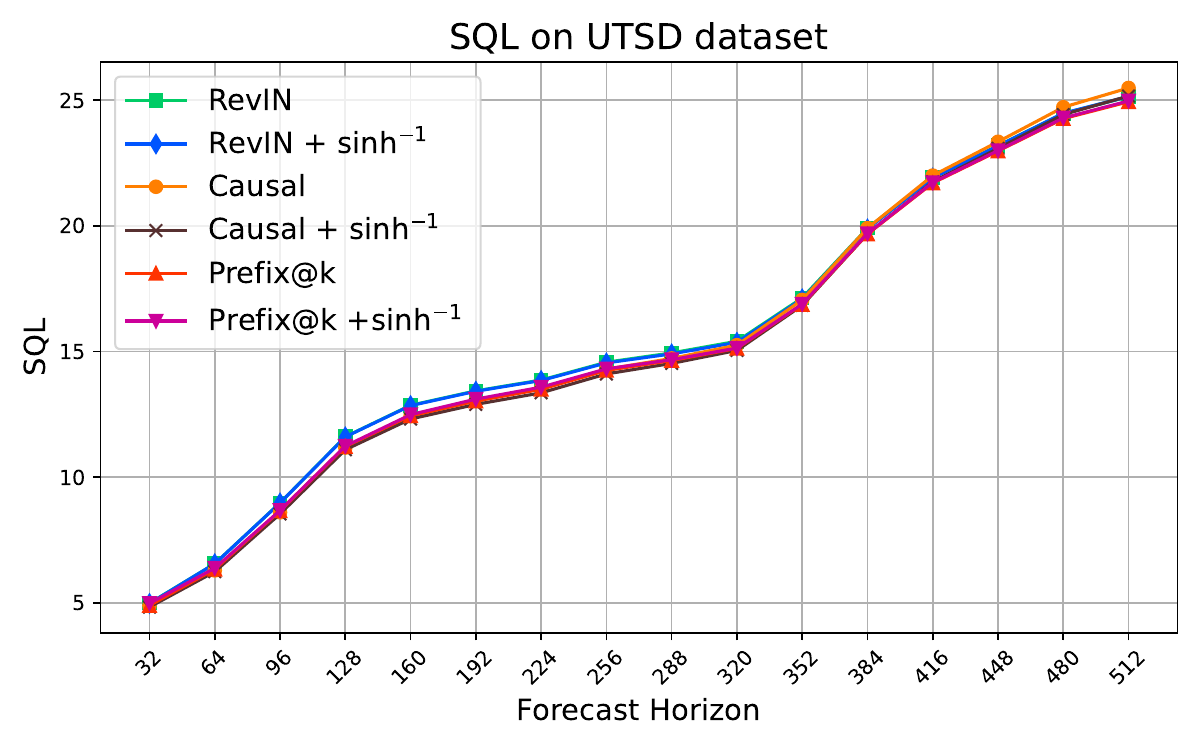}
        \caption{UTSD-12G 128}
        \label{fig:horizon_sql_128_utsd}
    \end{subfigure}
    \begin{subfigure}[t]{0.32\linewidth}
        \centering
        \includegraphics[width=\linewidth]{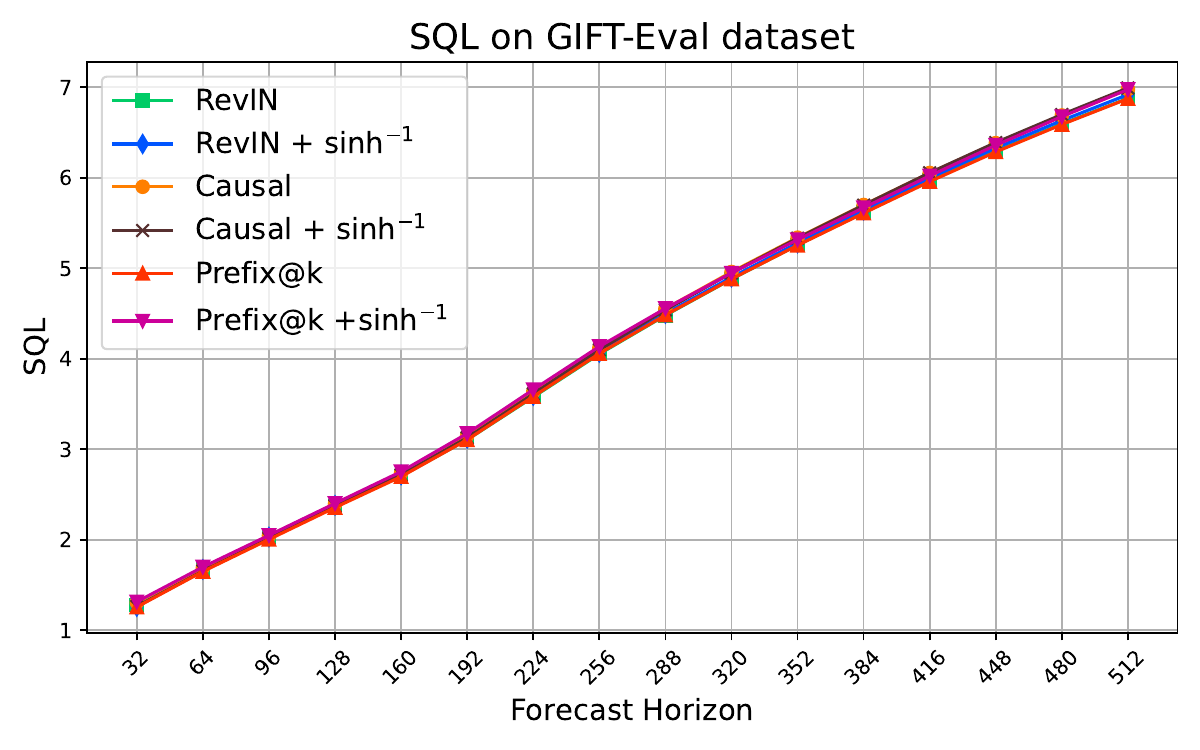}
        \caption{GIFT-Eval 128}
        \label{fig:horizon_sql_128_gift_eval}
    \end{subfigure}

    \begin{subfigure}[t]{0.32\linewidth}
        \centering
        \includegraphics[width=\linewidth]{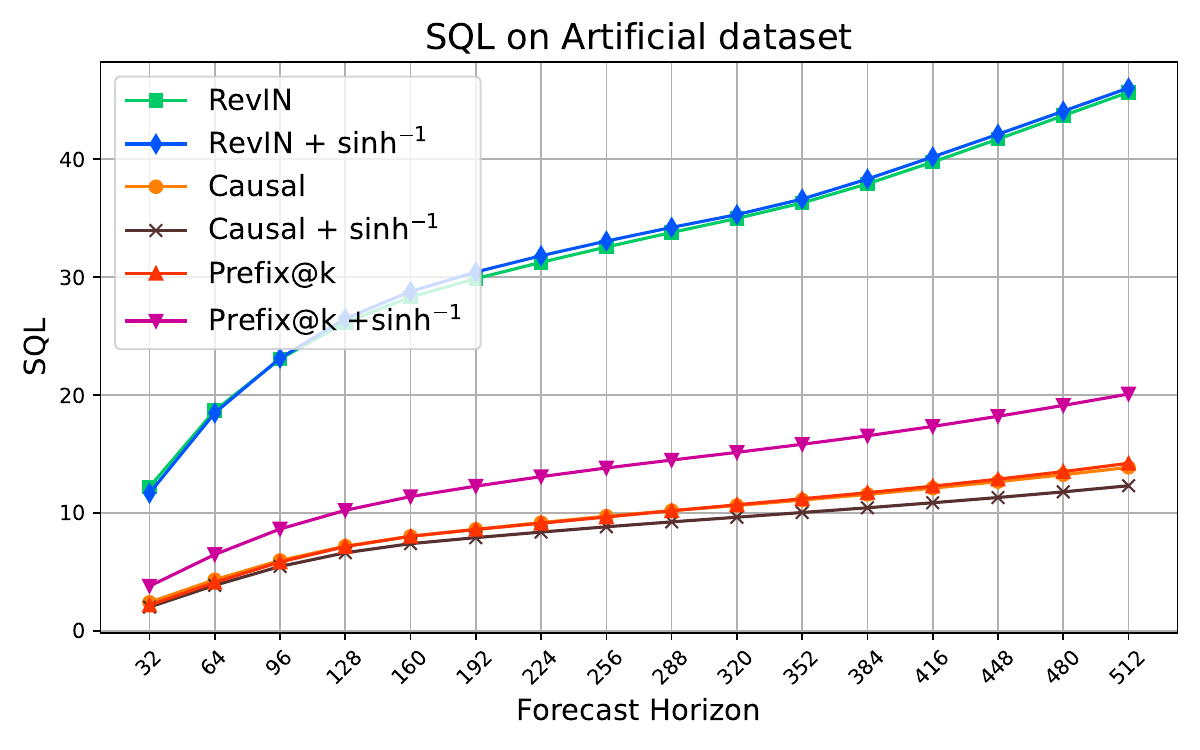}
        \caption{Synthetic 256}
        \label{fig:horizon_sql_256_artificial}
    \end{subfigure}
    \begin{subfigure}[t]{0.32\linewidth}
        \centering
        \includegraphics[width=\linewidth]{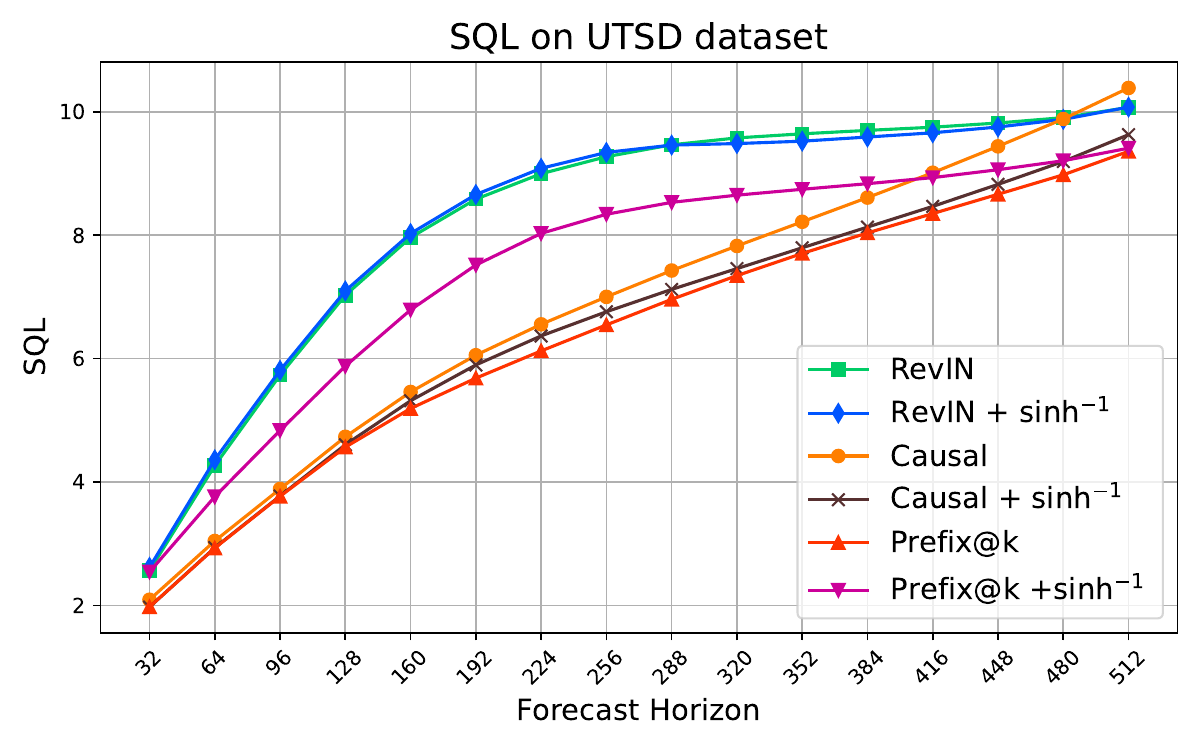}
        \caption{UTSD-12G 256}
        \label{fig:horizon_sql_256_utsd}
    \end{subfigure}
    \begin{subfigure}[t]{0.32\linewidth}
        \centering
        \includegraphics[width=\linewidth]{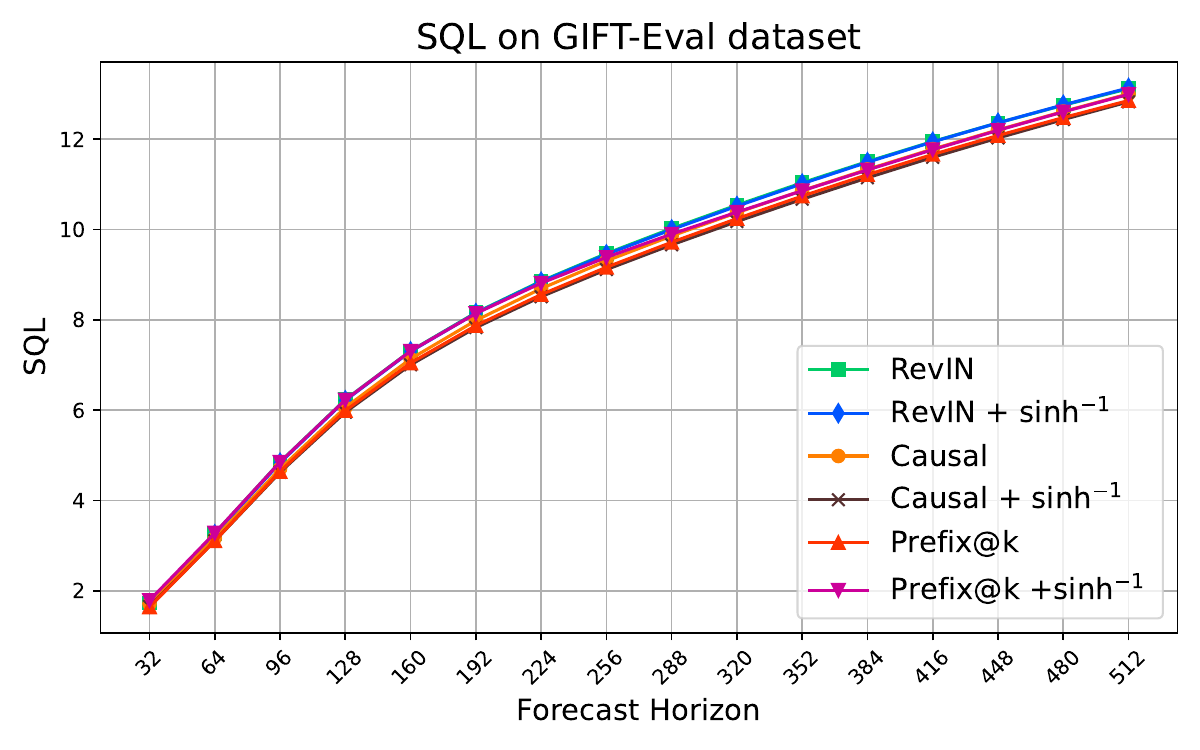}
        \caption{GIFT-Eval 256}
        \label{fig:horizon_sql_256_gift_eval}
    \end{subfigure}

    \begin{subfigure}[t]{0.32\linewidth}
        \centering
        \includegraphics[width=\linewidth]{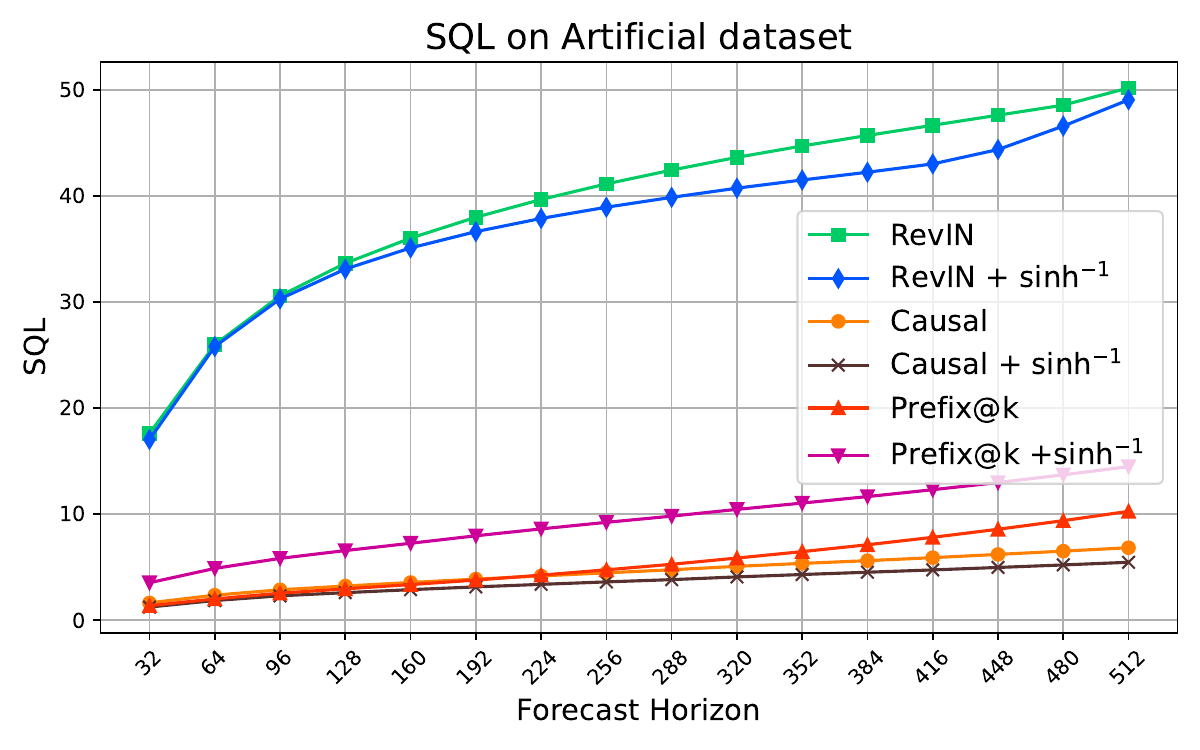}
        \caption{Synthetic 512}
        \label{fig:horizon_sql_512_artificial}
    \end{subfigure}
    \begin{subfigure}[t]{0.32\linewidth}
        \centering
        \includegraphics[width=\linewidth]{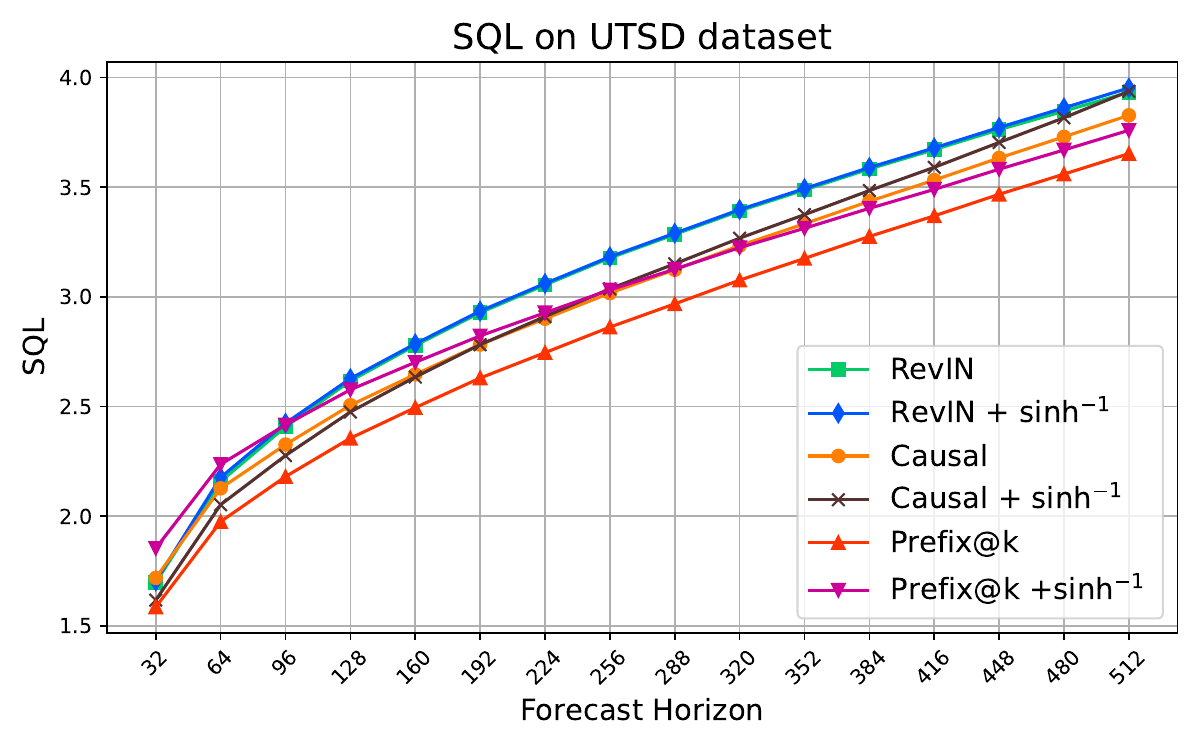}
        \caption{UTSD-12G 512}
        \label{fig:horizon_sql_512_utsd}
    \end{subfigure}
    \begin{subfigure}[t]{0.32\linewidth}
        \centering
        \includegraphics[width=\linewidth]{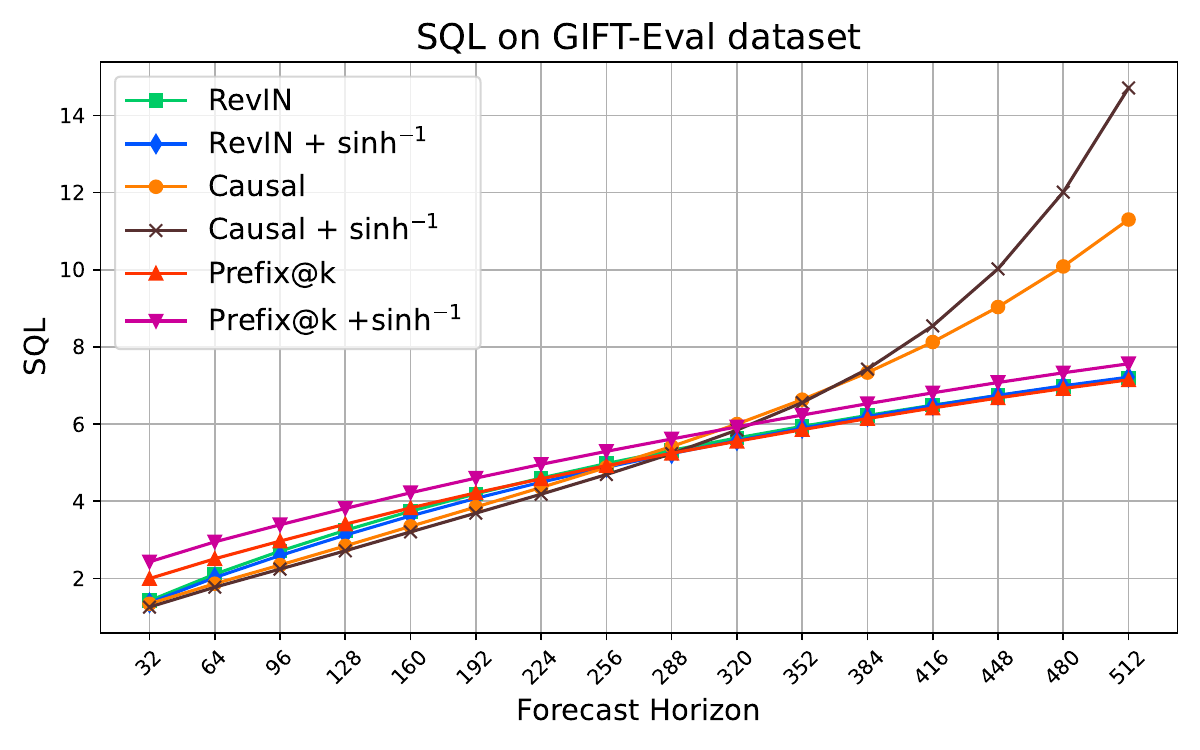}
        \caption{GIFT-Eval 512}
        \label{fig:horizon_sql_512_gift_eval}
    \end{subfigure}

    \begin{subfigure}[t]{0.32\linewidth}
        \centering
        \includegraphics[width=\linewidth]{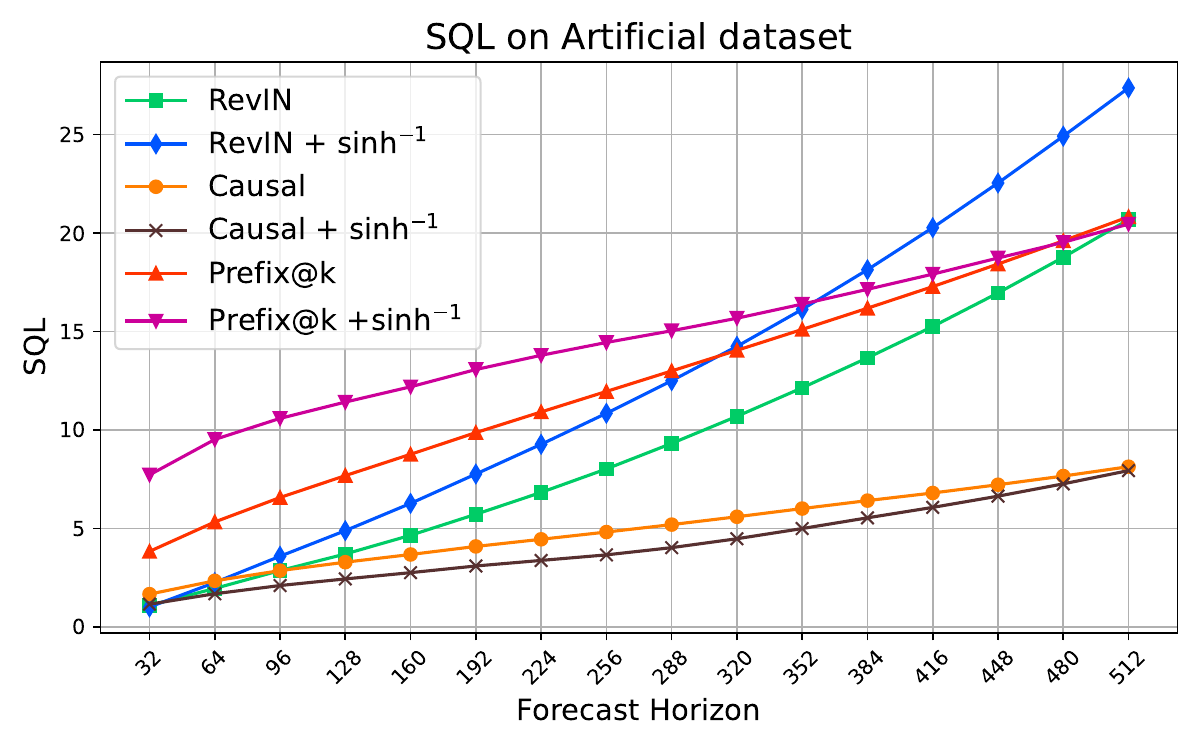}
        \caption{Synthetic 1024}
        \label{fig:horizon_sql_1024_artificial}
    \end{subfigure}
    \begin{subfigure}[t]{0.32\linewidth}
        \centering
        \includegraphics[width=\linewidth]{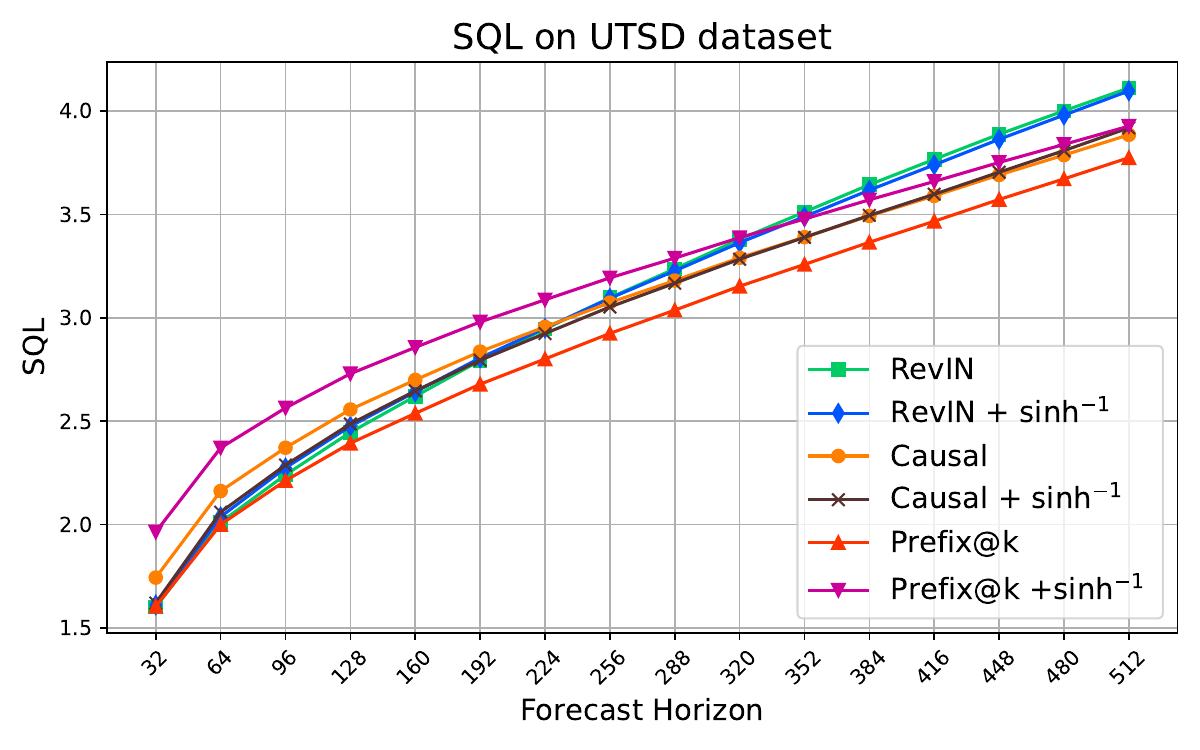}
        \caption{UTSD-12G 1024}
        \label{fig:horizon_sql_1024_utsd}
    \end{subfigure}
    \begin{subfigure}[t]{0.32\linewidth}
        \centering
        \includegraphics[width=\linewidth]{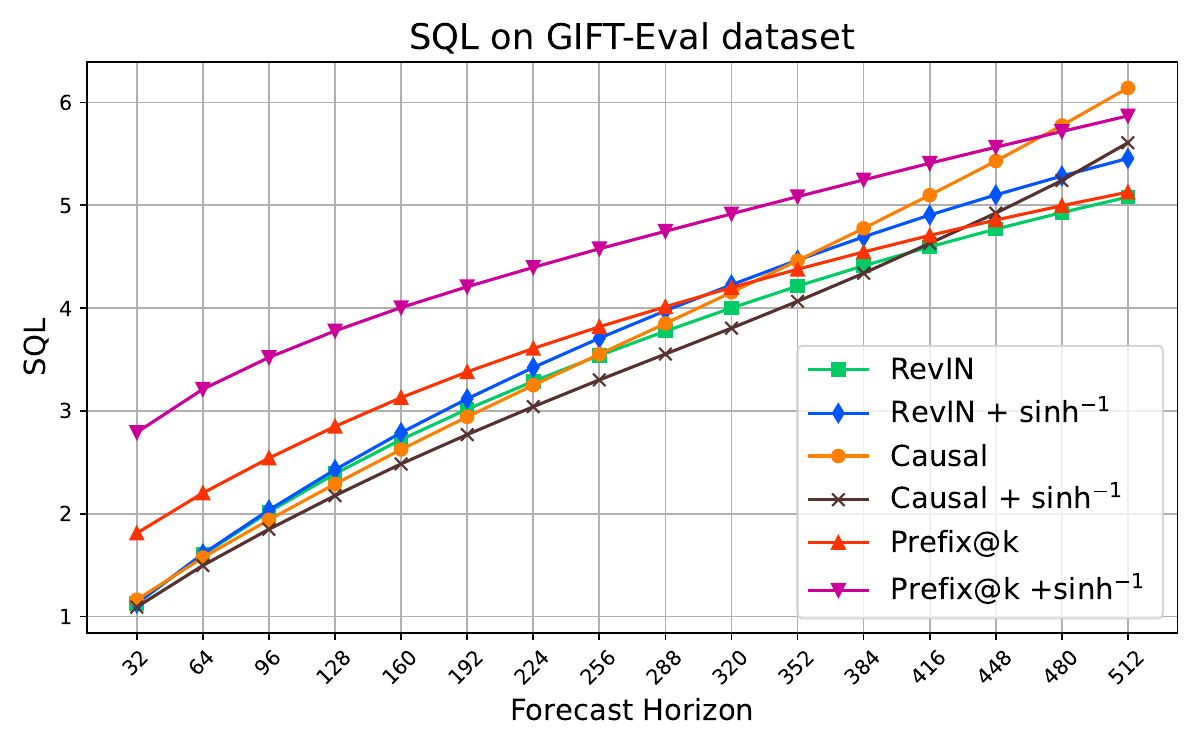}
        \caption{GIFT-Eval 1024}
        \label{fig:horizon_sql_1024_gift_eval}
    \end{subfigure}

    \caption{Scaled Quantile Loss as a forecasting performance metric across prediction horizons on all datasets, evaluated for four context lengths (128, 256, 512, and 1024).}
    \label{fig:horizon_sql_comparison}
\end{figure}

In Figure~\ref{fig:horizon_sql_comparison}, we report forecasting performance across prediction horizons using the Scaled Quantile Loss metric, aggregated over all datasets and context lengths.
On the synthetic benchmark, \causalrevinasinh{} and \causalrevin{} consistently achieve the best performance across all horizons and context lengths. In contrast, \revinasinh{} and \revin{} exhibit 
the weakest performance across horizons and context lengths, with the exception of the longest context length (1024), where they perform comparably to \wurevin{} and \wurevinasinh{} on average across horizons.
On the UTSD-12G benchmark, the results are less clear-cut. \wurevin{} appears to deliver the strongest overall performance, whereas \revinasinh{}, \revin{}, and \wurevinasinh{} tend to yield the weakest results.
Finally, on the GIFT-Eval benchmark, for context lengths up to 256, all models perform comparably across horizons. However, at a context length of 512, \causalrevinasinh{} and \causalrevin{} 
exhibit a pronounced scaling failure for forecasting horizons beyond 288. For a context length of 1024, the models again perform similarly on average across horizons, with 
the exception of \wurevinasinh{}, which shows consistently degraded performance across all horizons.

\subsection{Example of normalized signal and error analysis}\label{normalization_example}
\begin{figure}[h]
    \centering
    \begin{minipage}[t]{0.47\linewidth}
        \centering
        \includegraphics[width=0.9\linewidth]{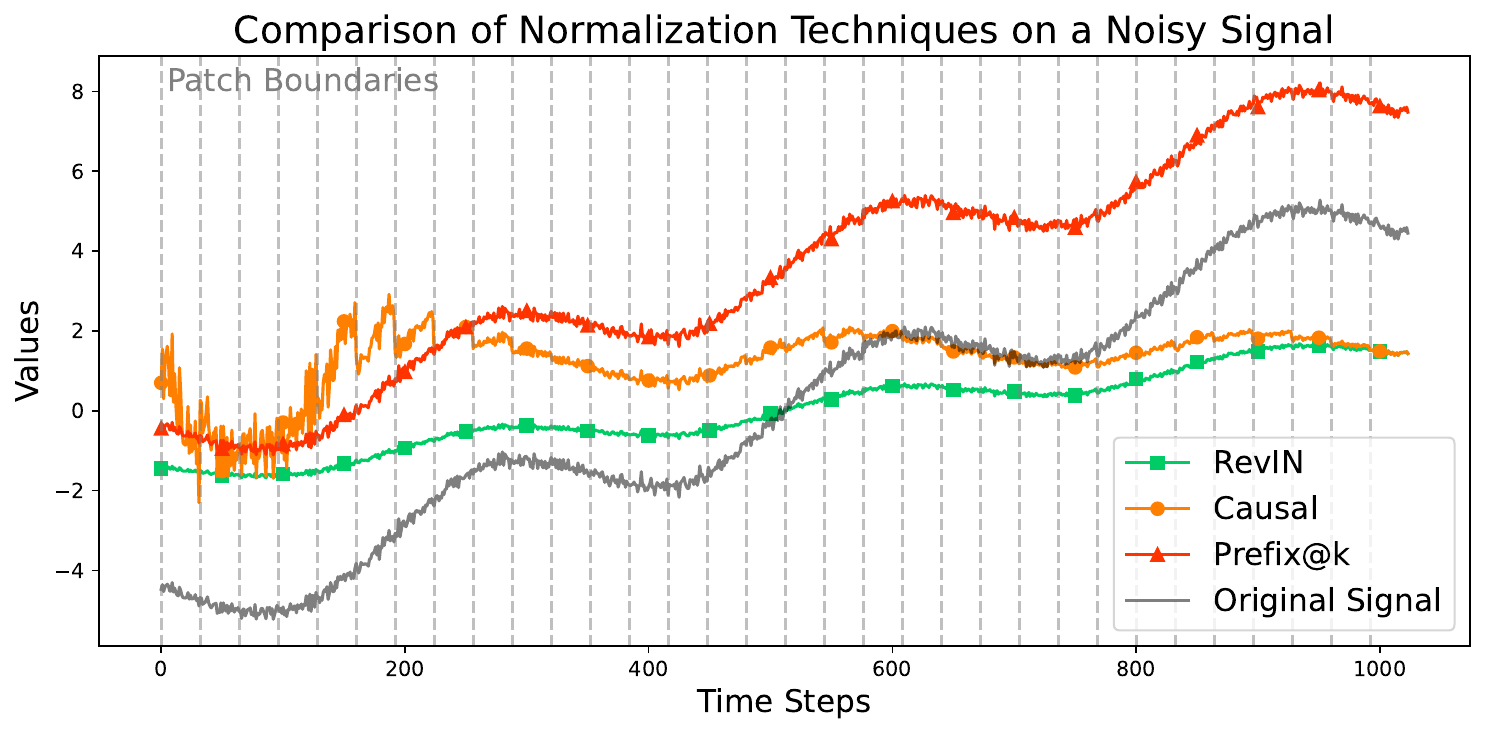}
        \caption{Example of normalization strategies on a simple increasing sinusoidal.}
        \label{fig:normalization_example}
    \end{minipage}
    \hspace{0.02\linewidth} 
    \begin{minipage}[t]{0.47\linewidth}
        \centering
        \includegraphics[width=0.9\linewidth]{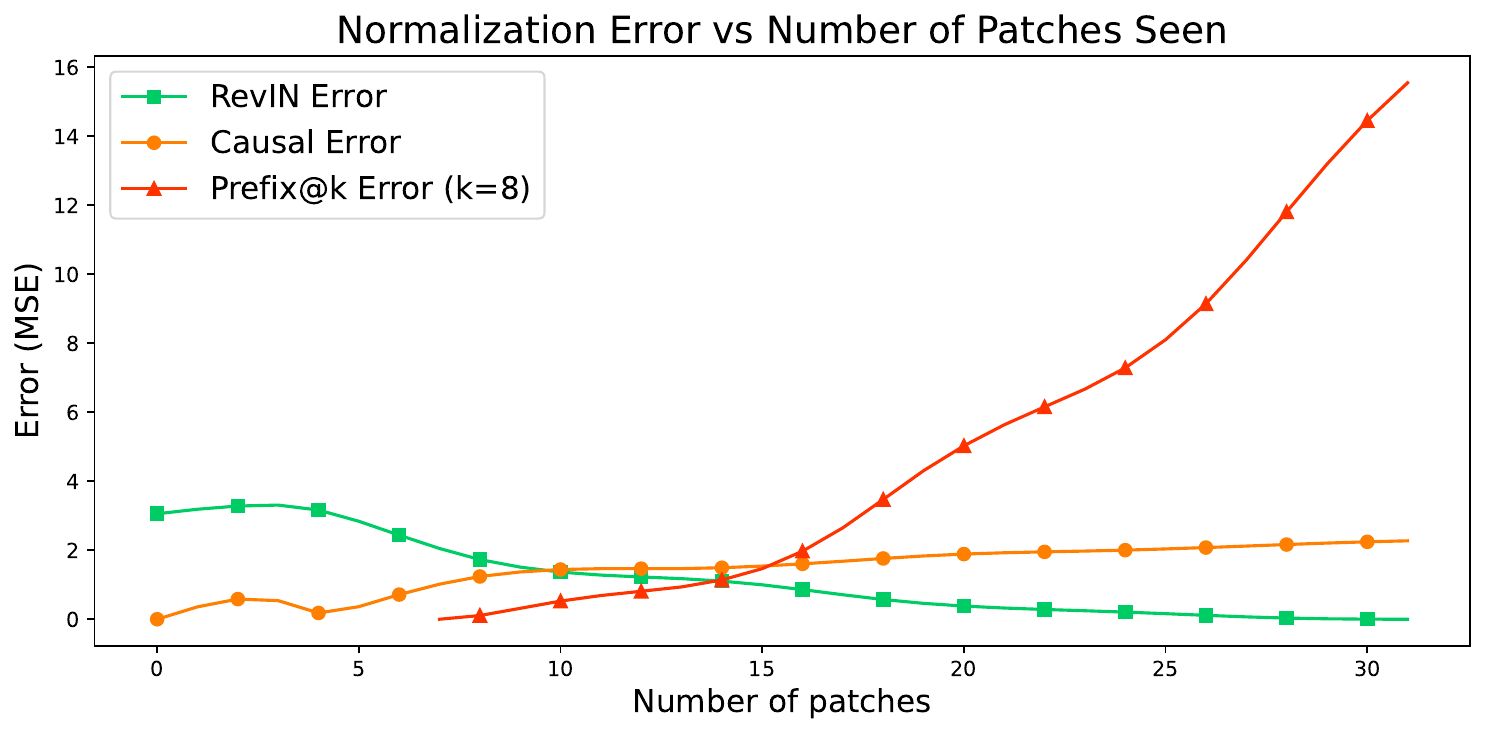}
        \caption{Normalization error against the more accurate normalization inference strategy.}
        \label{fig:normalization_error}
    \end{minipage}

    \vspace{0.5cm} 
\end{figure}

Figure \ref{fig:normalization_example} illustrates the behavior of the evaluated normalization strategies on a synthetic 
non-stationary sinusoidal signal.
The \revin{} strategy, leveraging global context statistics, produces the most stable representation with consistent zero-mean centering and 
unit-variance scaling across the entire sequence. However, as noted in Section \ref{normalization}, this result is achieved through non-causal information 
leakage during training. In contrast, the \wurevin{} strategy achieves a perfect representation only for the patch \(k\); once the distribution shifts, the 
normalization fails to track the evolving mean, leading to significant bias in later segments.
The \causalrevin{} strategy preserves a well-scaled representation but introduces visible non-smoothness. Because statistics are updated 
at each patch, the resulting normalized signal appears disjointed. While this approach handles the distribution shift more 
robustly than the \wurevin{} method, staying closer to a zero-mean representation, the lack of inter-patch continuity forces the model to manage
these abrupt scaling transitions.

Here, by the most "accurate" normalization strategy, we refer to the method that, given a context signal and a forecasting horizon, normalizes the input signal 
using identical statistics for all time steps. These statistics are computed once over the entire context available prior to the forecast threshold, 
thereby ensuring the absence of statistical leakage and avoiding any discontinuities between time steps.

Figure \ref{fig:normalization_error} quantifies the normalization discrepancy between each training strategy and the global inference-time optimum.
The \revin{} approach exhibits a high initial error that monotonically diminishes to zero. This trajectory reflects the "look-ahead" nature of 
the strategy: early patches are normalized using global future information during training that is not supposed to be available during inference, leading to 
a maximal mismatch that vanishes only at the final sequence element. Conversely, the \causalrevin{} strategy begins with zero error—as the first 
patch is perfectly aligned with its only available context—but shows a divergent error trend. As the sequence progresses, the causal prefix 
statistics increasingly deviate from the more accurate statistics, reaching maximum discrepancy at the last patch. Finally, the \wurevin{} strategy 
shows a sharp error divergence immediately following the initial prefix, highlighting its failure to track non-stationary signal shifts 
beyond the fixed prefix window.

\subsection{Training Without Normalization}\label{withoutnorm}

\begin{figure}[h]
    \centering
    \begin{minipage}[t]{0.49\linewidth} 
        \centering
        \includegraphics[width=1\linewidth]{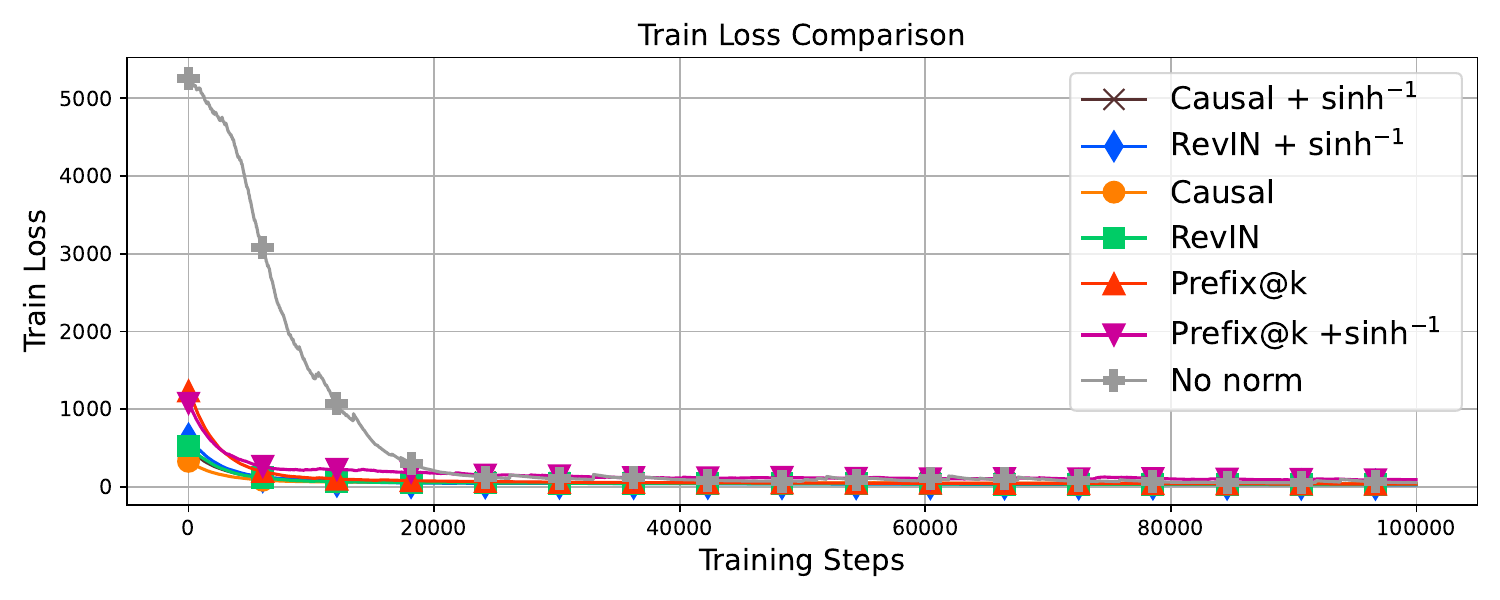}
        \caption{Training loss for first 100K steps.}
        \label{fig:lossall_nonorm}
    \end{minipage}
    \begin{minipage}[t]{0.49\linewidth} 
        \centering
        \includegraphics[width=1\linewidth]{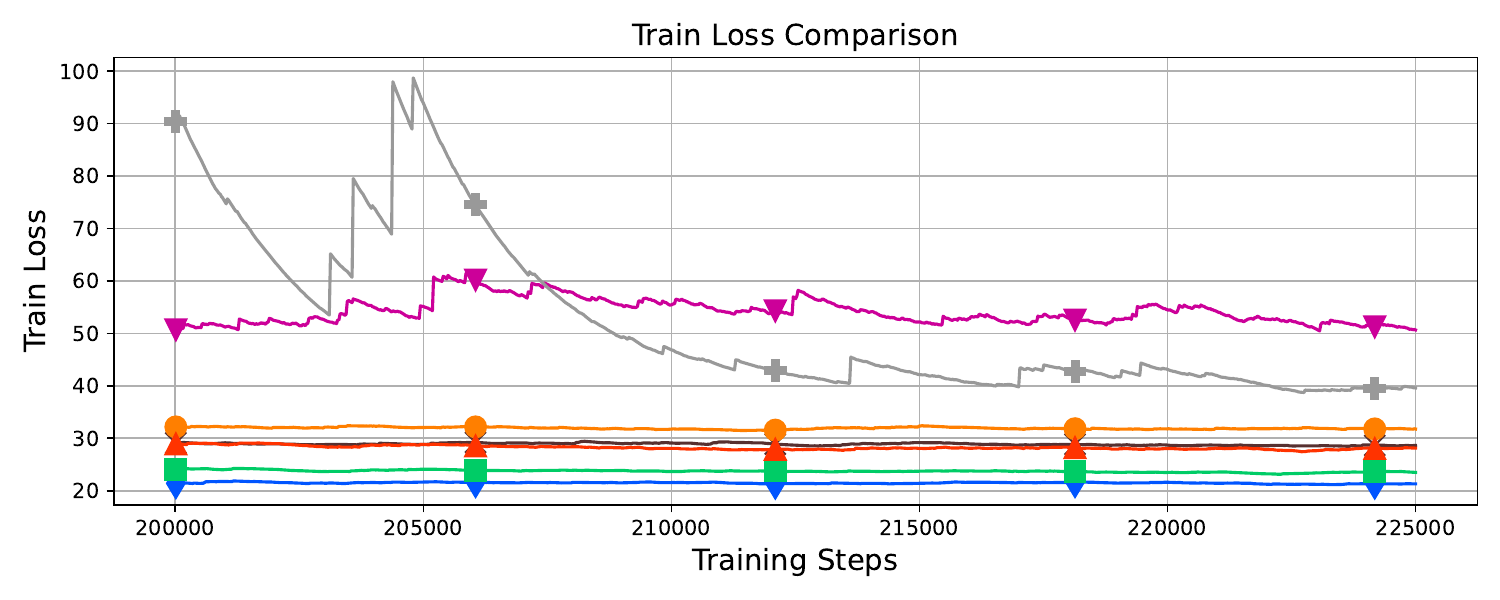}
        \caption{Training loss for 200K-225K steps.}
        \label{fig:lossend_nonorm}
    \end{minipage}%
\end{figure}

To further contextualize the performance of the evaluated normalization strategies, we additionally trained a model without any normalization strategy, using the raw input signal directly.
As shown in Figures~\ref{fig:lossall_nonorm} and~\ref{fig:lossend_nonorm}, the model trained without normalization exhibits a substantially higher training loss at initialization compared 
to all normalized variants. This behavior is expected, as the model must learn to operate directly on the raw, unnormalized signal and does not benefit from the denormalization layer, which rescales the model's predictions to the original signal range.
Nevertheless, as training progresses, the loss of the non-normalized model decreases and eventually matches, and even reaches a lower value than the worst-performing normalization strategy (\wurevinasinh{}) in the later stages of training.

\begin{figure}[h]
    \centering
    \includegraphics[width=\linewidth]{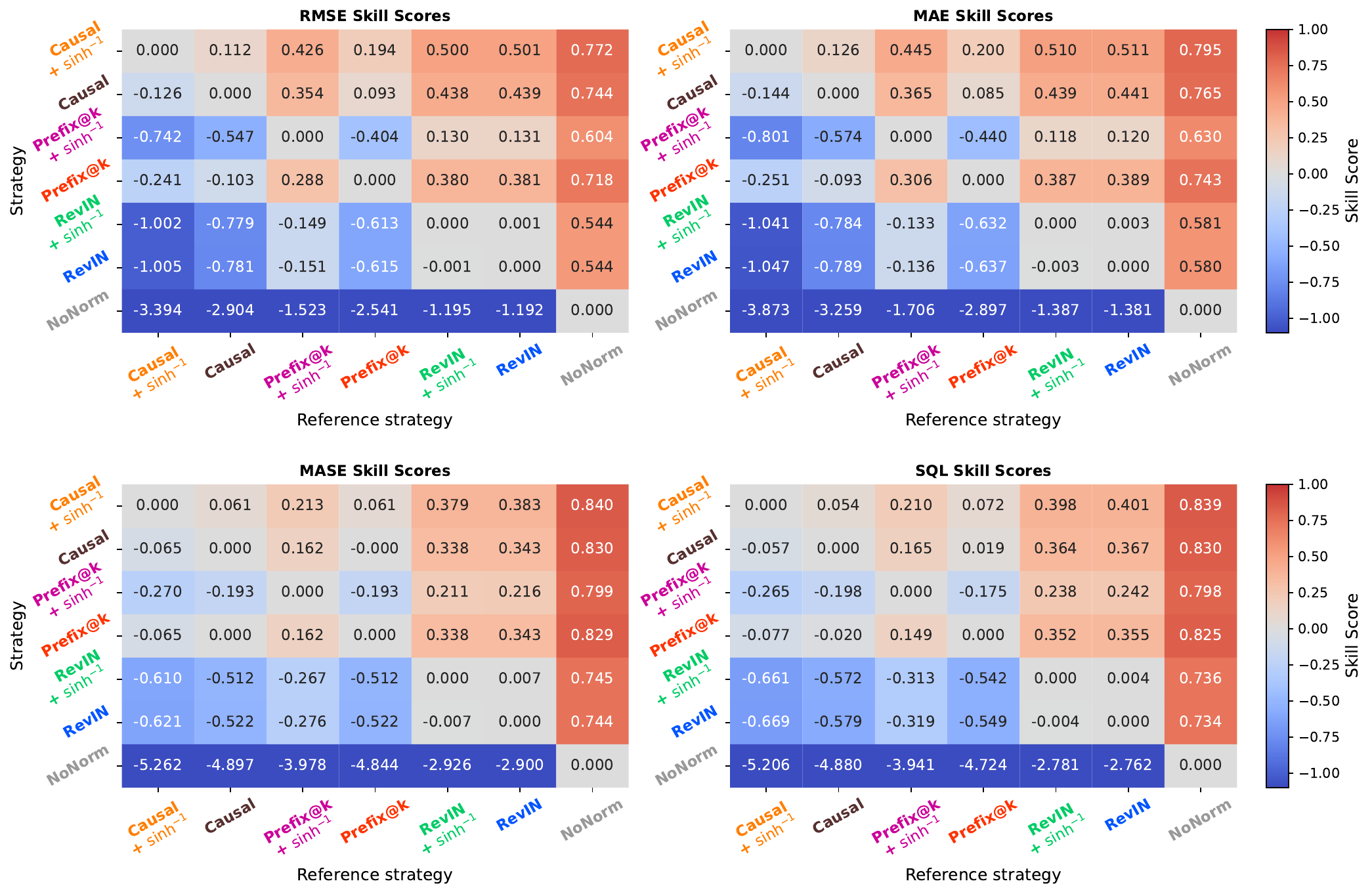}
    \caption{Pairwise skill scores for each normalization strategy, aggregated over context lengths \((128, 256, 512)\), datasets, and forecasting horizons \((32, 64, 96, \dots, 512)\).}
    \label{fig:skillscores_norm}
\end{figure}

Considering the forecasting performance, the non-normalized model achieves the worst results across all metrics and context lengths, as shown in Figure~\ref{fig:skillscores_norm}. 
This outcome is expected. 
These findings underscore the critical importance of normalization in training large-scale time-series models, as the absence of 
normalization results in substantially degraded performance compared to even the least effective normalization strategy.

\subsection{Limitations of this Work}\label{limitations}

Our study focuses on univariate time-series, leaving the systematic extension to multivariate settings for future work. 
While the presented normalization srategies can extend to multivariate signals by computing statistics per channel, 
interactions across variables and cross-channel normalization effects may introduce additional dynamics that we do not explore here.
Furthermore, our analysis is restricted to Transformer-based architectures. We do not study normalization strategies 
for state-space models~\citep{stateflow} or xLSTM-based approaches~\citep{tirex} trained under comparable conditions. 
Our analysis is restricted to mean–-variance-based normalization strategies, which remain the dominant choice in current large-scale time-series models. 
Alternative normalization families are therefore not considered in this work.

\subsubsection*{Acknowledgments}

The authors acknowledge ANR – FRANCE (French National Research Agency) for its financial support of SHARP project ANR-23-IAS4-0003.\\
This project was provided with computer and storage resources by GENCI at
IDRIS thanks to the grant 2025-AD011016510 on the supercomputer
Jean Zay's the V100 partition and CRIANN (Centre des Ressources Informatiques et Applications Numérique de Normandie, France) for providing computational resources.

\end{document}